\documentclass[pageno,noline]{jpaper}

\usepackage[normalem]{ulem}
\usepackage{caption}
\usepackage{fancyhdr}
\usepackage{hyperref}
\usepackage{graphicx}
\usepackage{amsmath}
\usepackage{amsfonts}
\usepackage[T1]{fontenc}
\usepackage{algpseudocode}
\usepackage[linesnumbered, ruled]{algorithm2e}
\usepackage{multirow}
\usepackage[flushleft]{threeparttable}
\usepackage[export]{adjustbox}
\usepackage{color}
\usepackage{fancyhdr}

\usepackage{array}
\newcolumntype{P}[1]{>{\centering\arraybackslash}p{#1}}
\newcolumntype{M}[1]{>{\centering\arraybackslash}m{#1}}
\newcolumntype{L}[1]{>{\raggedright\let\newline\\\arraybackslash\hspace{0pt}}m{#1}}

\SetKwRepeat{Do}{do}{while}%
\algnewcommand{\LeftComment}[1]{\(\triangleright\) #1}%
  
\newcommand{\var}{\texttt}
\algrenewcommand\textproc{}
\newcommand{\eg}{\emph{e.g.,}}
\newcommand{\ie}{\emph{i.e.}}

\SetKwFor{Loop}{Loop}{}{EndLoop}
\SetKw{Break}{break}

\begin{document}

\fancypagestyle{title}{%
  \setlength{\headheight}{22pt}%
  \fancyhf{}% No header/footer
  \renewcommand{\headrulewidth}{0pt}% No header rule
  \renewcommand{\footrulewidth}{0pt}% No footer rule
   \fancyhead[R]{Preliminary version \\ Final version to appear in ASPLOS 2019}
}%
\title{Packing Sparse Convolutional Neural Networks for Efficient Systolic Array Implementations: Column Combining Under Joint Optimization}
\author{
 H. T. Kung\\
 Harvard University\\
 Cambridge, MA, USA
 \and
 Bradley McDanel\\
 Harvard University\\
 Cambridge, MA, USA
 \and
 Sai Qian Zhang\\
 Harvard University\\
 Cambridge, MA, USA
}

\date{}
\maketitle
\thispagestyle{title}

\begin{abstract}
This paper describes a novel approach of packing sparse convolutional neural networks for their efficient systolic array implementations. By combining subsets of columns in the original filter matrix associated with a convolutional layer, we increase the utilization efficiency of the systolic array substantially (\eg~4x) due to the increased density of nonzeros in the resulting packed filter matrix. In combining columns, for each row, all filter weights but one with the largest magnitude are pruned. We retrain the remaining weights to preserve high accuracy. We demonstrate that in mitigating data privacy concerns the retraining can be accomplished with only fractions of the original dataset (e.g., 10\% for CIFAR-10). We study the effectiveness of this joint optimization for both high utilization and classification accuracy with ASIC and FPGA designs based on efficient bit-serial implementations of multiplier-accumulators. We present analysis and empirical evidence on the superior performance of our column combining approach against prior arts under metrics such as energy efficiency (3x) and inference latency (12x).

\end{abstract}

\section{Introduction}
Many recent hardware-based state-of-the-art deep learning accelerators use systolic arrays for efficient implementations of convolutional neural networks (CNNs). They leverage properties of systolic arrays such as parallel processing under the dataflow architecture, regular layout of processing elements, efficient inter-processor communication, and minimized I/O by being able to reuse the same data fetched from the memory many times~\cite{kungLeisersonSystolicArrays1978,kung1982systolic,Rojas1996}.
These systems, such as the
Google TPU~\cite{jouppi2017datacenter}, the ShiDianNao
accelerators~\cite{du2015shidiannao}, and numerous other efforts,
including~\cite{Merritt2018ARM},~\cite{chen2017eyeriss},~\cite{wang2017chain},~\cite{zhang2016cambricon},~\cite{wei2017automated},
have achieved low power consumption and high throughput.

In recent years, there have also been significant algorithmic advances for CNNs which have enabled orders of magnitude reduction in both the number of network parameters and the amount of computation for inference compared to prior well studied networks such as
VGG-16~\cite{simonyan2014very}. One of the more important model reduction techniques is weight pruning~\cite{han2015deep}, which has
shown that the majority of weights in a trained CNN can be pruned (set to 0) without significantly impacting the accuracy of the network.

However, it can be challenging to efficiently utilize the regular structure of systolic arrays given that these nonzero CNN weights are distributed in an irregular manner. In traditional approaches, zero weights still occupy systolic cells in the systolic array. 

In this paper we propose a novel approach, called \textit{column combining}, which can pack sparse convolutional networks for their efficient
systolic array implementations. In combining columns to increase the percentage of nonzeros in a packed CNN, within a group of combined columns, we prune all weights on conflicting rows but the one with the largest magnitude. We then bring up the classification accuracy of the pruned network via retraining. We iteratively perform these column-combining and network-retraining steps to improve both utilization efficiency of the systolic array and the classification accuracy of the network until a target model size is reached.

Thus our proposed column combining approach leverages a joint optimization opportunity present in CNNs. That is, for a CNN, we can optimize its topologies to fit the structure of the underlying computing hardware such as systolic arrays, while preserving most of its classification accuracy via network retraining. 

The main contributions of the paper are summarized as follows:

\begin{itemize}
\item \textbf{Column combining algorithm} (Section~\ref{sec:column-combine}) for packing sparse CNNs with unstructured sparsity for their efficient systolic array implementations. The method can retrain remaining filter weights after column-combine pruning using only fractions of the original training dataset in mitigating data privacy concerns (Section~\ref{sec:limited}). To ease data routing, a row permuting scheme is described (Section~\ref{sec:row-permute}) for a systolic array to output contiguous data items for those columns to be combined together in the systolic array of the next layer.
\item \textbf{Joint optimization methodology} (Algorithm 1 in Section~\ref{sec:column-combine}) aiming at achieving two objectives simultaneously---high utilization efficiency of the systolic array and high classification accuracy of the CNN. The methodology leverages opportunities presented in CNNs in training for both efficiency and accuracy simultaneously.
\item \textbf{Bit-serial systolic arrays} (Section~\ref{sec:bitserial-array}) to allow efficient multiplexing of multiple data streams into a single column in the array in support of column combining. In addition, bit-serial implementations provide flexibility in supporting accumulations at various precisions for the multiplier-accumulators (MACs) of systolic cells. In this work, we assume bit-serial implementations of 32-bit accumulations, except in Section~\ref{sec:asic-single} where we use 16-bit accumulations, and 8-bit weights and input data. Our approach extends naturally to other precisions.
\item \textbf{Cross-layer pipelining} (Section~\ref{sec:cross-layer-pipeline}) for CNN inference over a series of systolic arrays, one for each layer. This dramatically reduces the inference latency per input sample (\eg~an input image) and is especially important for realtime applications where it is common to process samples one at a time. 
\item \textbf{ASIC and FPGA designs} to validate performance gains of our approaches
(section~\ref{sec:eval}) in energy efficiency, area efficiency and latency.
\end{itemize}

%Section~\ref{sec:background} covers related work on systolic arrays, ASIC and FPGA designs, and recent algorithmic advances for efficient CNNs utilized in this work. Section~\ref{sec:column-combine} introduces the column combining algorithm which jointly optimizes for both classification accuracy and systolic array utilization. Section~\ref{sec:column-combine-design} describes the proposed systolic array design for the packed filter matrix format generated by the column combining algorithm. Section~\ref{sec:combine-analysis} provides empirical analysis on the parameters of column combining (Algorithm~\ref{alg:iterative-training}). Section~\ref{sec:limited} discusses model retraining with limited data to address privacy concerns. Section~\ref{sec:eval} evaluates the column combining system (Section~\ref{sec:column-combine-design}) with both ASIC and FPGA implementations across throughput, accuracy, area, and latency, comparing with other state-of-the-art systems. 

%Anonymous links to our PyTorch~\cite{paszke2017automatic} code used to train CNNs with column combining and Verilog code for both FPGA and ASIC synthesis is available at \url{https://docs.google.com/document/d/1rLoyVUMeOfQ26BtjNzx6AL1FdXucW-68b4ctP7ac2OI}.

\section{Background and Related Work}
\label{sec:background}

In this section, we first provide a brief review of the basic principle of using systolic arrays for the implementations of CNNs and introduce terminologies that we will use throughout. Then, we review related ASIC and FPGA accelerators for CNN inference, advances in CNN design, weight pruning, and input and weight quantization, all of which have led to large reductions in both model size and computation cost for training and inference.

\subsection{Systolic arrays for Convolutional Layers}
\label{sec:conv-layer}
\begin{figure*}
\centering
\subfloat[Computation of a convolutional layer]{%
    \includegraphics[height=3.3cm]{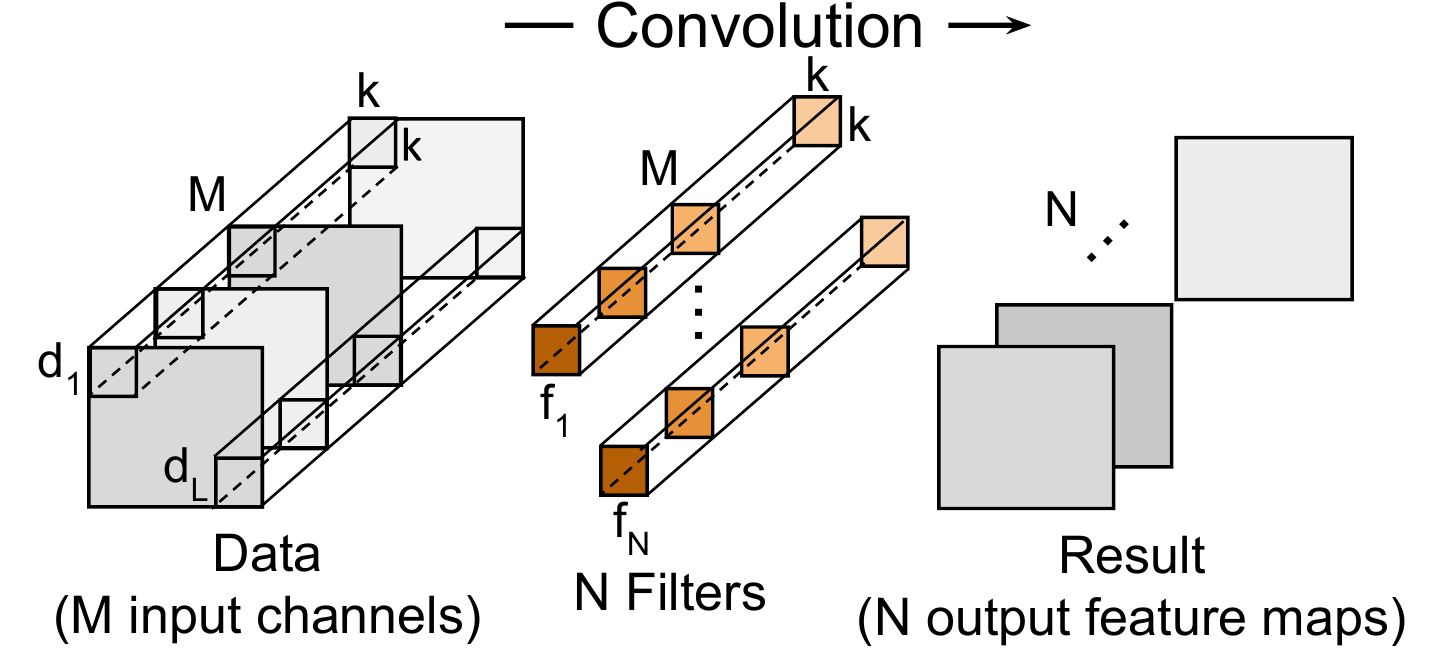}
    \label{fig:conv-layer}}\hfill
\subfloat[Equivalent matrix-matrix multiplication]{%
    \includegraphics[height=3.3cm]{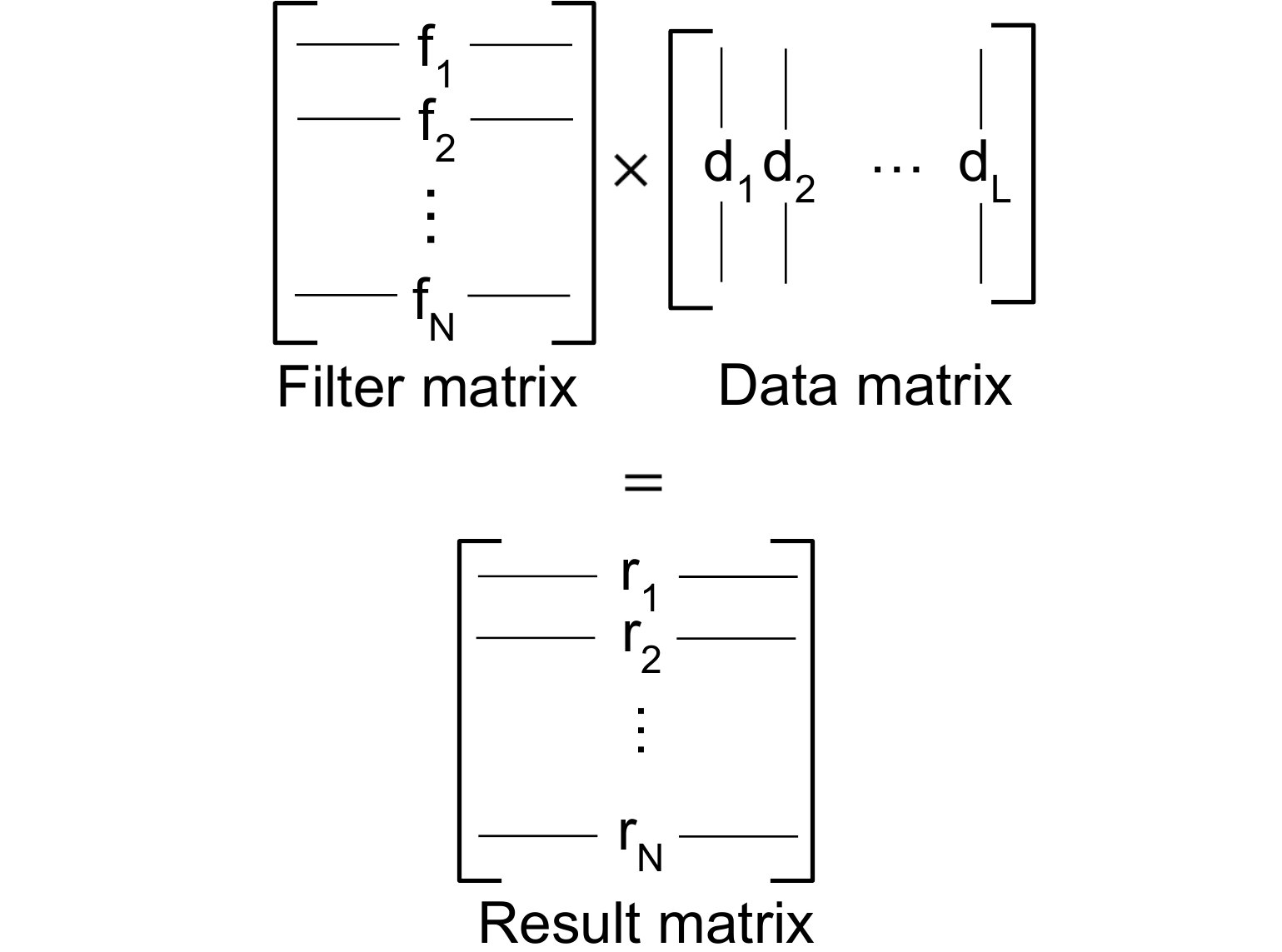}
    \label{fig:conv-mat}}\hfill
\subfloat[Weight-stationary Systolic Array]{%
    \includegraphics[height=3.3cm]{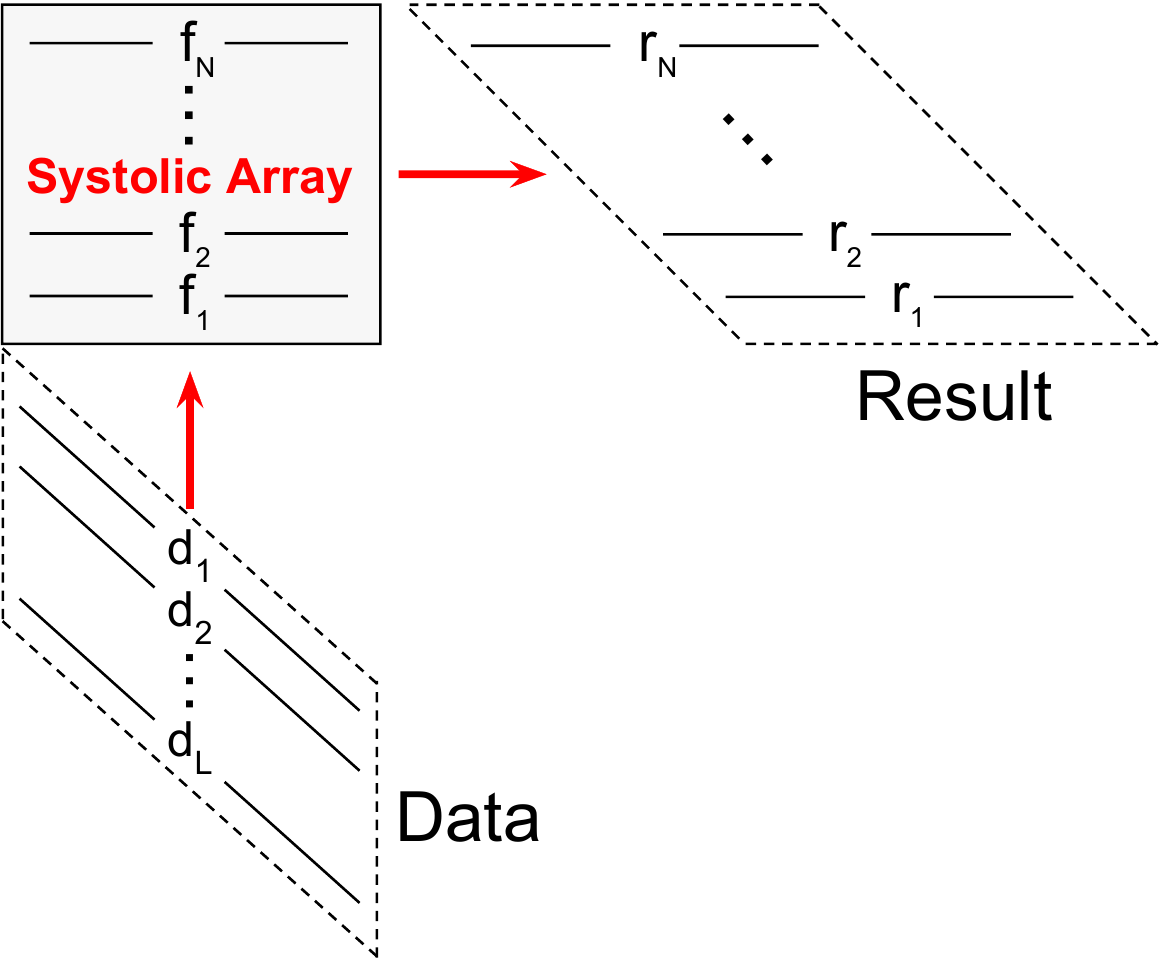}
    \label{fig:systolic-mm-with-skew}}
\caption{(a) Computation of a convolutional layer, (b) viewed as a matrix multiplication, and (c) deployed in a weight-stationary systolic array, with skewed input data and output results.}
\end{figure*}

It is well known that the bulk of the computation of a convolutional layer for a CNN can be viewed as a matrix-matrix multiplication. Suppose that a convolutional layer has N filters operating on a data volume of depth M, as depicted in Figure~\ref{fig:conv-layer}. Then, the result of the convolution computation is the matrix product of the filter matrix and the data matrix, as depicted in Figure~\ref{fig:conv-mat}.

Figure~\ref{fig:systolic-mm-with-skew} depicts a systolic array design for this matrix multiplication. It is a \textit{weight-stationary} systolic array in the sense that filter weights stored in the array will not move during computation, whereas input data continuously move bottom-to-top and result data accumulate left-to-right. For systolic array synchronization, items in the data and result matrices are properly skewed, as shown in the figure. We assume throughout the paper this weight-stationary systolic array design.

\subsection{ASIC and FPGA Accelerators for CNNs}
Over the past several years, there has been extensive work on constructing efficient ASIC and FPGA designs for CNNs which generally consider well studied networks such as LeNet-5~\cite{lecun1998gradient}, AlexNet~\cite{krizhevsky2012imagenet}, and VGG-16~\cite{simonyan2014very} including~\cite{sankaradas2009massively,qiu2016going,zhang2015optimizing, zhang2016caffeine,zhao2017accelerating,zhao2017aep,park2016fpga,park2017scale,reagen2016minerva,shen2017escher,shen2017maximizing,song2017pipelayer,bang201714,baiduhotchips}. One of the main considerations for such systems is minimizing the number of off-chip DRAM accesses for fetching the CNN weights, input samples and intermediate layer results, as these incur significant energy consumption~\cite{han2015deep}. Therefore, a main focus of accelerator design is mapping CNN computations in such a way that input and weights are fetched only once for all usages within a layer~\cite{chen2014diannao,chen2014dadiannao,du2015shidiannao,ma2017optimizing}. Another orthogonal direction is designing memory systems that are more suitable to the regular structure of CNN inference computation~\cite{rhu2016vdnn,clemons2016patch,wang2016re,wang2017dlau}. In Section~\ref{sec:eval-asic}, we show our design achieves state-of-the-art performance in terms of energy efficiency.

FPGAs allow for faster development time and therefore are often used to explore various new research areas for CNNs, such as low-precision and binary networks~\cite{courbariaux2016binarized,umuroglu2017finn}, novel training regimes~\cite{dicecco2017fpga}, and model compression through weight pruning or novel CNN structures~\cite{han2017ese,ding2017c}. In Section~\ref{sec:eval}, we validate the performance of our filter matrix packing algorithm with an FPGA implementation. Additionally, we compare our implementation to previous state-of-the-art FPGA results~\cite{truenorthcifar,park2016fpga,ding2017c,zhao2017accelerating}.

\subsection{CNNs with Simplified Filter Structures}
\label{sec:cnn}
\begin{figure}
    \centering
    \includegraphics[width=0.9\columnwidth]{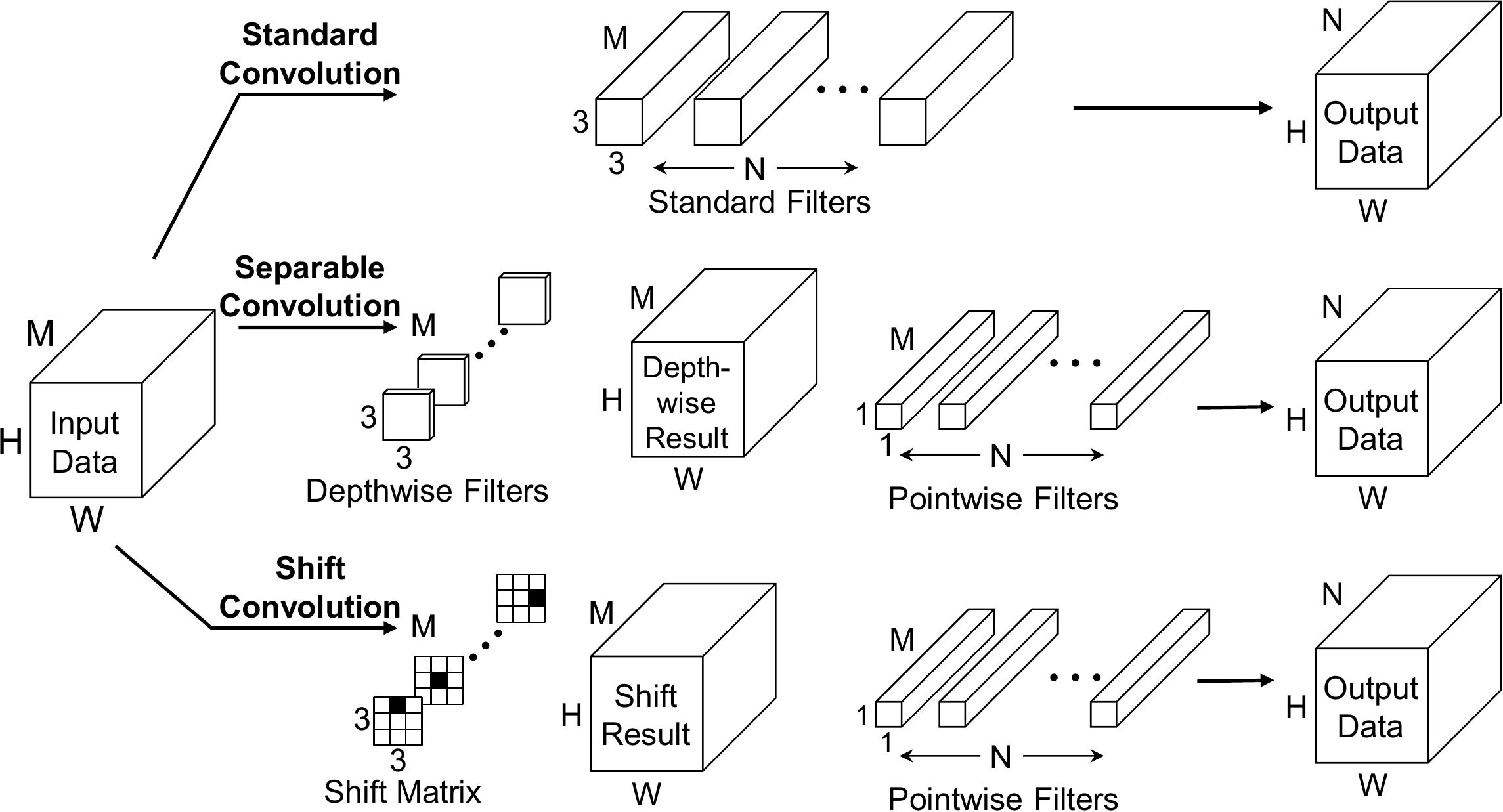}
    \caption{Standard, separable, and shift convolution.}
    \label{fig:convolution}
\end{figure}

Figure~\ref{fig:convolution} compares standard CNNs to two recent CNN variants, \textit{separable convolution}~\cite{chollet2016xception,howard2017mobilenets} and \textit{shift convolution}~\cite{wu2017shift}, as shown in Figure~\ref{fig:convolution}. Separable convolution decouples a standard convolution layer into two smaller convolution layers (depthwise convolution and pointwise convolution) in order to reduce both model size and amount of computation. Each pointwise filter has only a single weight for each channel, and therefore does not utilize neighboring pixels in the spatial dimensions (width and height). Shift convolution replaces the depthwise convolution layer with a shift operation that does not require any learned weights. In Section~\ref{sec:column-combine-design}, we leverage shift convolution to construct a network that consists only of pointwise layers, as it greatly simplifies the structure of computation in each layer. 

%Shift convolution takes separable convolution a step further, by replacing the depthwise convolution layer with a shift operation that does not require any learned weights. The shift operation spatially translates each input channel a small amount (one pixel) and adds zero padding to maintain the input shape. For instance, the first input channel of shift convolution as depicted in Figure~\ref{fig:convolution} is translated 1 pixel upward, then the top row of pixels are removed and a row of zeros are added to the bottom. By applying small shifts to the input channels, each output pixel produced by the following pointwise layer utilizes neighboring pixels across the channels. In our design described in Section~\ref{sec:column-combine-design}, we leverage shift convolution to construct a network that consists only of pointwise layers, as it greatly simplifies the structure of computation in each layer. 

\subsection{Weight Pruning During Training}
\label{sec:prune}
Weight pruning methods aim to reduce the number of weights in a trained CNN by removing (pruning) unimportant weights. These pruning techniques have shown that many well studied networks such as AlexNet and VGG-16 have a large number of weights (up to 90\%) that can be pruned without any impact on classification accuracy~\cite{wen2016learning,narang2017block,gray2017blocksparse,huang2017condensenet,he2017channel,luo2017thinet}.

In Section~\ref{sec:column-combine}, we propose an iterative pruning procedure, similar to CondenseNet~\cite{huang2017condensenet}, but at the finest granularity of individual weights. This iterative pruning method gradually removes the smallest magnitude weights during training. This leads to sparse models (as low as 10\% nonzero in each convolution layer) which still achieve similar performance to the baseline dense models. Additionally, as outlined in Section~\ref{sec:column-combine}, we prune weights in such a way as to improve the utilization efficiency of the CNN when deployed in the systolic array design for sparse CNNs described in Section~\ref{sec:column-combine-design}.

\subsection{Input and Weight Quantization}
Quantization is another direction in accelerating inference computations. In this work, we take a simple linear fixed-point quantization scheme~\cite{lin2016fixed}. We quantize both the inputs and weights to an 8-bit fixed-point representation from the 32-bit float-point representation~\cite{lin2015neural,gysel2016hardware} used during training. This quantization has been shown to lead to minimal performance degradation even on challenging datasets~\cite{lin2016fixed}. Within a layer, the accumulation is done with 32-bit integers, which adds complexity to the bit-serial systolic array design and is discussed in Section~\ref{sec:bitserial-array}.

\section{Column Combining}
\label{sec:column-combine}
As discussed in Section~\ref{sec:prune}, training a CNN with weight pruning leads to small but highly sparse models with unstructured nonzero weights, which is not directly amenable to efficient implementation in systolic arrays traditionally designed for dense matrix-matrix multiplication. In this section, we propose a column combining algorithm, which is an iterative training procedure that jointly optimizes the CNN for both classification accuracy and utilization efficiency when deployed in the proposed systolic array described in Section~\ref{sec:column-combine-design}.

\subsection{Terminologies and definitions}
\label{sec:terminologies}
Suppose that we are given the filter matrix of weights associated with a convolutional layer of the CNN (see Figure~\ref{fig:conv-layer}). The columns of this filter matrix which have nonzero weights on a row are said to be \textit{conflicting} on the row, and the row is said to be a \textit{conflicting row} for these columns. 
By \textit{column combining}, we mean combining a given group of columns into a single \textit{combined} column. In a combined column, for the columns which conflict on a row, all nonzero weights on the row are pruned except for the one with the largest magnitude. We refer this pruning process as \textit{column-combine pruning}.

We say a group of columns has $x$ conflicts if a total of $x$ weights will be pruned when combining columns in the group. We say that a group of columns meets the \textit{limited-conflict condition} for certain $\gamma$ value, if the group has at most $\gamma$ conflicts per row on average. The $\gamma$ value can less than 1. For example, if $\gamma = 0.5$, then for every two rows at most one weight is pruned on average.

\subsection{Column Combining Overview}
\label{sec:column-combining-overview}
Given a sparse filter matrix, we first partition it into \textit{column groups}, and then for each group we combine its columns to form a single combined column by applying column-combine pruning. We aim at achieving two objectives simultaneously. First, we pack the given sparse filter matrix into a dense matrix, called a \textit{packed filter matrix}, with as a few combined columns as possible to allow efficient systolic array implementations. Second, we minimize the impact of column-combine pruning on classification accuracy.

For high-density packing, we adopt a \textit{dense-column-first} combining policy that favors selections of combining columns which result in high-density combined columns, where the density of a column is the percentage of nonzeros in the column. For high classification accuracy, we then retrain the remaining weights after column-combine pruning. The algorithm involves some parameters: $\alpha$ (the maximum number of combined columns), $\beta$ (the initial pruning percentage number) and $\gamma$ (the average number of conflicts per row allowed for each group). Their typical values are $\alpha = 8$, $\beta = 20$ and $\gamma = 0.5$.

\begin{figure}
    \centering
    \includegraphics[width=\columnwidth]{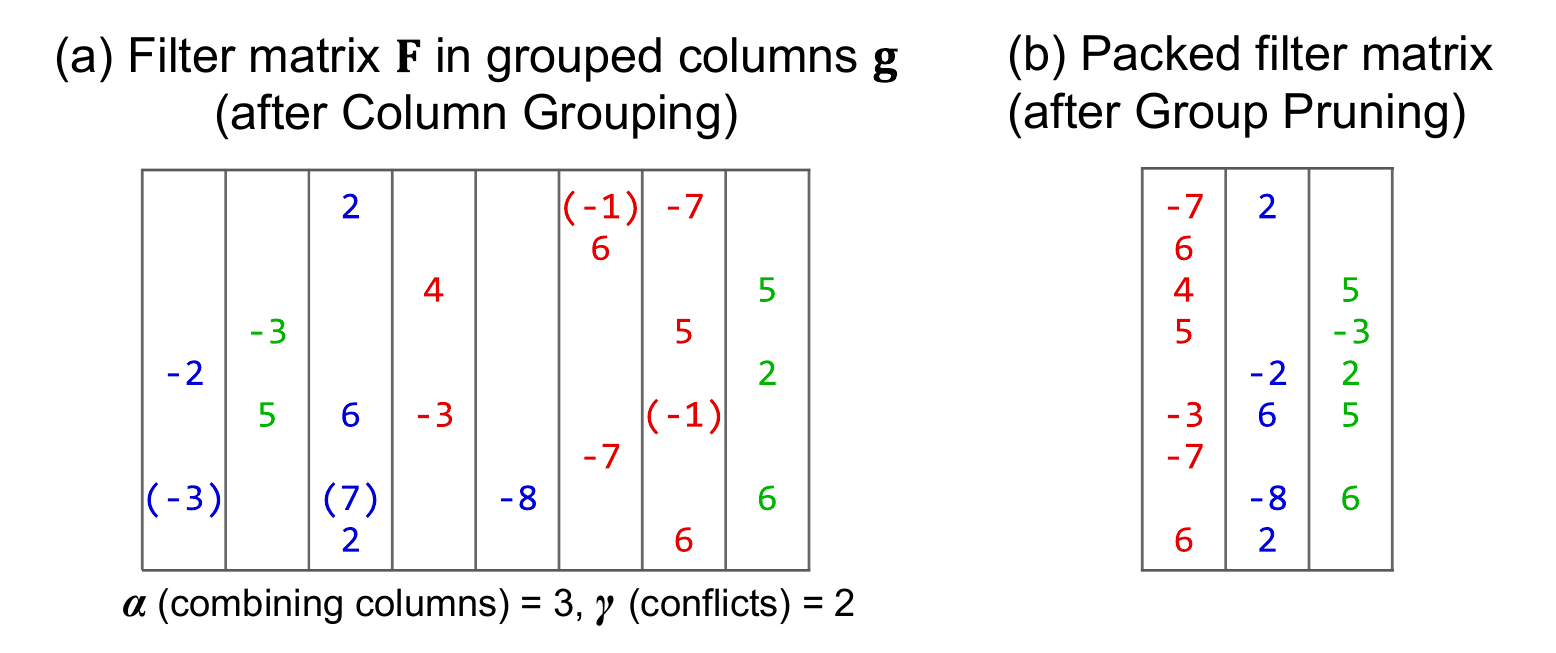}
    \caption{Example of combining columns.}
    \label{fig:column-combine}
\end{figure}

Figure~\ref{fig:column-combine} depicts a column combining example. In (a), a filter matrix $\mathbf{F}$, associated with a sparse convolutional layer, is divided along columns into three groups (blue, green, and red). The zero-valued weights in $\mathbf{F}$ due to previous pruning steps are omitted for illustration clarity. The objective of column grouping is to select columns that, when combined, achieve high packing efficiency (\ie~are mostly nonzeros). As we show in Section~\ref{sec:combine-analysis}, a high packing efficiency translates to a high utilization efficiency, as more MACs will perform useful computation by storing nonzero weights. A small number of conflicting elements $\gamma$ are allowed between the columns in a group. For instance, in the blue group, (-3) in column 1 conflicts with (7) in column 3 and -8 in columns 5. The conflicting (-3) and (-7) weights are pruned and -8 is kept as it has the largest magnitude. In (b), each group is combined into a single column in order to be loaded into a column in the proposed systolic array (as discussed in Section~\ref{sec:column-combine-design}). 

\subsection{Column Combining Algorithm}
\label{sec:combine-alg}
The column combining scheme, outlined in Section~\ref{sec:column-combining-overview}, joins columns in a sparse filter matrix that do not have significant conflicts. Algorithm~\ref{alg:iterative-training}, which calls Algorithm~\ref{alg:combine} and Algorithm~\ref{alg:group-prune}, is the top level algorithm used to train sparse CNNs that can be implemented with systolic arrays of high utilization efficiency. The training process works in an iterative fashion, where at each iteration the model is pruned and packed so that it fits more efficiently in the systolic array. In Section~\ref{sec:combine-analysis}, we provide analysis on the effect that each parameter of Algorithm~\ref{alg:iterative-training} has on both classification accuracy and utilization efficiency.

\begin{algorithm}
\caption{Iterative Training with Column Combining}
\label{alg:iterative-training}
\footnotesize
\DontPrintSemicolon
  \KwIn{$\mathbf{C}$ is a CNN with $L$ convolution layers\newline
  $\alpha$ is the maximum number of combined columns per group \newline
  $\beta$ is the initial pruning factor (\eg~$10\%$ of weights) \newline
  $\gamma$ is the number of conflicts (\ie~pruned weights) allowed on average per row for each group \newline
  $\rho$ is the target number of nonzero weights after column combining in $\mathbf{C}$ and is used as the stopping criteria 
  }
  \KwOut{$\mathbf{\hat{C}}$ is a pruned version of $\mathbf{C}$ with combined columns for each of the $L$ sparse convolution layers  \newline
  $\mathbf{G}$ are the column groups for each of the $L$ layers in $\mathbf{\hat{C}}$}
  $\mathbf{\hat{C}}$ $\gets$ $\mathbf{C}$; \;
  \LeftComment{Prune and retrain until $\rho$ target is reached}\;
  \While{$||\mathbf{\hat{C}}||_0 > \rho$}{
    \For{$l \gets 1 \text{ to } L$}{
      \LeftComment{\textbf{Step 1} Perform initial pruning by removing the smallest magnitude weights up to an $\beta$ percentage} \;
      $\mathbf{\hat{C}}_l \gets \var{prune}(\mathbf{\hat{C}}_l$, $ \beta)$; \;
      \vspace{2mm}
      
      \LeftComment{\textbf{Step 2} Form groups by combining columns} \;
      $\mathbf{G}_l \gets \var{group-columns}(\mathbf{\hat{C}}_l$, $ \alpha$, $ \gamma)$; \;
      \vspace{2mm}

      \LeftComment{\textbf{Step 3} Prune conflicts in groups} \;
      $\mathbf{\hat{C}}_l \gets \var{group-prune}(\mathbf{\hat{C}}_l$, $ \mathbf{G}_l$); \;
    }
    \LeftComment{\textbf{Step 4} Network retraining} \;
    $\mathbf{\hat{C}} \gets \var{train}(\mathbf{\hat{C}})$; \;
    $\beta \gets 0.9\cdot\beta$; \Comment{Decay $\beta$ by constant factor} \;
  }
\end{algorithm}

\begin{algorithm}
\caption{Column Grouping (\var{group-columns})}
\label{alg:combine}
\footnotesize
\DontPrintSemicolon
  \KwIn{$\mathbf{F} \in \mathbb{R}^{N\times MWH}$ a filter matrix with $N$ rows and $MWH$ columns \newline
  $\alpha$ is the maximum number of combined columns per group \newline
  $\gamma$ is the number of conflicts (\ie~nonzero weights) allowed on average per row in each group
  }
  \KwOut{$\mathbf{g}$ are the $H$ groups of columns in $\mathbf{F}$}
  
  $\mathbf{g} \gets [\{\}]$; \;
  $\mathbf{u} \gets \{1, 2, \dots,M\}$;\;
  \Loop{}{
      \LeftComment{exit if all columns are in a group} \;
      \lIf{$\mathbf{u} = \emptyset$}{\Break;}
  
      $c \gets \var{pop}(\mathbf{u})$; \Comment{select ungrouped column $c$}\;
  
      \LeftComment{compute densities $\mathbf{d}$ between $\mathbf{g}$ and $c$} \;
      $\mathbf{d} \gets$ \var{pairwise-density}($\mathbf{F}$, $\mathbf{g}$, $c$); \;
      
      \LeftComment{compute number of conflicting weights between $\mathbf{g}$ and $c$} \;
      $\mathbf{o} \gets$ \var{pairwise-overlap}($\mathbf{F}$, $\mathbf{g}$, $c$); \;
     
      \LeftComment{select the group with the highest density while satisfying both the group size $\alpha$ and the overlap $\gamma$ constraints} \;
      $\mathbf{g}_h \gets$ \var{densest-group($\mathbf{g}$, $\mathbf{d}$, $\mathbf{o}$, $\alpha$, $\gamma$)};\;
      
      $\mathbf{g}_h \gets g_h \cup c$; \Comment{add $c$ to the group $\mathbf{g}_h$}\;
  }
\end{algorithm}

\begin{algorithm}
\caption{Column-Combine Pruning (\var{group-prune})}
\label{alg:group-prune}
\footnotesize
\DontPrintSemicolon
  \KwIn{$\mathbf{F} \in \mathbb{R}^{N\times MWH}$ a filter matrix with $N$ rows and $MWH$ columns \newline
  $\mathbf{g}$ are the $H$ groups of columns in $\mathbf{F}$
  }
  \KwOut{$\mathbf{\hat{F}}$ is $\mathbf{F}$ with conflicting entries within each group pruned }
  $\mathbf{\hat{F}} \gets \mathbf{F}$; \;
  \LeftComment{Iterate over $H$ groups and prune all but one entry per row in each group} \;
  \For{$h \gets 1 \text{ to } H$}{
    $\mathbf{\hat{F}}_h \gets \mathbf{\hat{F}}[:, \mathbf{g}_h]$; \Comment{Submatrix of $\mathbf{\hat{F}}$ containing columns in $\mathbf{g}_h$} \;
    \For{$n \gets 1 \text{ to } N$}{
        $w \gets \var{max}(|\mathbf{\hat{F}}_h[n]|)$; \Comment{Find largest magnitude weight $w$ in row}\;
        $\var{found} \gets$ false; \;
        \For{$m \gets 1 \text{ to } \textup{\var{size}}(\mathbf{g}_h)$}{
          \eIf{\textup{\var{found}} or $|\mathbf{\hat{F}}_h[n][m]| < w$}{
            $\mathbf{\hat{F}}_h[n][m] \gets 0$; \Comment{Prune (set to $0$)}\;
          }{
              $\var{found} \gets$ true; \;
          }
        }
    }
  }

\end{algorithm}

\subsection{Explanations for Column Combining Algorithm}
The limited-conflict condition assures that in column combining for each group (Algorithm~\ref{alg:combine}) at most $\gamma$ weights are pruned per row on average (Algorithm~\ref{alg:group-prune}). This helps minimize the impact of column-combine pruning on classification accuracy. The fact that each group can have at most $\alpha$ columns (e.g., $\alpha$ = 8) limits the degree of multiplexing that systolic cells (described in Section~\ref{sec:bitserial-array}) need to support, while allowing as many as $\alpha$ columns to be combined in order to achieve high packing density. 

The initial pruning in the beginning of each iteration can decrease the number of iterations required to reach a target number of nonzero weights for the sparse CNN. This is useful, when column-combine pruning is set to be less aggressive by using a relatively small $\gamma$ (e.g., $\gamma$ = 0.5) in order to minimize its impact on classification accuracy. Each iteration retrains the network resulting from the initial pruning and column-combine pruning. This mitigates the impact of these pruning operations on classification accuracy. Finally, we note that the dense-column-first combining policy is analogous to that of some popular bin-packing algorithms which pack large items first.

\subsection{Row Permutation for Contiguous Column Groups}
\label{sec:row-permute}
\begin{figure}
    \centering
    \includegraphics[width=0.95\columnwidth]{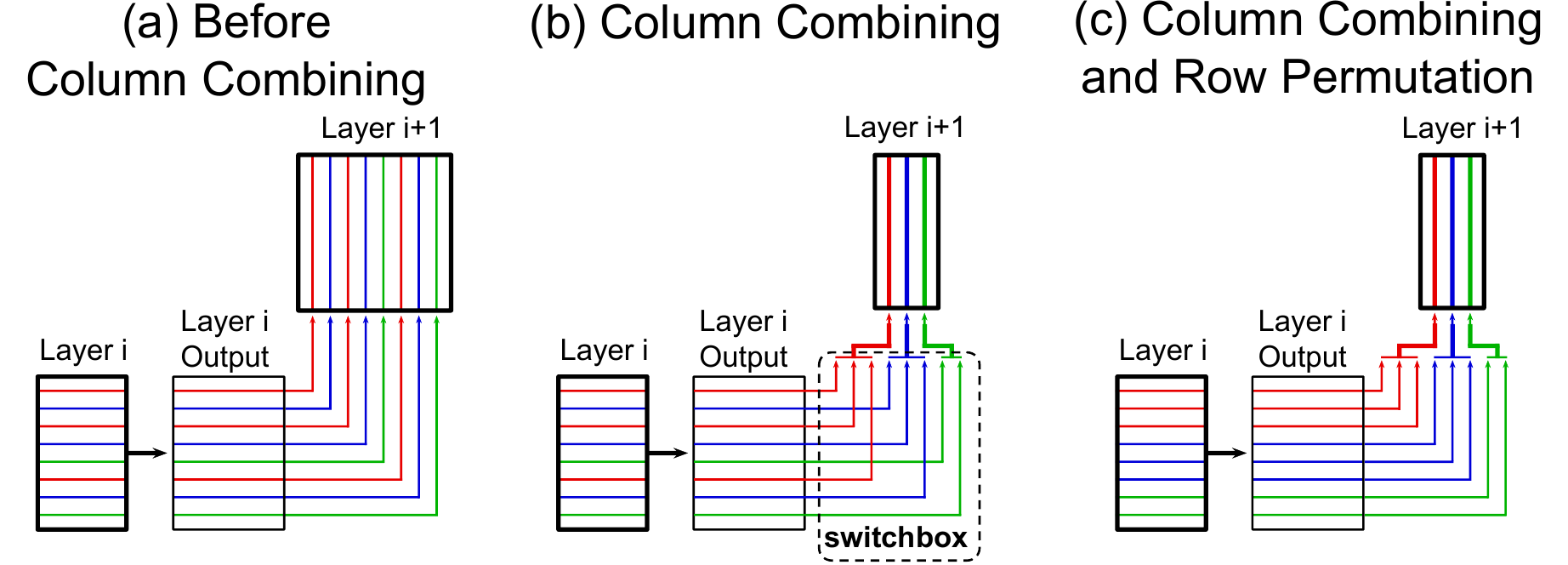}
    \caption{Applying Column Combining and Row Permutation.}
    \label{fig:row-permuting}
\end{figure}

We can permute rows of a filter matrix of the current layer to ensure that the columns from the same group for the next layer are output next to each other. In Figure~\ref{fig:row-permuting}, systolic arrays for various layers are denoted as rectangles with a thick black boundary. In (a), a systolic array of eight columns for layer i+1 is for an original sparse filter matrix of this layer consisting of three column groups, indicated in three colors, for column combining. In (b), column combining is performed on the three column groups, which results in a reduced systolic array of three columns for layer i+1. This reduced systolic array is for a packed filter matrix consisting of three combined columns. A relatively expensive switchbox function is needed for routing output of layer i to input of the reduced systolic array for layer i+1. In (c), by permuting the rows of the layer i filter matrix according to the column groups in layer i+1, we avoid the expensive switchbox. A simple counter that counts the data items in each group can be used instead.

Note that such row permutations are valid, as the column combining operation on a filter matrix are not affected by row permutations on the previous filter matrix. Thus, row permutations for layer i can be determined by the column groups of a row permuted filter matrix for layer i+1. This makes the columns within each group contiguous and removes the need to reorder the output using a switchbox at inference runtime.

\subsection{Cross-layer Pipelining of CNN Inference under Column Combining and Row Permutation}
\label{sec:cross-layer-pipeline}

In many realtime scenarios, single sample latency is a more important metric than throughput, as an input sample (\eg~an image) must be processed as soon as it is received by the system, and therefore cannot be processed in large batches. 

\begin{figure}
    \centering
    \includegraphics[width=0.85\columnwidth]{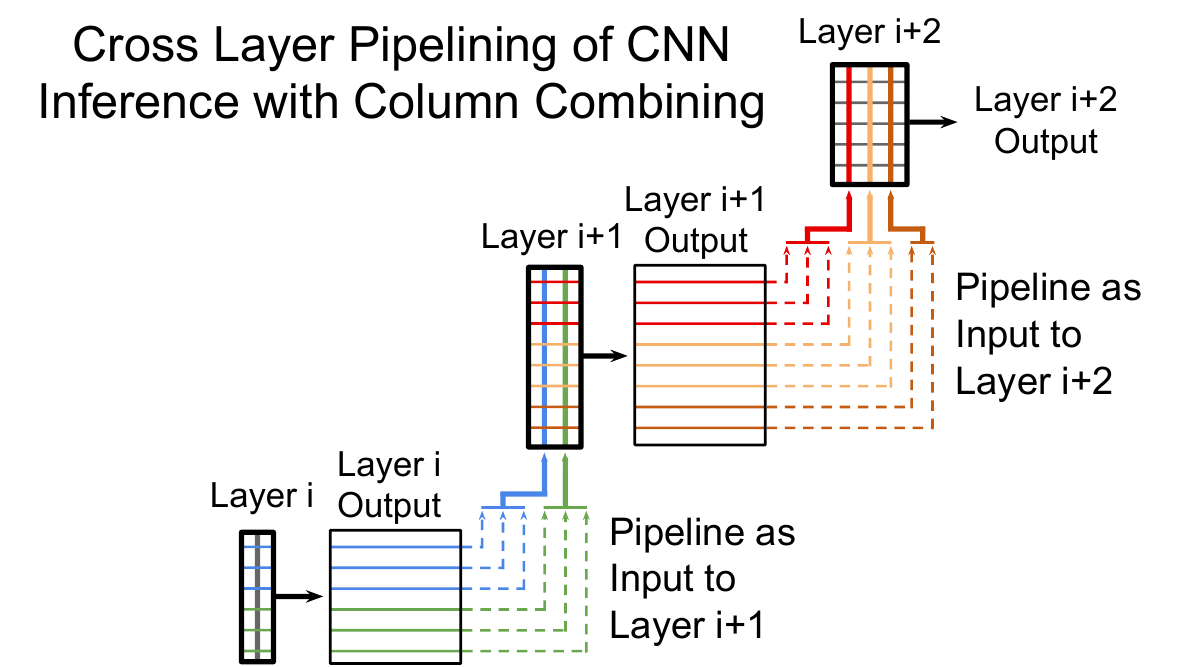}
    \caption{Pipelining CNN inference across three layers with column combining and row permutation applied to each layer.}
    \label{fig:cross-layer-pipelining}
\end{figure}

To address this concern, we propose cross-layer pipelining in the sense that we will pipe the output data elements from the previous layer immediately as input to the next layer as soon as it exits from the systolic array.  Figure~\ref{fig:cross-layer-pipelining} shows this pipelining approach for three sparse CNN layers (Layer i, Layer i+1, and Layer i+2), each deployed in a separate systolic array after column combining and row permutation have been applied to each layer. The dashed lines emitted from each layer output denote that each data element is immediately pipelined into the next layer. In Section~\ref{sec:pipeline-results}, we show that this approach reduces the inference latency for our ASIC implementation of LeNet-5 by 3.5$\times$. Having the effect of narrowing systolic arrays for convolutional layers of a CNN, column combining can reduce data skew, which further reduces the latency.

\section{Systolic Array System Description for Column Combining}
\label{sec:column-combine-design}
In this section, we describe the systolic array system and its components in support of the proposed column combining approach presented in Section~\ref{sec:column-combine}.

\subsection{Systolic Array System Components}
\begin{figure}
    \centering
    \includegraphics[width=\columnwidth]{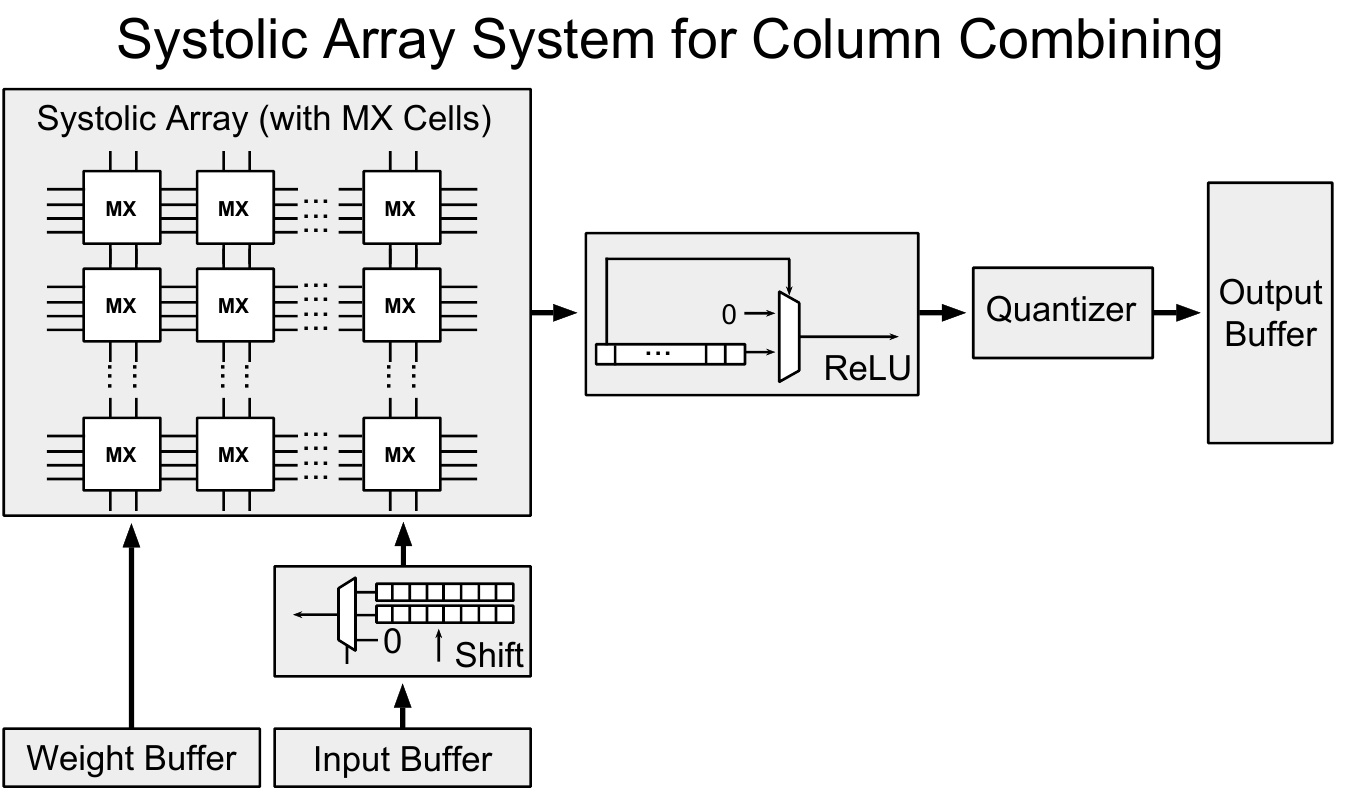}
    \caption{Systolic array system.}
    \label{fig:system-architecture}
\end{figure}

The systolic array system for column combined CNNs is shown in Figure~\ref{fig:system-architecture}. The filter weights corresponding to layers of a CNN are stored in the weight buffer. The weights for a CNN layer can then be loaded into the MX cells of the systolic array (discussed in Section~\ref{sec:bitserial-array}) before matrix multiplication is performed with the input data. The input data is loaded from the input buffer and passed through the shift block (discussed in Section~\ref{sec:shift}). The shift block performs shift operations, as depicted in Figure~\ref{fig:convolution}, and passes the output to the systolic array in a bit-serial fashion, which then performs matrix multiplication with the weights stored in the systolic cells. The output of each row in the systolic array is passed to the ReLU block (discussed in Section~\ref{sec:relu}), which performs the ReLU activation function. Finally, the result from the ReLU block is passed to the quantization block and stored in the output buffer. 

\subsection{Bit-serial Systolic Arrays}
\label{sec:bitserial-array}
\begin{figure}
    \centering
    \includegraphics[width=0.9\columnwidth]{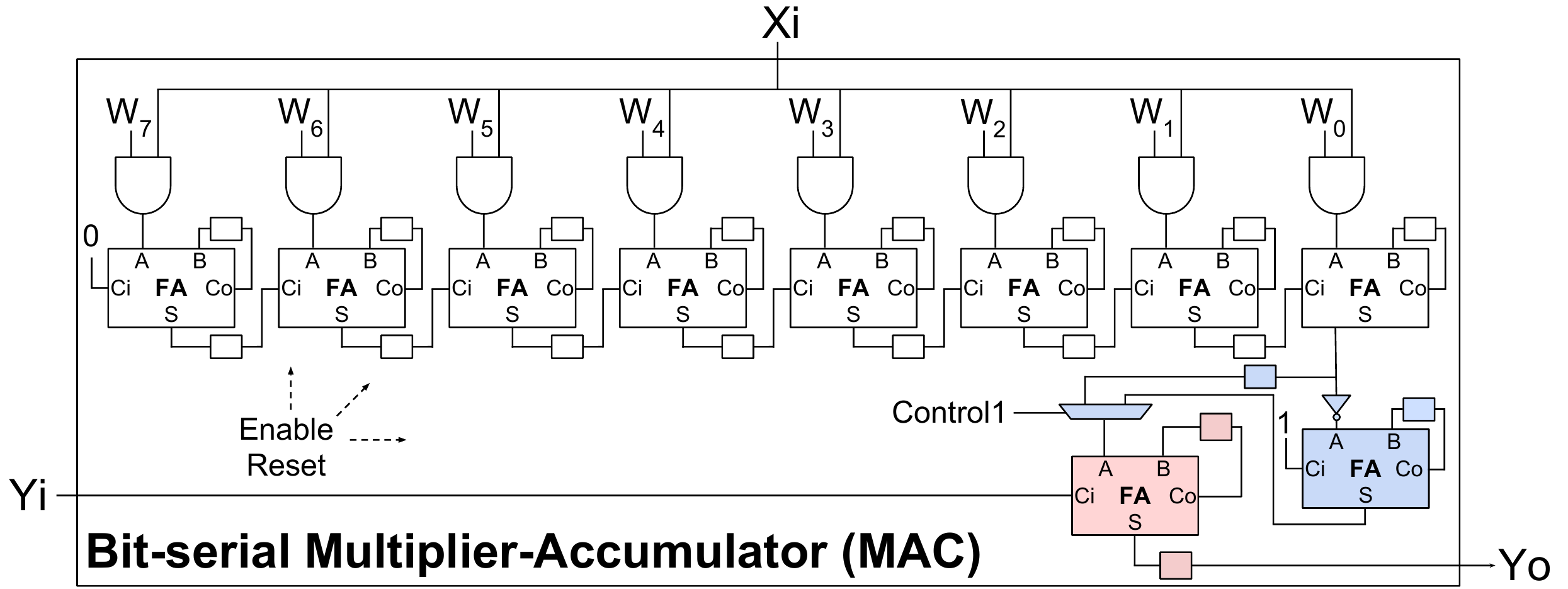}
    \caption{Bit-serial multiplier-accumulator (MAC).}
    \label{fig:bitserial-mac}
\end{figure}

In this section, we describe our bit-serial implementation of systolic arrays for matrix multiplication. Figure~\ref{fig:bitserial-mac} show our proposed bit-serial MAC design which is used across all systolic array implementation for 8-bit input Xi and 8-bit filter weight W. The white logic elements implement the bit-serial multiplication between the input Xi and the absolute value of the filter weight. The blue logic elements negate the product based on the sign of the filter weight. The pink full adder performs bit-serial addition between the product and the input accumulation Yi.

\begin{figure}
\centering
\subfloat{%
    \includegraphics[height=3.25cm]{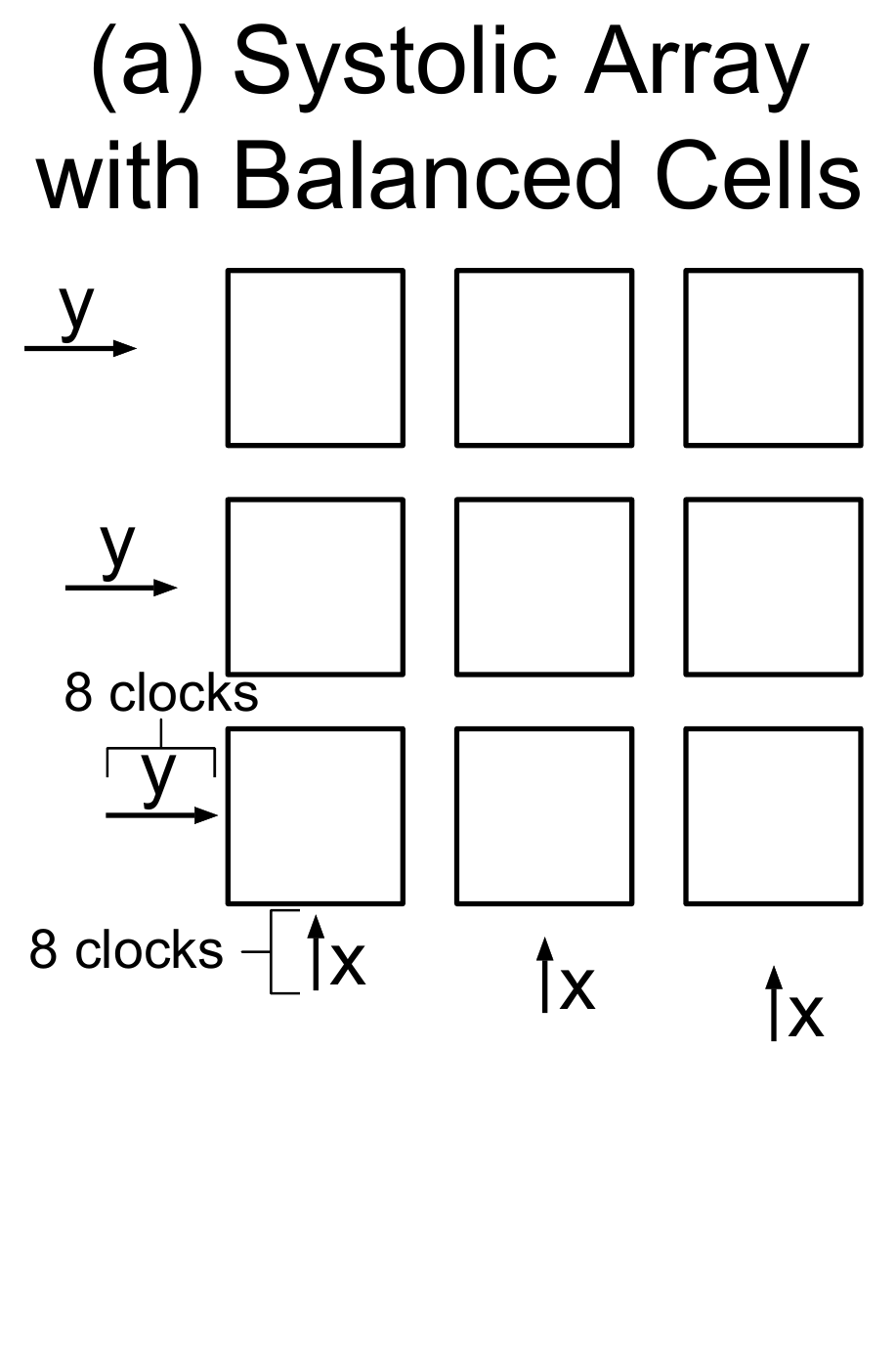}
    \label{fig:array-balanced}}\hfill
\subfloat{%
    \includegraphics[height=3.25cm]{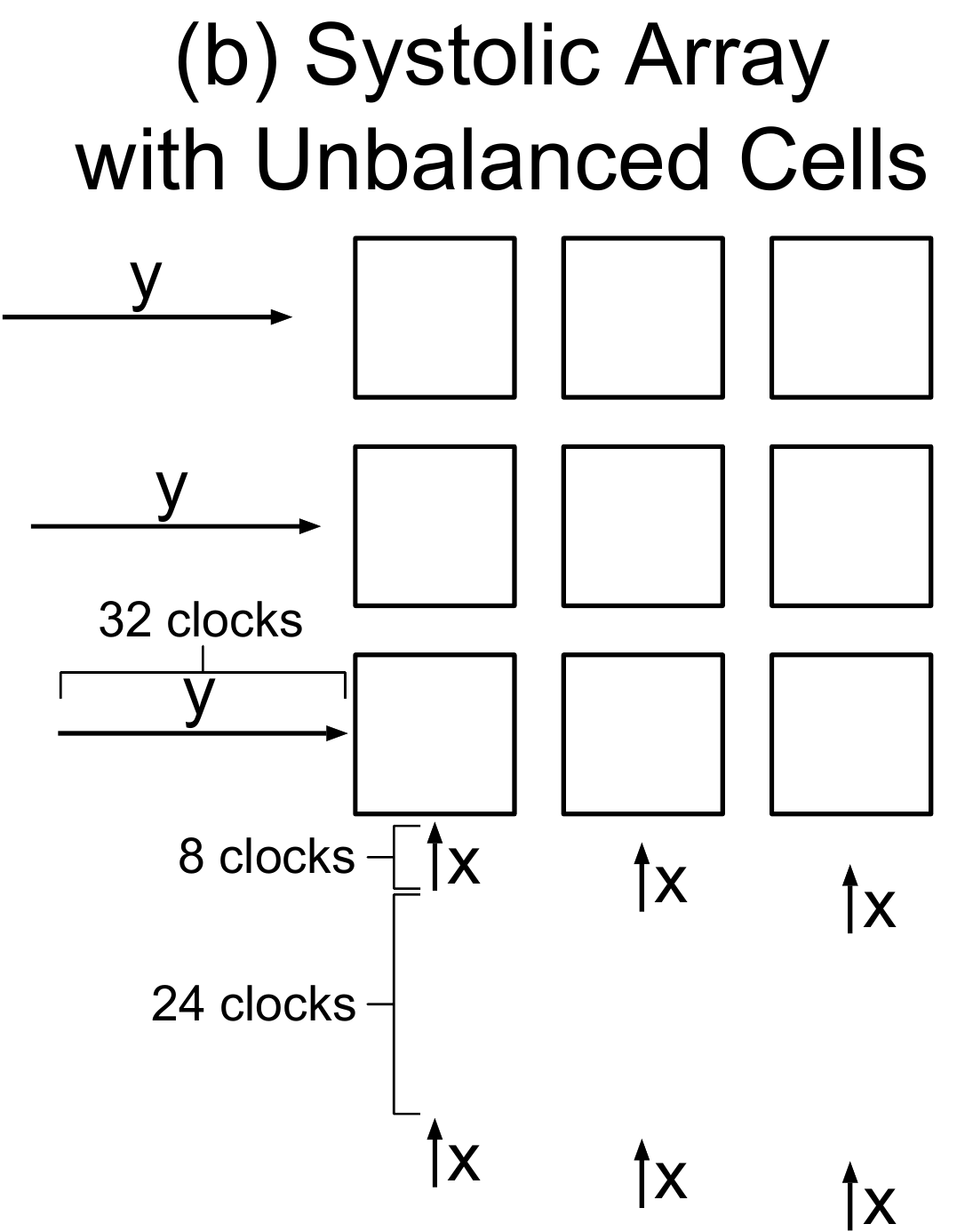}
    \label{fig:array-unbalanced}}\hfill
\subfloat{%
    \includegraphics[height=3.25cm]{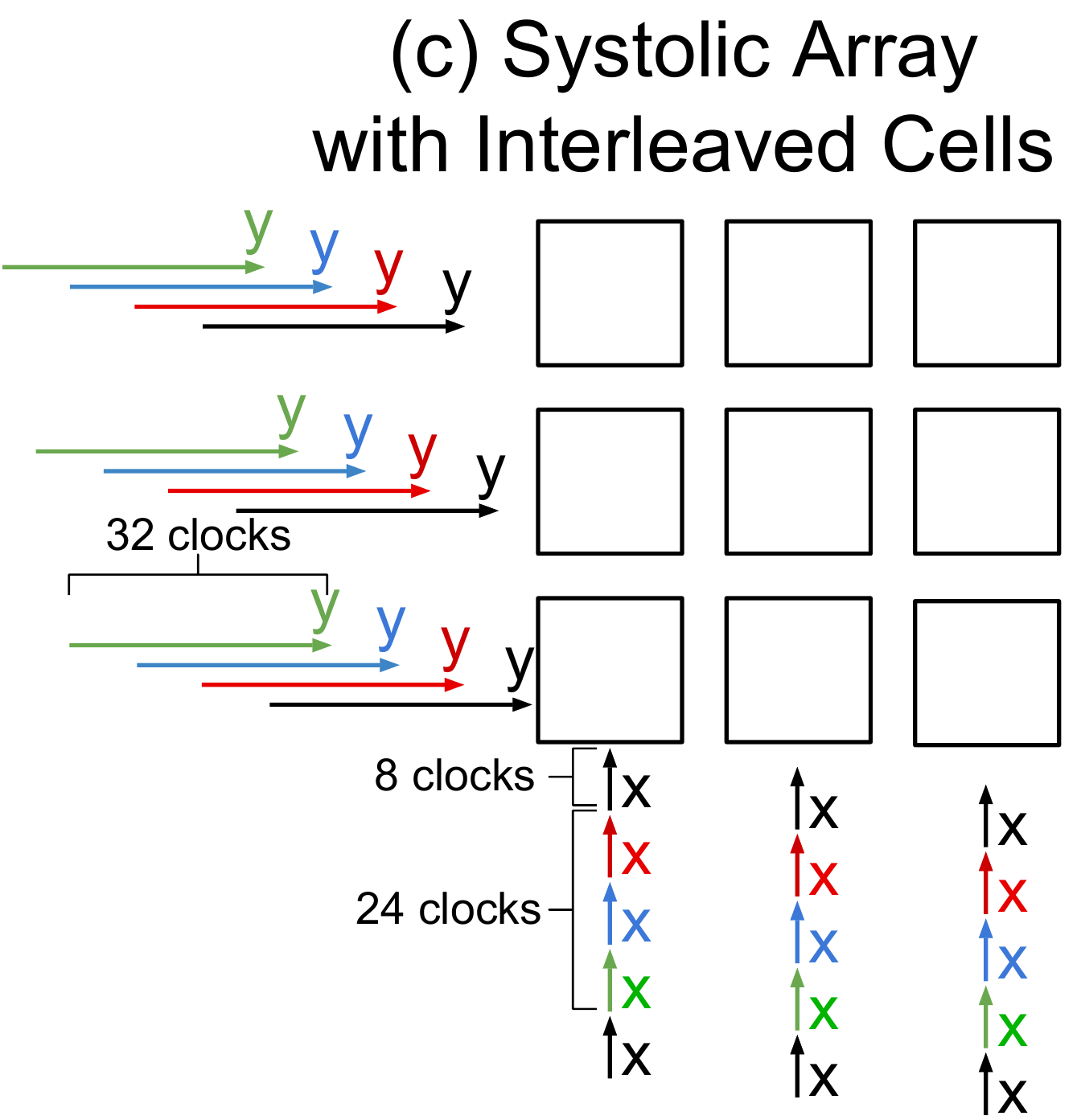}
    \label{fig:array-interleaved}}\hfill
\caption{Systolic arrays under mixed precision settings.}
\end{figure}

We illustrate the scheme with a $3\times3$ bit-serial systolic array for multiplying a $3\times3$ filter matrix and a $3\times$M data matrix, as depicted in Figure~\ref{fig:array-balanced}. We pre-store in the systolic cell (or simply cell) at position ${(i,j)}$ the corresponding filter weight W$_{i,j}$ in the filter matrix. Data arrive from the bottom of the array. Matrix multiplication results come out from the right side of the array.

\begin{figure}
\centering
\subfloat[Balanced Cell]{%
    \includegraphics[height=1.9cm]{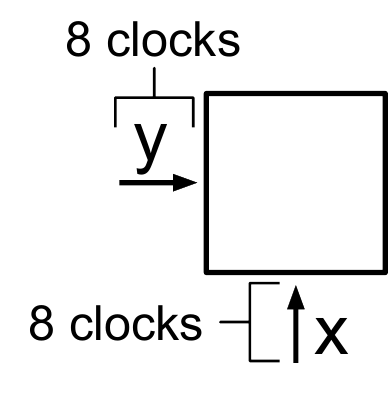}
    \label{fig:cell-balanced}}\hfill
\subfloat[Unbalanced Cell]{%
    \includegraphics[height=1.9cm]{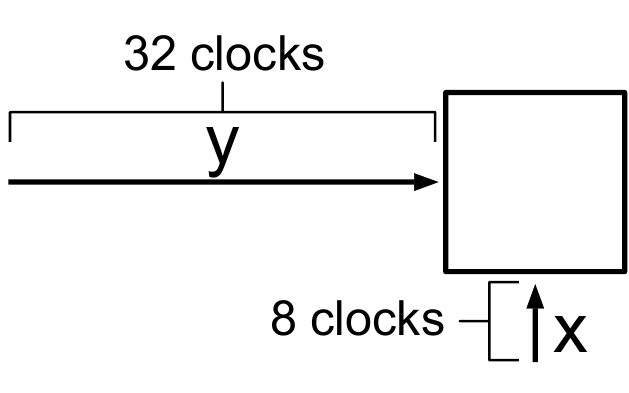}
    \label{fig:cell-unbalanced}}\hfill
\subfloat[Interleaved Cell]{%
    \includegraphics[height=1.9cm]{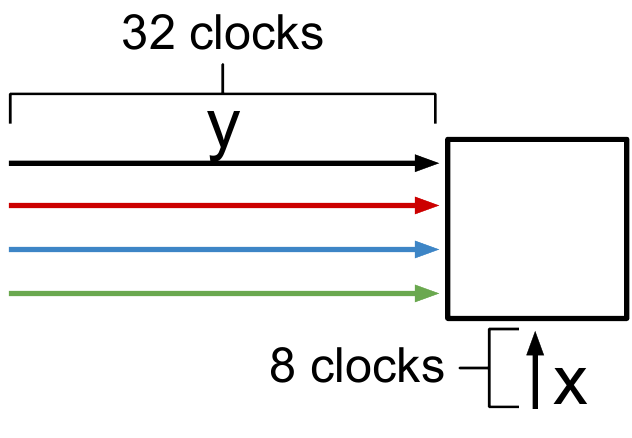}
    \label{fig:cell-interleaved}}
\caption{Systolic cells under different computation settings.}
\end{figure}

\begin{figure}
    \centering
    \includegraphics[width=0.85\columnwidth]{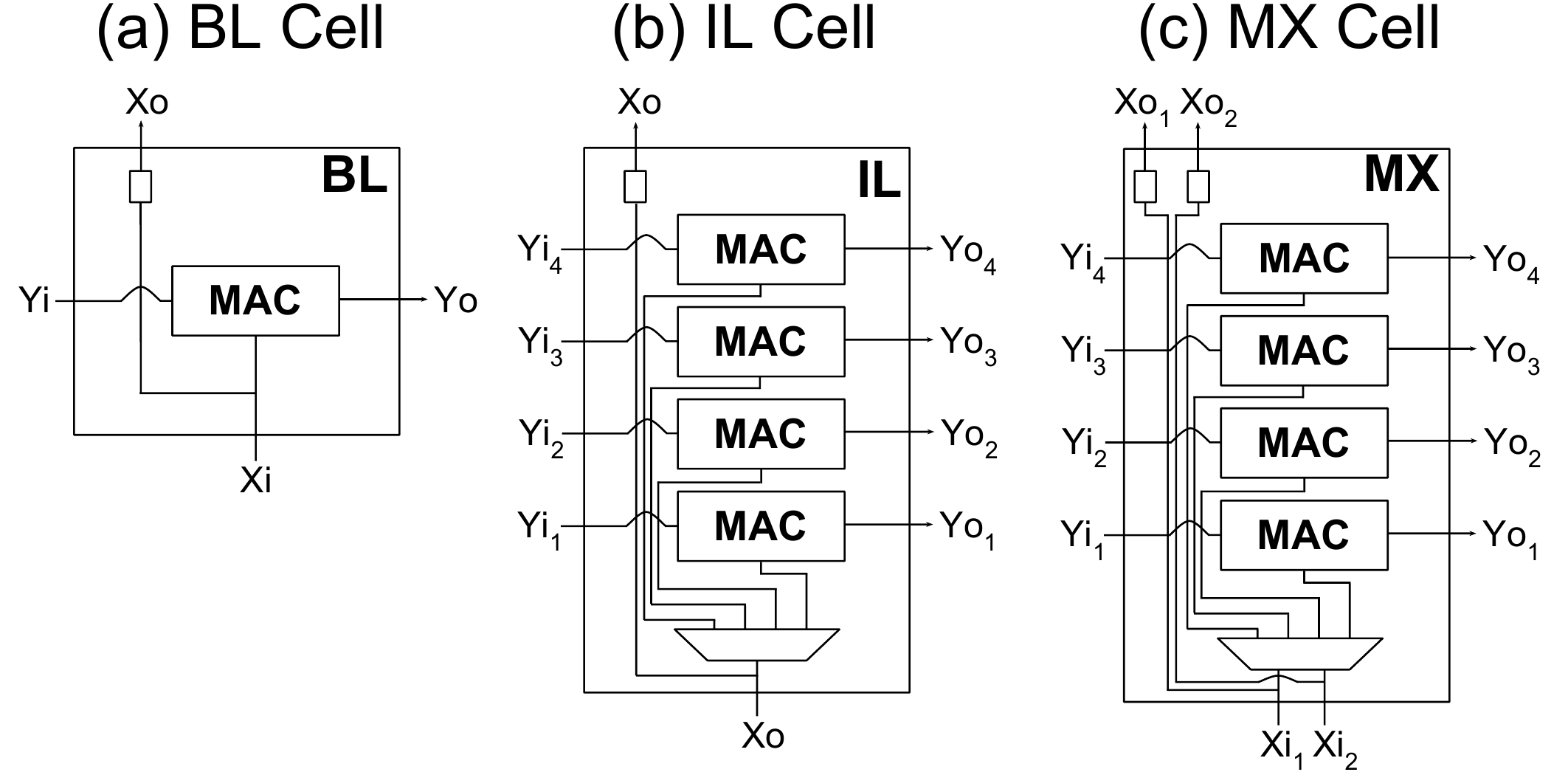}
    \caption{Systolic cell types used for the corresponding systolic array in Figure~\ref{fig:bitserial-systolic-arrays}.}
    \label{fig:bitserial-cells}
\end{figure}

First, consider a simple scenario where each systolic cell has \textit{balanced} I/O and computation time. This is the case when input data, filter weights and accumulation values use words of the same length. Suppose that they are all 8-bit. In this case, under the bit-serial MAC implementation of Figure~\ref{fig:bitserial-mac}, we will have a systolic cell as depicted in Figure~\ref{fig:cell-balanced} or a BL cell in Figure~\ref{fig:bitserial-cells}.  In the corresponding systolic array, as depicted in Figure~\ref{fig:array-balanced}, for data synchronization purposes, neighboring input and accumulation data streams are skewed by one clock to accommodate the communication delay between the cells. However, this simple scenario is not applicable to high-precision accumulation that is necessary for holding the partial result of matrix multiplication~\cite{lin2016fixed}.

To accommodate high-precision accumulation, bit-serial systolic cells will have \textit{longer} computation time than I/O. Suppose that input data and filter weights are 8-bit and accumulation values are 32-bit. In this case, under a bit-serial MAC implementation of Figure~\ref{fig:bitserial-mac} with k = 32, we have the systolic cell as depicted in Figure~\ref{fig:cell-unbalanced}. In the corresponding systolic array, as depicted in Figure~\ref{fig:array-unbalanced}, there is a 24-clock gap between words in each input data stream. The gap allows for the additional computation time required beyond the I/O time.

\begin{figure}
\centering
\subfloat{%
    \includegraphics[height=3.2cm]{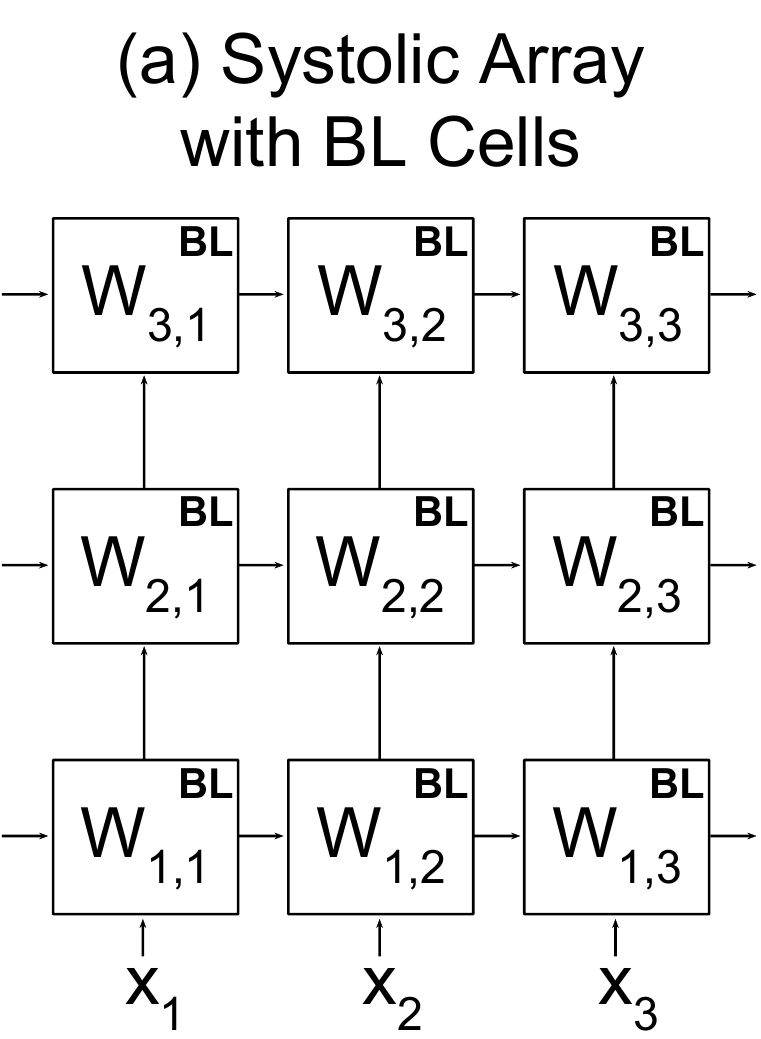}
    \label{fig:bitserial-BL}}\hfill
\subfloat{%
    \includegraphics[height=3.2cm]{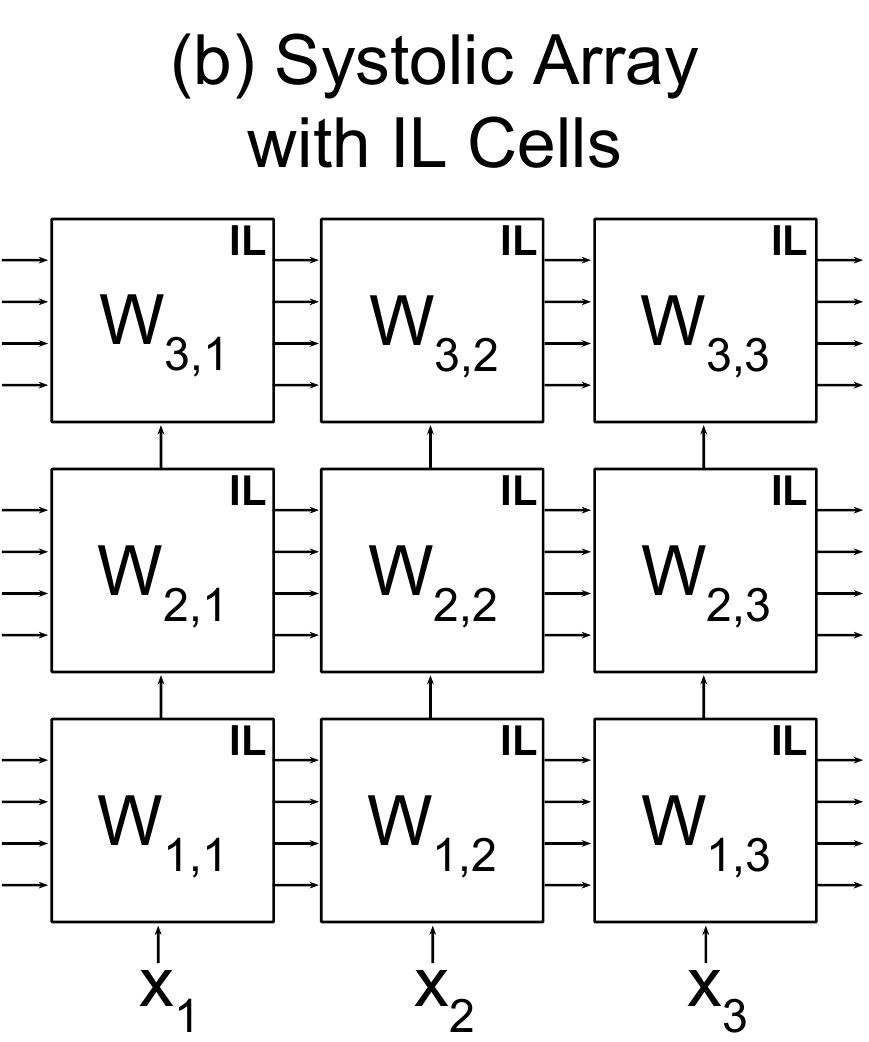}
    \label{fig:bitserial-IL}}\hfill
\subfloat{%
    \includegraphics[height=3.2cm]{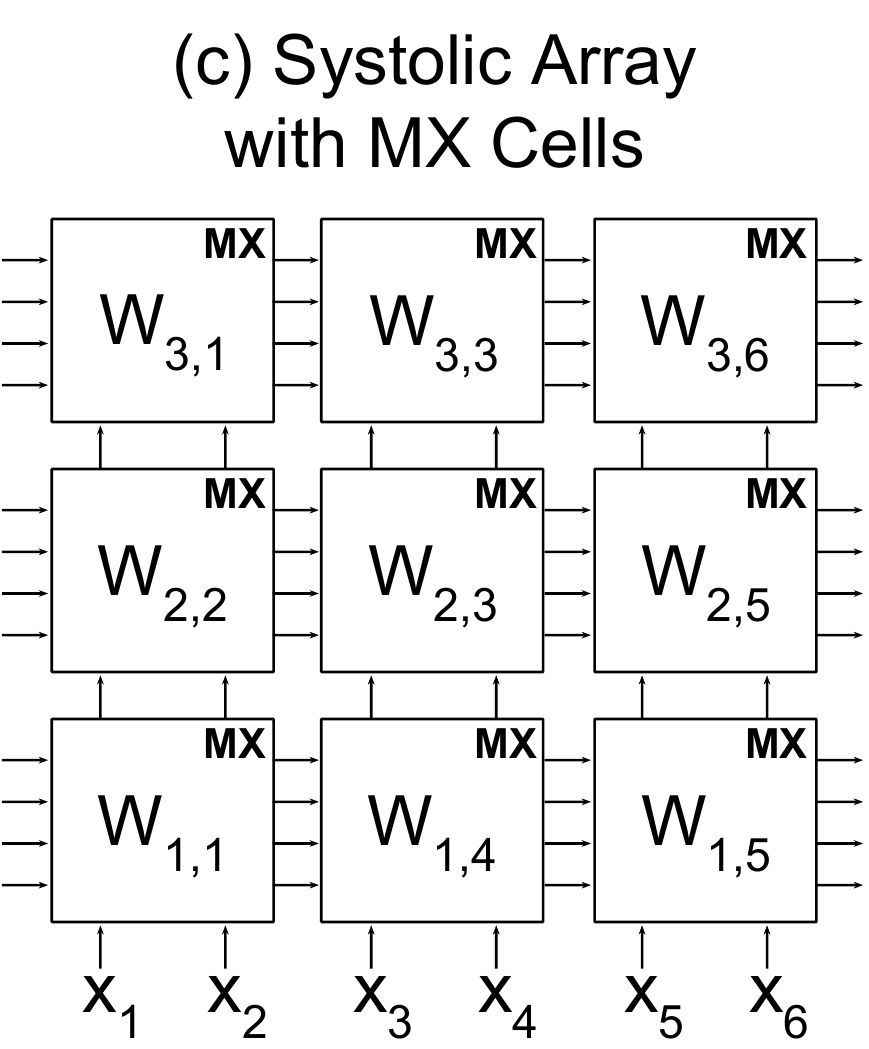}
    \label{fig:bitserial-MX}}\hfill
\caption{Three types of systolic arrays based on the three cell designs in Figure~\ref{fig:bitserial-cells}.}
\label{fig:bitserial-systolic-arrays}
\end{figure}

We can fill in these gaps for each cell by processing \textit{four} independent input data streams simultaneously in an interleaved manner, while expanding the processing power and accumulation data path by 4$\times$, as depicted in Figure~\ref{fig:cell-interleaved} and the IL cell in Figure~\ref{fig:bitserial-cells}. The corresponding systolic array is depicted in Figure~\ref{fig:array-interleaved} with more details in Figure~\ref{fig:bitserial-IL}.

Given the input channel groups determined by the column combining algorithm, we now describe an efficient systolic array implementation which can utilize the combined input channel groups. In Figure~\ref{fig:bitserial-cells}, the multiplexed input (MX) cell, takes in two x inputs, from two input channels, utilizes one of them inside each MAC, and forwards both x to the cell above. Note that while for illustration simplicity this figure shows only two instances of input x$_i$, in our ASIC and FPGA designs we pack up to 8 channels (requiring 8 instances of input x$_i$) into a single cell. This highlights the importance of the bit-serial design, as in the case of 8 inputs, each cell takes in only 8 bits per cycle, as opposed to a bit-parallel design where each cell would require 64 inputs per cycle in the case of 8-bit input.

Figure~\ref{fig:bitserial-MX} shows how a systolic array connects the MX cells. In this example, for the first column, the first and third rows (filters) use input channel 1, denoted by the $\mathsf{W_{1,1}}$ and $\mathsf{W_{3,1}}$ weights stored within the cells, and the second row uses input channel 2, denoted by the $\mathsf{W_{2,2}}$ weight stored in the cell. As shown, these channels indexes are after row permutation (Section~\ref{sec:row-permute}), and are therefore guaranteed to be contiguous.

\subsection{Shift Block}
\label{sec:shift}
\begin{figure}
\centering
    \subfloat{%
    \includegraphics[height=3.6cm]{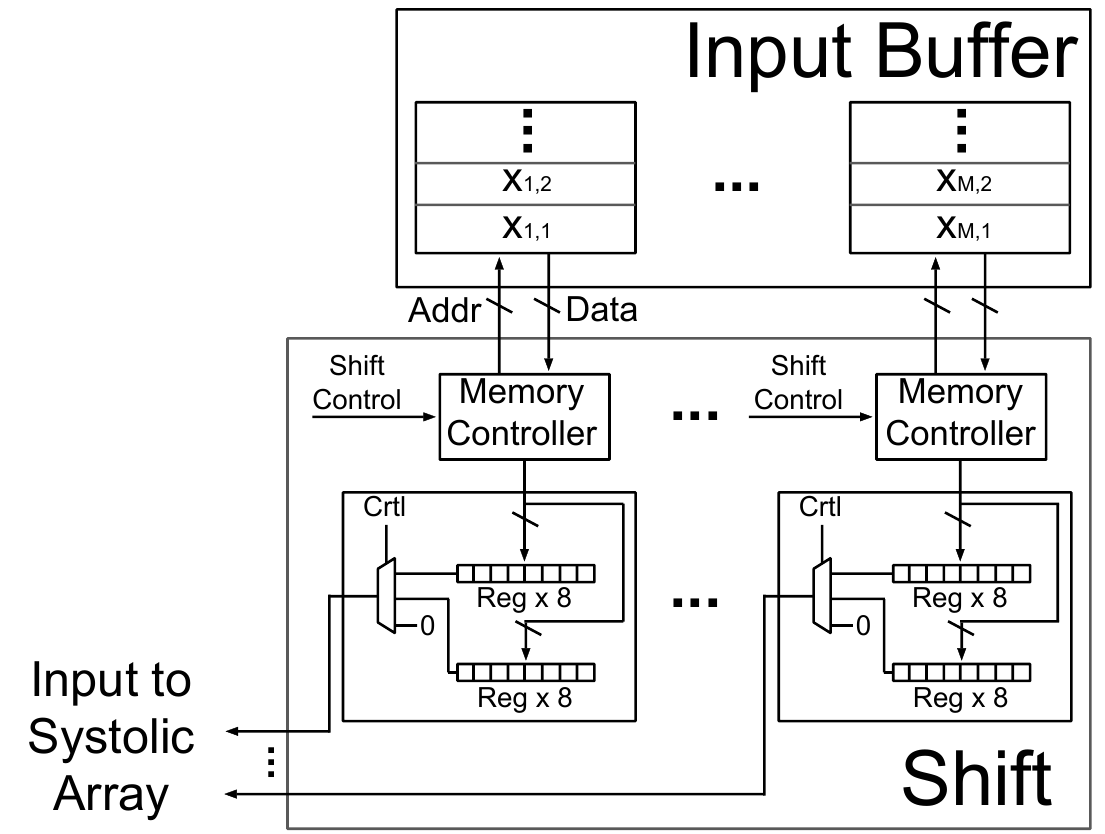}}\hfill
    \subfloat{%
    \includegraphics[height=1.6cm]{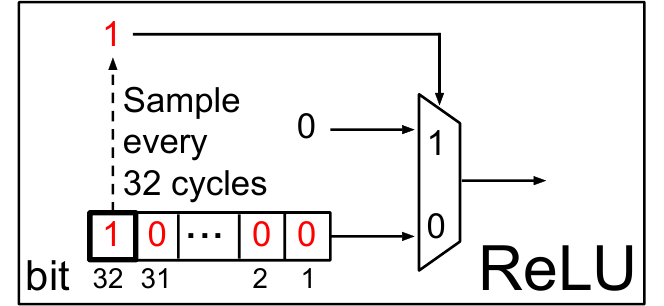}}\hfill
\caption{Shift and ReLU blocks.}
\label{fig:shift-relu}
\end{figure}

Figure~\ref{fig:shift-relu} shows the design for shift operation. Based on the direction of the spatial translation specified by the shift control signal, the memory controller fetches the corresponding 8 bits input maps from the input buffer to the register array, which generates the input to the systolic arrays in a bit-serial fashion. We use double buffering to prefetch the next data tile so that the output time can overlap with the data transfer overhead from the input buffer to register arrays.

\subsection{ReLU and Quantization}
\label{sec:relu}
Figure~\ref{fig:shift-relu} shows the design for ReLU operation. The 32-bit input stream comes in a bit-serial fashion and is stalled in a register array until the last bit arrives. The sign of the integer number represented by the 32-bit input stream is determined by the most significant bit (32nd bit). If the 32nd bit is 1, then the multiplexer outputs a 32-bit stream of 0, otherwise the multiplexer simply outputs the input stream. The output from the ReLU block is then re-quantized and saved in the output buffer. This output can then be transferred to the input buffer to be used as input for the following layer.

\section{Performance Analysis for the Column Combining Algorithm}
\label{sec:combine-analysis}

We analyze our column combining approach described in Section~\ref{sec:column-combine} on two datasets MNIST~\cite{lecun1998mnist} (28$\times$28 greyscale images of handwritten digits) and CIFAR-10~\cite{krizhevsky2014cifar} (32$\times$32 RGB images of objects). We evaluate the approach on three well studied datasets: Lenet-5~\cite{lecun1998mnist} on MNIST and VGG-16~\cite{simonyan2014very} and ResNet-20~\cite{he2016deep} on CIFAR-10. Each convolution layer in all networks is replaced by shift followed by pointwise convolution~(Shift Convolution in Figure~\ref{fig:convolution}) to fit our systolic array system design covered in Section~\ref{sec:column-combine-design}. All networks are trained using Stochastic Gradient Descent (SGD) with an initial learning rate $\eta$ of 0.05 for Lenet-5 and 0.2 for VGG-16 and ResNet-20.  A Nesterov momentum of 0.9~\cite{ruder2016overview} is used for all networks. A cosine shape learning rate schedule~\cite{loshchilov2016sgdr} is used to decay the learning rate over each iteration of Algorithm~\ref{alg:iterative-training}, ending at 20\% of the initial initial learning rate $\eta$. After the target number of weights has been reached, 100 additional epochs of training is performed, while decaying the learning rate to $0$. Unless stated otherwise, $\beta$ is set to $20\%$ for all networks.

\subsection{Iterative Training with Column Combining}
\begin{figure*}
\centering
    \subfloat[Iterative Training with Column Combining]{%
    \includegraphics[width=0.32\textwidth]{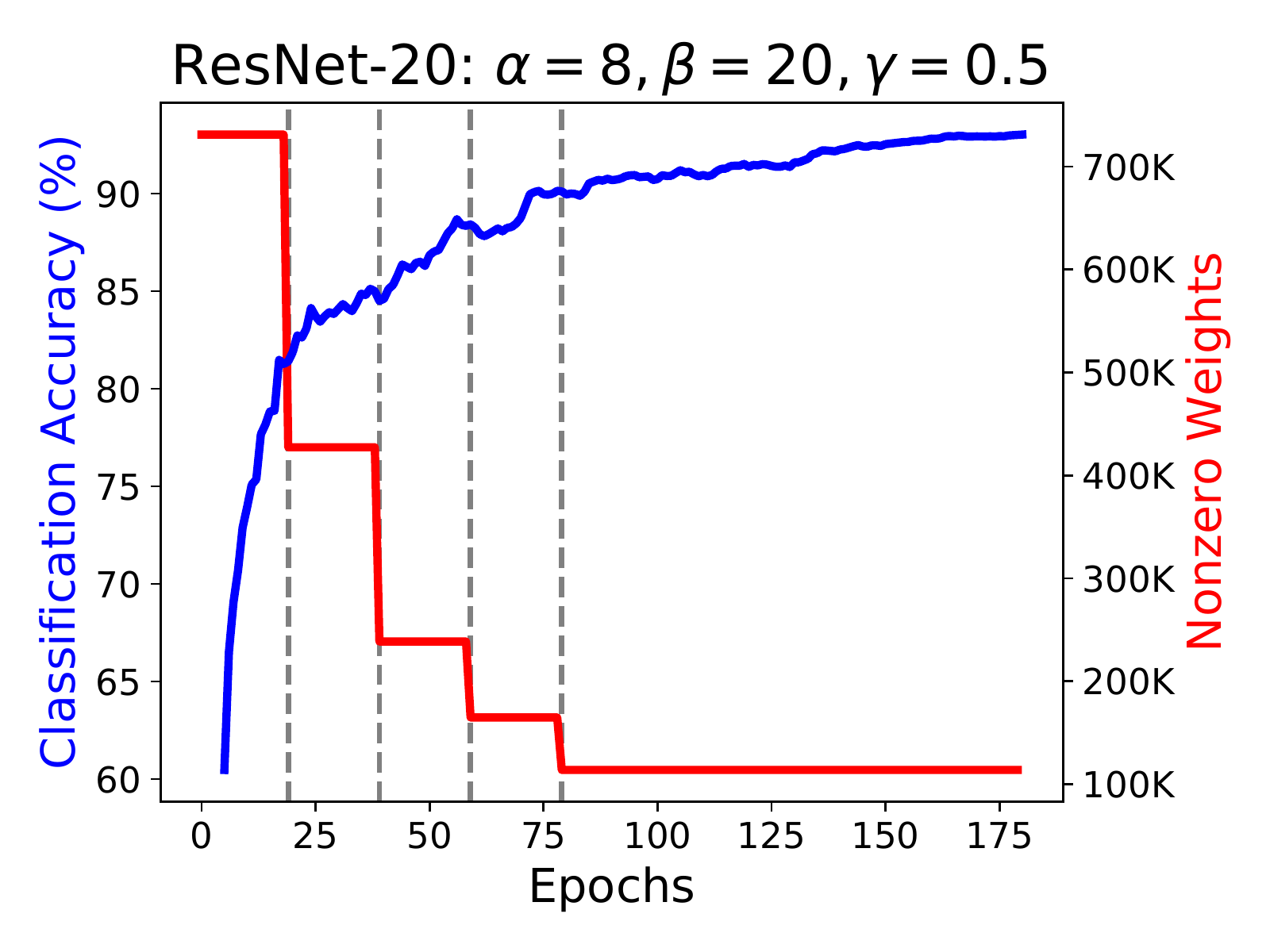}
    \label{fig:epoch-acc-weight}}\hfill
    \subfloat[Impact of $\alpha$]{%
    \includegraphics[width=0.32\textwidth]{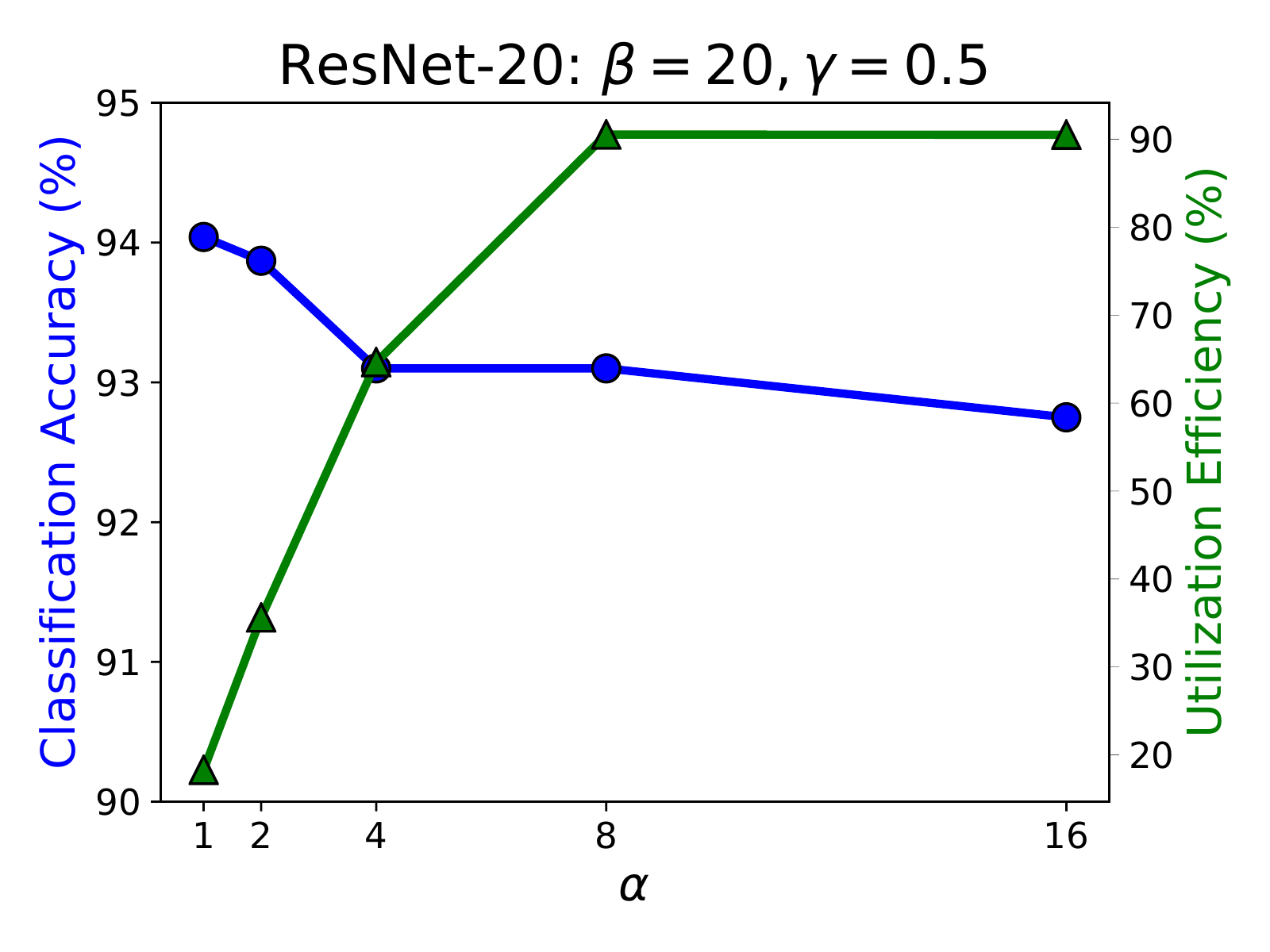}
    \label{fig:alpha}}\hfill
    \subfloat[Impact of $\gamma$]{%
    \includegraphics[width=0.32\textwidth]{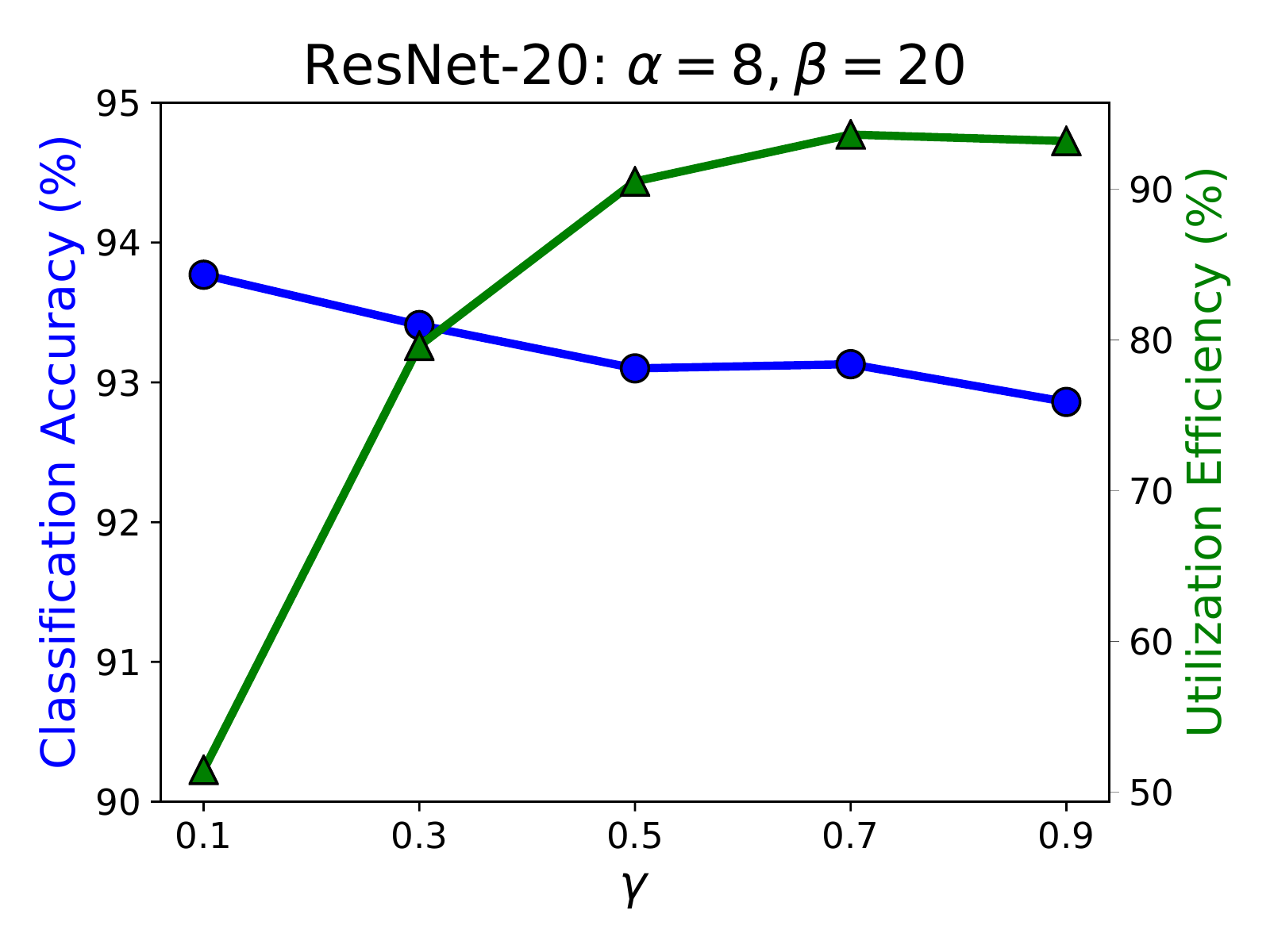}
    \label{fig:gamma}}\hfill
\caption{(a) Classification accuracy and number of nonzero weights over training epochs (grey vertical lines denote pruning). (b) Increasing $\alpha$ allows for more columns to be added to a single combined column. (c) Increasing $\gamma$ greatly improves utilization efficiency while minimally impacting classification accuracy.}
\end{figure*}

Training a network with column combining occurs over a series of iterations (Algorithm~\ref{alg:iterative-training}), where, at each iteration, weights are pruned to decrease the model size and increase the utilization efficiency when deployed in the systolic array. After each pruning stage, retraining is performed to recover the loss in classification accuracy due to the pruned weights. Figure~\ref{fig:epoch-acc-weight} shows the classification accuracy and number of nonzeros weights for the ResNet-20 model over each training epoch. The dashed vertical lines denote the beginning of an iteration of Algorithm~\ref{alg:iterative-training}, where initial pruning (using $\beta$) and column-combine pruning (using $\alpha$ and $\gamma$) are performed. At each epoch, the number of weights in the model is shown by the red line. The first iteration of pruning decreases the model size most substantially (from 740K to 440K nonzero weights), and subsequent pruning stages decreases the model size by smaller amounts due to the reduced $\beta$ value. When the target number of nonzeros weights is reached, 125K in this instance, a final 100 epochs of training is performed, which improves the classification accuracy by an additional 5\%. 

\subsection{Impact of Number of Columns per Group}
The number of columns allowed to be added to a group during column grouping (Algorithm~\ref{alg:combine}) is determined by the $\alpha$ parameter. Figure~\ref{fig:alpha} shows the classification accuracy and utilization efficiency for 5 ResNet-20 models for the CIFAR-10 dataset trained using Algorithm~\ref{alg:iterative-training} with $\beta=20$ and $\gamma=0.5$ while varying $\alpha$ from 1 to 16. At $\alpha=1$, no column combining or column-combine pruning is performed, as only a single column is allowed per group. This network is equivalent to a standard systolic array operating on sparse filter matrices and achieves a utilization efficiency of under $20\%$. Note that, for this analysis, utilization efficiency and packing efficiency are interchangeable. As $\alpha$ is increased, the utilization efficiency improves up to $90\%$ at $\alpha=8$, with the classification accuracy dropping by approximately 1\% due to column-combine pruning. For $\alpha=16$, there is no improvement in utilization efficiency, as columns cannot be further combined due to the higher degree of conflicts between the remaining nonzero weights. 

\subsection{Impact of the Limited-Conflict Condition}
The limited-conflict condition, as described in Section~\ref{sec:terminologies}, allows for $\gamma$ conflicting entries per row on average between columns within a group. All but the largest magnitude weight among conflicting weights are pruned during column-combine pruning (Algorithm~\ref{alg:group-prune}). Figure~\ref{fig:gamma} shows how classification accuracy and utilization efficiency vary as a function of $\gamma$ for 5 ResNet-20 networks trained on the CIFAR-10 dataset. Larger values of $\gamma$ allow for more conflicts between the columns in a group and therefore prune more weights, possibly with relatively large magnitudes, in order to achieve higher utilization efficiency across all layers in the CNN.  This dramatically increases the utilization efficiency from 52\% ($\gamma=0.1$) to 93\% ($\gamma=0.5$). As discussed in the previous subsection, column-combine pruning has a small impact on classification accuracy (around 1\%) since retraining is performed after each round of pruning in order to allow the remaining weights to adjust to the loss of the pruned weights.

\subsection{Dramatic Tiling Reduction in Partitioned Matrix Multiplication with Column Combining}
\label{sec:tiling}
\begin{figure}
\centering
    \subfloat{%
    \includegraphics[height=3.9cm]{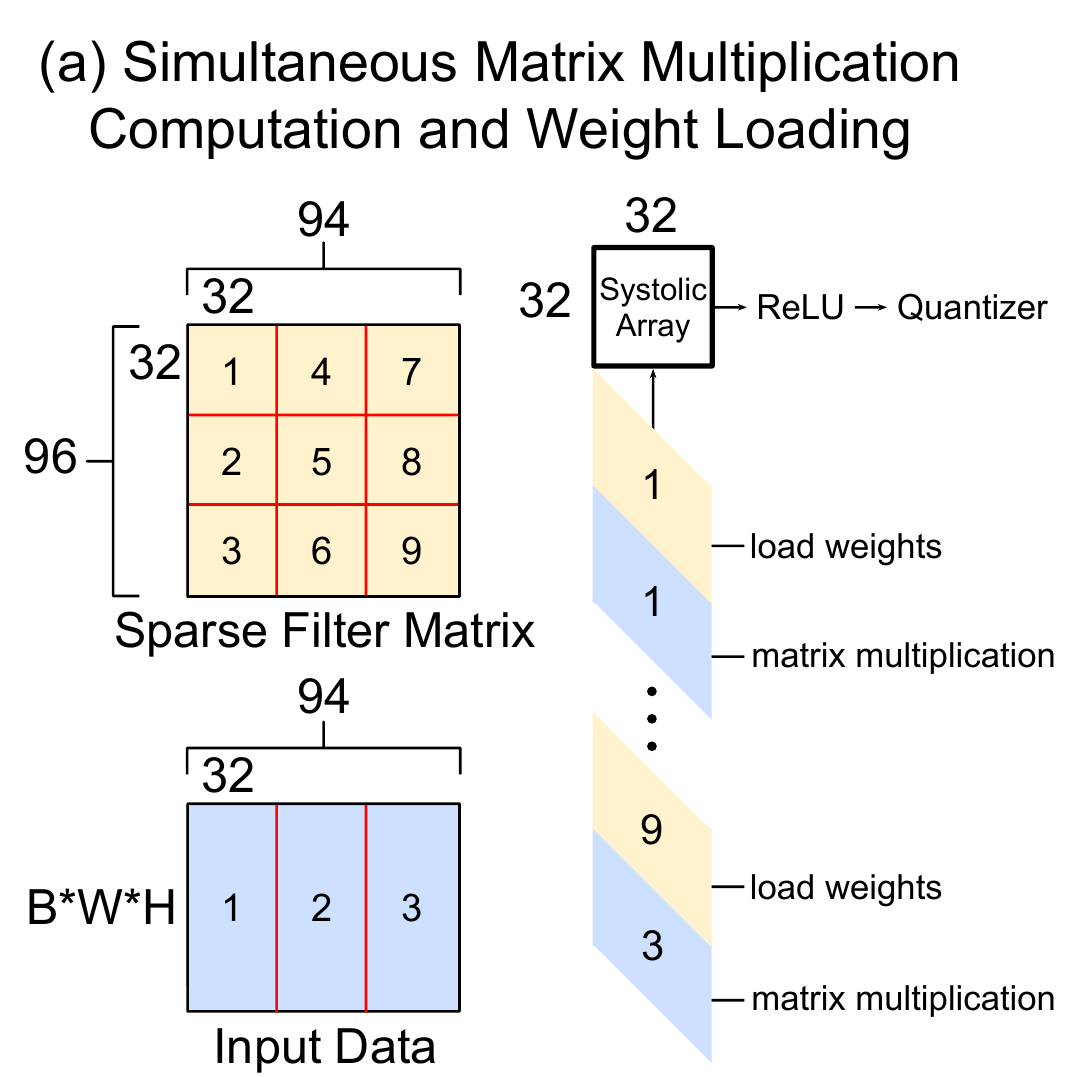}
    \label{fig:pipeline-loads}}\hfill
    \subfloat{%
    \includegraphics[height=3.9cm]{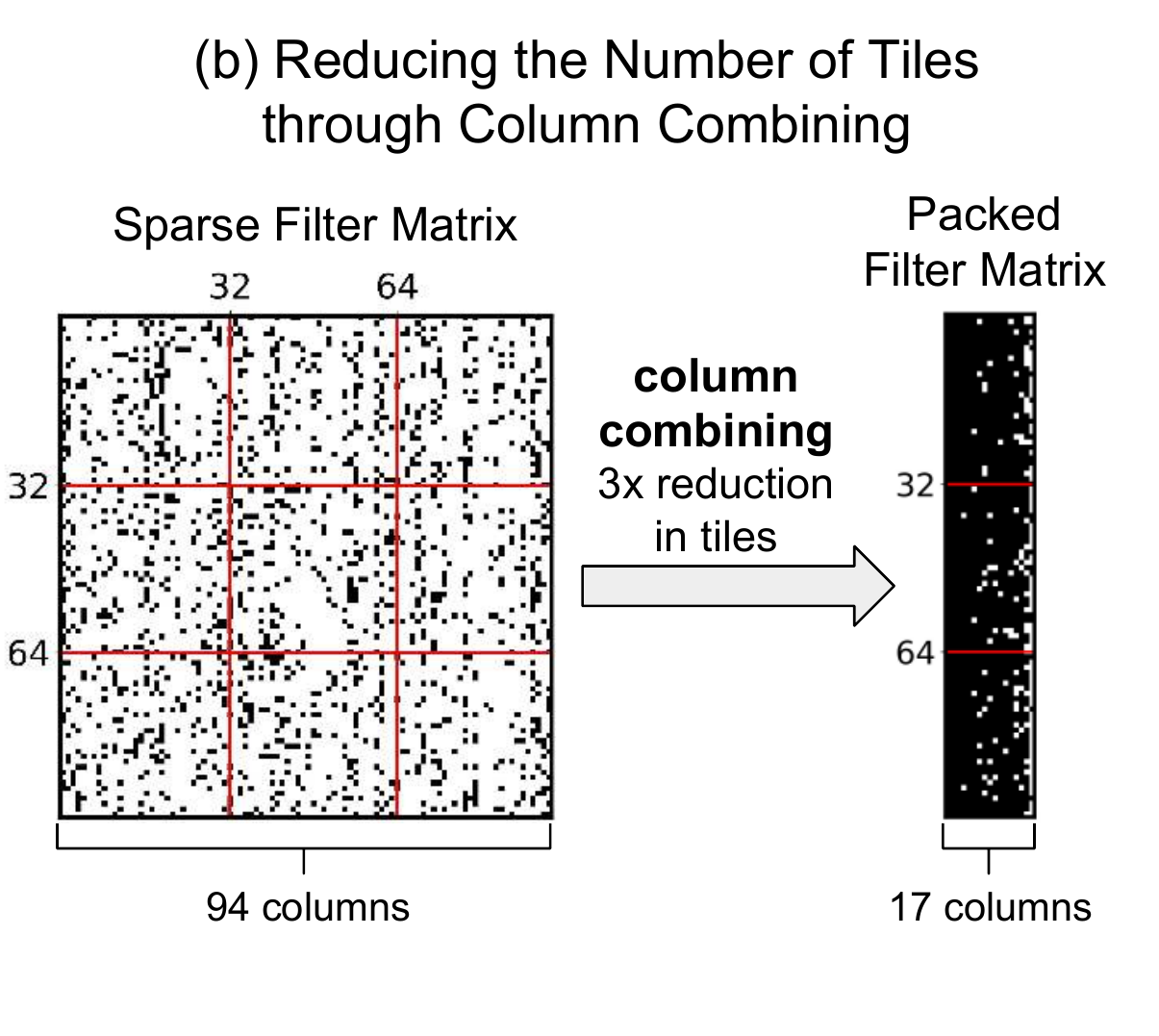}
    \label{fig:tile-reduction}}\hfill
\caption{(a) Partitioned matrix multiplication with systolic array alternating between weight loads and matrix multiplication. (b) Column Combining reduces the number of tiles.}
\end{figure}

When a systolic array is smaller than the weights of a convolutional layer, matrix multiplication can be performed in multiple passes, where each pass executes matrix multiplication between a submatrix of the layer weights and the corresponding input data. Figure~\ref{fig:pipeline-loads} shows how this partitioning process is performed on a sparse filter matrix of (96 rows by 94 columns), which is larger than the systolic array (32 rows by 32 columns). The filter matrix is partitioned into 9 tiles, each with a maximum size of 32 by 32, and the input data is tiled in a similar manner along the columns, but not along the rows (batch size $\times$ image width $\times$ image height). 

The full matrix multiplication is performed by alternating between weight loads and matrix multiplications for each of the submatrices (tiles). The filter matrix and input data enter the systolic array as depicted in a skewed fashion in order to maintain synchronization within the systolic array. Note that every systolic cell is busy all the time, either doing the matrix multiplication computation or loading the weights for the next tile. ReLU and quantization are performed on the output of the systolic array after the final tile for a set of rows in the filter matrix. (Note that in Section~\ref{sec:eval}, we evaluate settings where the CNN must be partitioned in tiles, as shown in Figure~\ref{fig:pipeline-loads} and also settings where the each layer can fit entirely into a systolic array which does not require partitioning.)

We have used a ResNet-20 model in the performance study for our proposed column combining scheme. For illustration purposes, consider here the third layer of model. Figure~\ref{fig:tile-reduction} shows a sparse filter matrix and a corresponding packed filter matrix after column combining, which is stored in the systolic array with MX cells (described in Section~\ref{sec:column-combine-design}). As in Figure~\ref{fig:pipeline-loads}, the sparse filter matrix has 96 rows and 95 columns, with only 16\% of the weights being nonzeros. For a 32$\times$32 systolic array, this sparse filter matrix is partitioned into 9 tiles (denoted by the red lines) in order to perform the full matrix multiplication. The packed filter matrix is the output after column combining, which has arranged the 94 columns of the sparse filter matrix into 17 groups. Each group is a single column in the packed filter matrix format, which can then be loaded into the systolic array. This packed format has 89\% nonzeros and requires only 3 tiles to perform matrix multiplication (a 3$\times$ reduction).

\begin{figure}
\centering
    \subfloat[Number of Tiles in Systolic Array]{%
    \includegraphics[width=0.48\columnwidth]{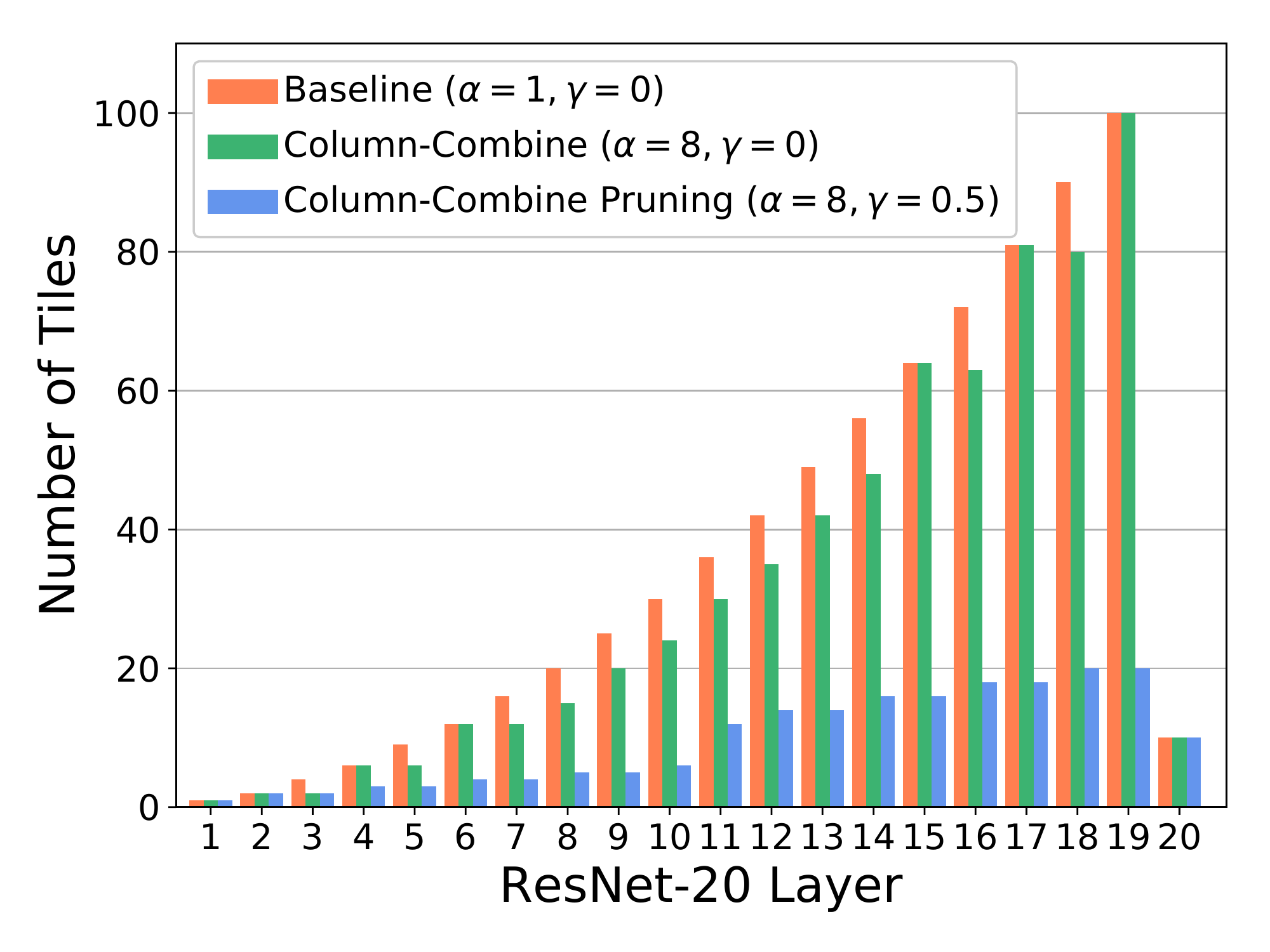}
    \label{fig:tiles-analysis}}\hfill
    \subfloat[Training with Limited Data]{%
    \includegraphics[width=0.48\columnwidth]{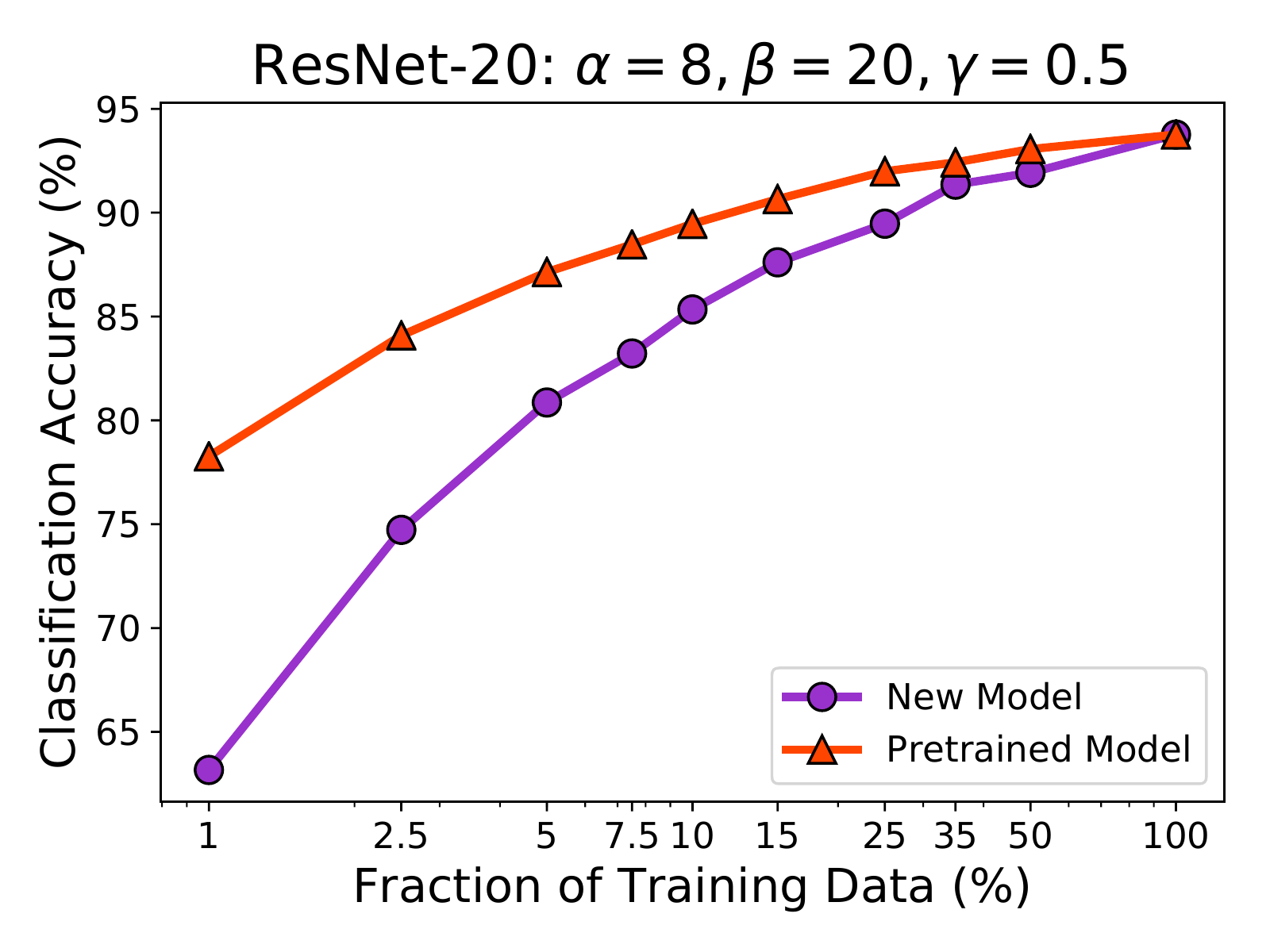}
    \label{fig:limited}}\hfill
\caption{(a) Number of tiles for a 32$\times$32 systolic array. (b) Comparing training a new model to training a pretrained model with column combining on limited datasets.}
\end{figure}

Figure~\ref{fig:tiles-analysis} shows the number of tiles required to perform matrix multiplication with a 32$\times$32 systolic array for each layer in ResNet-20 models trained using Algorithm~\ref{alg:iterative-training} under three different parameter settings. All three settings set $\beta = 20$. The baseline setting trains the CNN with standard pruning but without column combining or column-combine pruning ($\alpha=1, \gamma=0$). The column-combine setting uses the same CNN trained in the baseline setting, but allows for column combining without column-combine pruning ($\alpha=8, \gamma=0$). Finally, the column-combine pruning setting trains the CNN with column combining and performs column-combine pruning to remove conflicting entries ($\alpha=8, \gamma=0.5$). The column-combine setting only reduces the number of tiles over the baseline setting by 10\% at most. By comparison, the column-combine pruning setting reduces the number of tiles by a substantial margin across all layers and achieves at 5$\times$ reduction in the largest layer (layer 19). Generally, this shows that it is difficult to effectively combine sparse columns, as a single conflict in any row for a potential group will make the combination invalid. By adding a modest amount of column-combine pruning (\eg~$\gamma=0.5$) the combining algorithm is able to substantially improve the utilization efficiency and decrease the number of tiles. 

\section{Column Combining with Limited Datasets}
\label{sec:limited}
In many real world scenarios, customers may provide pretrained models to vendors to be deployed on their device~(\eg~a mobile device). In these settings, a customer may not wish to divulge datasets used to train the model to the vendor for a number of reasons, such as the dataset containing sensitive private information or being a competitive advantage. In this scenario, model pruning is difficult, as pruning weights without retraining leads to a significant degradation in classification accuracy. 

We propose that these data privacy concerns can be mostly mitigated by providing only a subset of the original dataset to perform the proposed column combining iterative training process. Figure~\ref{fig:limited} compares the effects of column combining on a pretrained dense ResNet-20 model, trained on the full CIFAR-10 train dataset, to a new network, such as depicted in Figure~\ref{fig:epoch-acc-weight}, over different fractions of training data. The largest difference in performance between the two approaches is when only 1\% of the full training data is used (a 15\% difference in classification accuracy), as the weights in the pretrained model are already initialized to reasonable values. At 15\% of the full training data, the pretrained model can achieve over 90\% classification accuracy. This shows that a small amount of training data can be sufficient to perform column combining while still maintaining a relatively high classification accuracy. By comparison, training a new model requires 35\% of the training dataset to achieve an over 90\% classification accuracy. Our model pruning and retraining method can be view as part of a larger area of research shared by teacher-student networks~\cite{romero2014fitnets,tarvainen2017mean} and curriculum learning~\cite{bengio2009curriculum}. 

\section{Hardware Implementation Experiments and Performance Evaluation}
\label{sec:eval}
In this section, we evaluate performance of our column combining system described in Section~\ref{sec:column-combine-design} based on design experiments in both ASIC and FPGA. Throughout, we compare designs in terms of performance metrics concerning accuracy, throughput, area efficiency and energy efficiency. Additionally, we pay attention to performance for single or a small number of input samples, \eg~the end-to-end latency, or energy requirement, in processing a single input sample such as a 28x28 grey-scale image over LeNet 5. As stated earlier in Section~\ref{sec:cross-layer-pipeline}, in realtime scenarios, single sample latency is a more important metric than throughput, as an input sample must be processed immediately for early prediction. 
 
Section~\ref{sec:eval-asic} describes our ASIC implementation and compares it against a baseline systolic array without column combining on three different CNNs (Lenet-5, VGG-16, and ResNet-20). In addition, we compare our ASIC design with other state-of-the-art ASIC accelerators for LeNet-5. 
In Section~\ref{sec:optimality-in-energy}, we provide a mathematical analysis on optimality in energy efficiency. We argue that for CNNs such as LeNet-5 which incurs a relatively small I/O energy compared to MAC operations, packing these CNNs with column combining leads to systolic array designs which are near optimal in energy efficiency. 
Section~\ref{sec:eval-fpga} compares our FPGA implementation with other FPGA CNN accelerators on CIFAR-10. 
Section~\ref{sec:pipeline-results} compares single-sample latency of our ASIC implementations with and without cross-layer pipelining. 

\subsection{ASIC Implementation and Evaluation}
\label{sec:eval-asic}
We synthesize our ASIC design using the Synopsys Design Compiler~\cite{synopsysdesigncompiler} with 45nm NanGate Open Cell Library~\cite{nangatelib} and CACTI 7.0~\cite{cacti}. We estimate the hardware performance of static random-access-memory (SRAM) with CACTI 7.0 and synthesize the remaining components of the design including Systolic Arrays with MX cells (Section~\ref{sec:bitserial-array}), Shift (Section~\ref{sec:shift}), ReLU and Quantization (Section~\ref{sec:relu}) using the Synopsys Design Compiler. 

We analyze our ASIC implementation across two scenarios. First, in Section~\ref{sec:asic-tiling}, we compare the bit-serial systolic array without column combining (Figure~\ref{fig:bitserial-IL}) to our bit-serial design with column combining (Figure~\ref{fig:bitserial-MX}), where a single systolic array is used to process all CNN layers with tiling as presented in Section~\ref{sec:tiling}. Then, in Section~\ref{sec:asic-single}, we compare our column combining ASIC implementation against prior ASIC implementations of LeNet-5. In the second scenario, we can fit each layer entirely into a systolic array and therefore do not require tiling.

\subsubsection{Systolic Array Comparison using Tiling}
\label{sec:asic-tiling}
To analyze our ASIC implementation of column combining, we implement the three networks discussed in Section~\ref{sec:combine-analysis} (LeNet-5, VGG-16 and ResNet-20) using a single systolic array of size 32$\times$32 and perform partitioned matrix multiplication as shown in Figure~\ref{fig:pipeline-loads}. For this scenario, 32-bit accumulation is used for all networks. We report energy consumption for processing one input sample for each CNN across the three column combining algorithm parameter settings presented in Section~\ref{sec:tiling}. The baseline setting uses standard pruning without column combining ($\alpha=1, \gamma=0$), the column-combine setting allows for column combining without column-combine pruning ($\alpha=8, \gamma=0$), and the column-combine pruning setting performs column-combine pruning to improve utilization efficiency by removing conflicting entries ($\alpha=8, \gamma=0.5$). 

\begin{figure}
\centering
\includegraphics[width=\columnwidth]{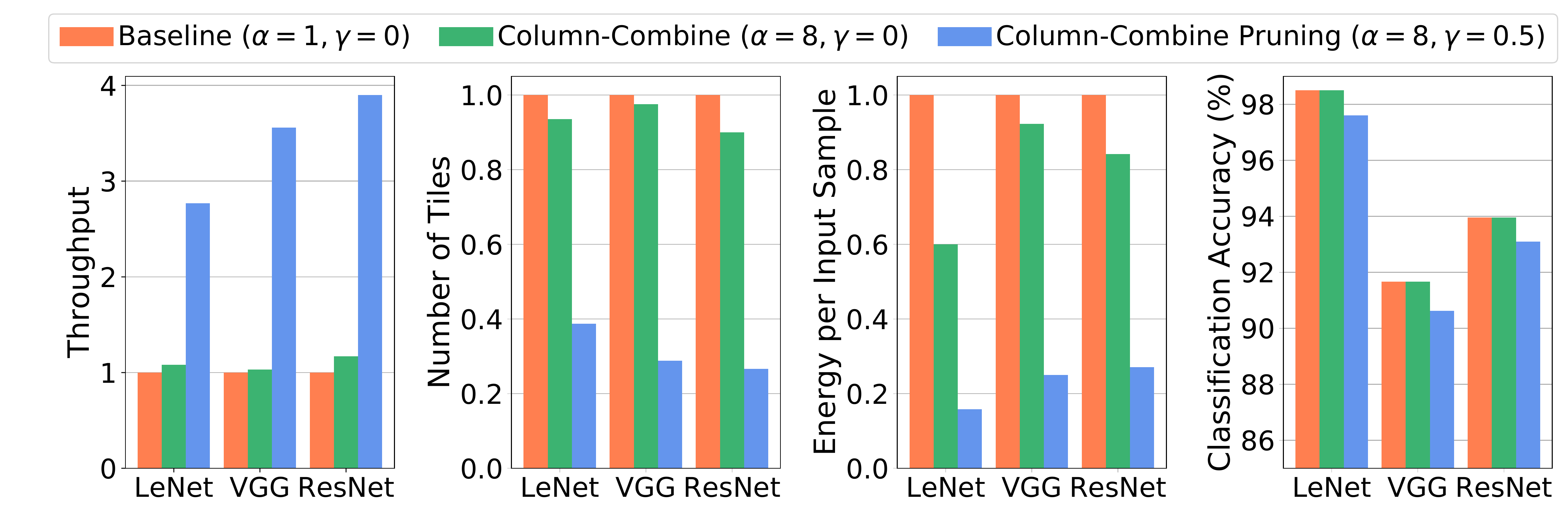}
\caption{Performance of baseline and column combining ASIC implementations using tiling as in Section~\ref{sec:tiling}.}
\label{fig:asic-all}
\end{figure}

Figure~\ref{fig:asic-all} depicts the throughput, number of tiles required to perform matrix multiplication across all layers, energy consumption per input sample, and classification accuracy for each CNN across the three parameter settings. For all the three CNN structures, the column-combine pruning setting greatly reduces the energy consumption and number of tiles by 4$\times$ to 6$\times$ over the other two settings. Furthermore, the column-combine pruning setting has $3\times$ to $4\times$ greater throughput compared to the other settings across all networks.

\subsubsection{Comparison Against Prior Designs on LeNet-5}
\label{sec:asic-single}

We compares our ASIC implementation of LeNet-5, trained on MNIST, to prior state-of-the-art CNN accelerator designs. For this scenario, we use 16-bit accumulations for the systolic array, as all layers are small in terms of filter sizes and therefore do not require 32-bit accumulations. With 16-bit accumulations, a single MAC operation will take half amount of cycles compared with 32-bit accumulations. All other designs use LeNet-5 (except for SpiNNaker~\cite{khan2008spinnaker} which uses a Deep Belief Network and TrueNorth~\cite{akopyan2015truenorth} which uses a Spiking Neural Network). Two SC-DCNN~\cite{ren2017sc} designs were chosen for comparison: SC-DCNN (type a) has higher classification accuracy while SC-DCNN (type b) has higher energy efficiency. To compare with these designs, we implement two configurations of LeNet-5, Ours (design 1) and Ours (design 2), by running the column-combining algorithm with two different target numbers of nonzero weights $\rho$ (8k for design 1 and 5K for design 2). Both designs use ($\alpha=8,\beta=20,\gamma=0.5$).

Table~\ref{table:asic-lenet} compares all designs in terms of accuracy, area efficiency, and energy efficiency. Generally, our design has both the highest area efficiency and energy efficiency across all the designs. Compared to SC-DCNN (type a), our design 1 achieves a 2.2$\times$ improvement in area efficiency and a 3$\times$ improvement in energy efficiency, while also attaining a higher classification accuracy. Similarly, our design 2 realizes a higher classification accuracy than SC-DCNN (type b), while achieving a 1.4$\times$ improvement in area efficiency and a 1.7$\times$ improvement in energy efficiency. 

\begin{table}
\centering
 \begin{adjustbox}{width=\columnwidth,center}
 \begin{tabular}{|l c c c c c|}
 \hline
  Platform & Network & Platform & Accuracy & Area Eff. & Energy Eff. \\
 \hline
  Ours (design 1) & CNN & ASIC & 98.32\% & 46603 & 658053  \\
  Ours (design 2) & CNN & ASIC & 97.61\% & 64716 & 869402 \\
  SC-DCNN (type a) & CNN & ASIC & 98.26\% & 21439 & 221287 \\
  SC-DCNN (type b) & CNN & ASIC & 96.64\% & 45946 & 510734 \\
  2x Xeon W5580 & CNN & CPU & 98.46\% & 2.5 & 4.2 \\
  Tesla C2075 & CNN & GPU & 98.46\% & 4.5 & 3.2\\
  SpiNNaker & DBN & ARM & 95.00\% & N/A & 166.7 \\
  TrueNorth & SNN & ASIC & 99.42\% & 2.3 & 9259 \\
 \hline
 \end{tabular}
 \end{adjustbox}
 \caption{Comparison of our ASIC implementations of LeNet-5 to other CNN accelerators on MNIST.}
 \label{table:asic-lenet}
\end{table}
\subsection{Optimality in Energy Efficiency}
\label{sec:optimality-in-energy}
%Our designs stand out from the other designs in terms of energy efficiency for two main reasons: (1) high utilization efficiency of systolic arrays (MACs in the array) in implementing sparse CNNs resulting from column combining, and (2) efficient hardware implementations resulting from regular control and layout of systolic arrays. 
We provide an analysis showing that our systolic array design can achieve near-optimal energy efficiency.
%when the energy cost for I/O is significantly lower than that for MAC operations.
%During training, the number of nonzero weights has been greatly decreased, via weight pruning and channel combining. This reduces the number of MAC operations and therefore the energy consumption of the hardware. 
The total energy consumption of processing an input sample is:
%on a specific hardware platform, 
%$E_{total}$ is:
\setlength{\abovedisplayskip}{5pt}
\setlength{\belowdisplayskip}{5pt}
\begingroup\makeatletter\def\f@size{9}\check@mathfonts
\[E_{total} = E_{comp} + E_{mem} = E_{mac} \times N_{mac} +  E_{mem} = E_{mac} \times c N_{mac}^{opt} + E_{mem}\]\endgroup
where $E_{comp}$ and $E_{mem}$ are the energy consumption for all MAC computations and SRAM, respectively, $E_{mac}$ is the energy consumption for a single MAC operation, $N_{mac}$ is the number of MAC operations in the pruned network, and $N_{mac}^{opt}$ is the optimal number of MAC operations. 
Let $c (c\geq 1)$ denotes the ratio between $N_{mac}$ and $N_{mac}^{opt}$. Suppose that all designs have the same $E_{mac}$ and $E_{mem}$. Then the energy efficiency of a design is:
\begingroup\makeatletter\def\f@size{9}\check@mathfonts
$$\text{Energy Eff.} = \frac{1}{E_{total}} \nonumber 
                       = \frac{1}{E_{comp} + E_{mem}} \nonumber
                       = \frac{1}{E_{mac} \times c N_{mac}^{opt} + E_{mem}}$$\endgroup
and the optimal energy efficiency is:
\begingroup\makeatletter\def\f@size{9}\check@mathfonts
$$\text{Optimal Energy Eff.} = \frac{1}{{E_{mac} \times N_{mac}^{opt} + E_{mem}}}$$\endgroup
We have observed from synthesized results that when the input size is relatively small,
$r = \frac{E_{mem}}{E_{comp}}$ 
tends to be small. For example, $r = 0.06$ and $r = 0.1$ for LeNet-5 and ResNet-20, respectively. In this case, we have
\begingroup\makeatletter\def\f@size{9}\check@mathfonts
\begin{align}
\frac{\text{Energy Eff.}}{\text{Optimal Energy Eff.}} & = \frac{E_{mac}\times N_{mac}^{opt} + E_{mem}}{E_{mac}\times c N_{mac}^{opt} + E_{mem}} = %\frac{\frac{1}{c}E_{comp} + E_{mem}}{E_{comp} + E_{mem}} \nonumber \\
                                                          % & =
                                                          \frac{\frac{1}{c} + r}{1 + r} \approx \frac{1}{c} \nonumber
\end{align}
\endgroup
Note that ${1/c}$ is the packing efficiency achievable by column combining. Thus when ${r}$ is small, the ratio between Energy Eff. and Optimal Energy Eff. is mostly denominated by the packaging efficiency.

Consider, for example, the scenario depicted in Figure~\ref{fig:gamma} (c), for $\gamma = 0.5$. Column combining can achieve a packing efficiency of about 94.5\% with a modest degradation of classification accuracy of about 0.7\% in absolute percentage. Thus in this case the energy efficiency of our design is about 94.5\% of the optimal energy efficiency, for small ${r}$.

\subsection{FGPA Implementation and Evaluation}
\label{sec:eval-fpga}
For our FPGA implementation, we use the Xilinx XCKU035-1FBVA676C chip~\cite{xcku035}. We synthesize our design using the Xilinx Vivado Design Suite (2017.4)~\cite{vivado}. We use 32-bit accumulation for the systolic array implementation.

Table~\ref{table:cifar10-comparison} compares our ResNet-20 implementation to other FPGA implementations for CIFAR-10 in terms of classification accuracy and energy efficiency. We notice that our design achieves an accuracy of 93.1\%, which is around 5-6\% higher than other models. Moreover, our design achieves a 3$\times$ improvement on energy efficiency over the next best design. While it is possible for the other designs to increase the accuracy by using more hardware, it is hard for them to attain a low energy efficiency as our design.
 
\begin{table}
\centering
 \begin{adjustbox}{width=\columnwidth,center}
 \begin{tabular}{|l c c c c|} 
 \hline
 & \cite{truenorthcifar} & \cite{zhao2017accelerating} & \cite{ding2017c} & Ours \\
 \hline\hline
 Frequency (MHz)                     & N/A & 143 & 100 & 150 \\
 Precision (data/weight)                           & N/A & 1 & 16 & 8 \\
 Classification Accuracy        & N/A & 87.73\% & 88.3\% & 93.1\% \\
 %Throughput (frames/second)          & 1249 & 168 & 726 & 1903 \\
 Energy Efficiency (frames/joule)    & 6109 & 1320 & 36 & 18830 \\
 \hline
 \end{tabular}
 \end{adjustbox}
 \caption{Comparison of our ResNet-20 model to state-of-the-art FPGA implementations for CIFAR-10.}
 \label{table:cifar10-comparison}
\end{table}

\subsection{Dramatic Reduction in End-to-end Inference Latency with Cross-layer Pipelining}
\label{sec:pipeline-results}
In this section, we evaluate the FPGA performance of cross-layer pipelining, described in Section~\ref{sec:cross-layer-pipeline}, in terms of reduced end-to-end inference latency for a single sample on LeNet-5 and ResNet-20. We found that cross-layer pipelining reduces the latency significantly by $3.5\times$ and $9.3\times$ compared to without pipelining for LeNet-5 and ResNet-20, respectively. 

Furthermore, we compare our column combined ResNet-20 model with cross-layer pipelining on FPGA to other hardware implementations including GPU, CPU and FPGA accelerators trained on CIFAR-10. Table~\ref{table:latency-comparison} shows the classification accuracy and end-to-end latency for a single input sample of each design. The latency 652$\mu$s of \cite{lin2017binary} shown in Table~\ref{table:latency-comparison} only includes the latency for all convolutional layers (thus >652). Our design achieves an end-to-end latency over $12\times$ smaller than next best implementation, while also obtaining a higher classification accuracy. 

\begin{table}
\centering
 \begin{adjustbox}{width=\columnwidth,center}
 \begin{tabular}{|l c c c c c|} 
 \hline
 & CPU\cite{zhao2017accelerating} & GPU\cite{zhao2017accelerating} & \cite{zhao2017accelerating} & \cite{lin2017binary} & Ours \\
 \hline\hline
 %Frequency (MHz)                & N/A & N/A & 143 & N/A & 150 \\
 Classification Accuracy        & 88.42\% & 88.42\% & 88.42\% & 85.88\% & 93.1\% \\
 Latency (microseconds/frame)   & 14800 & 730 & 5940 & >652 & 55.68 \\
 \hline
 \end{tabular}
 \end{adjustbox}
 \caption{Comparison of our ResNet-20 model with cross-layer pipelining to state-of-the-art CNN accelerators for CIFAR-10.}
 \label{table:latency-comparison}
\end{table}
%Although the bit-serial design have a lower input rate than bit-parallel in general, 
%This shows the cross-layer pipelining can dramatically reduce the end-to-end latency that low latency inference can still be achieved with a bit-serial design by beginning computation as soon as it becomes available.

\section{Conclusion}
In this paper, for CNN inference, we have presented a solution to a long-standing parallel processing challenge about how one can make efficient use of regular parallel processing arrays, such as systolic arrays, for \textit{sparse} computations. Specifically, for a given sparse CNN, we have proposed a novel approach of using column combining to pack the filter matrix associated with each convolutional layer for its efficient systolic array implementation. In combining columns, we prune all weights on conflicting rows but the one with the largest magnitude. We then bring up the classification accuracy of the pruned network via retraining. We iterate on this column-combining and network-retraining step to improve both utilization efficiency of the systolic array and the classification accuracy of the network. This joint optimization has become feasible for sparse CNN inference. That is, for a CNN, we can optimize its topologies to fit the structure of the underlying computing hardware such as a given systolic array, while preserving most of its classification accuracy via network training. 

Being able to transform sparse computations to fit highly efficient regular processor arrays is powerful. As demonstrated in the paper, our proposed column combining approach can increase the utilization efficiency of a systolic array by approximately 4$\times$, with a slight increase in the complexity of systolic cells for providing multiplexing (MX) support. This has led to superior performance of our proposed method against prior arts under metrics such as energy efficiency (3$\times$) and inference latency (12$\times$).

%\section*{Acknowledgements}
% For final version

\bibliographystyle{plain}
\bibliography{main.bbl}

\begin{thebibliography}{10}

\bibitem{cacti}
Cacti: An integrated cache and memory access time, cycle time, area, leakage,
  and dynamic power model.
\newblock \url{https://github.com/HewlettPackard/cacti}.

\bibitem{synopsysdesigncompiler}
Design compiler: Rtl synthesis.
\newblock
  \url{https://www.synopsys.com/support/training/rtl-synthesis/design-compiler-rtl-synthesis.html}.

\bibitem{nangatelib}
Nangate freepdk45 open cell library.
\newblock \url{http://www.nangate.com/?page_id=2325}.

\bibitem{vivado}
Vivado design suite - hlx editions productivity. multiplied.
\newblock \url{https://www.xilinx.com/products/design-tools/vivado.html}.

\bibitem{xcku035}
Xilinx inc. xcku035-1fbva676c.
\newblock
  \url{https://www.digikey.ca/product-detail/en/xilinx-inc/XCKU035-1FBVA676C/122-1989-ND/6132038}.

\bibitem{akopyan2015truenorth}
Filipp Akopyan, Jun Sawada, Andrew Cassidy, Rodrigo Alvarez-Icaza, John Arthur,
  Paul Merolla, Nabil Imam, Yutaka Nakamura, Pallab Datta, Gi-Joon Nam, Brian
  Taba, Michael Beakes, Bernard Brezzo, Jente Kuang, Rajit Manohar, William
  Risk, Bryan Jackson, and Dharmendra Modha.
\newblock Truenorth: Design and tool flow of a 65 mw 1 million neuron
  programmable neurosynaptic chip.
\newblock {\em IEEE Transactions on Computer-Aided Design of Integrated
  Circuits and Systems}, 34(10):1537--1557, 2015.

\bibitem{bang201714}
Suyoung Bang, Jingcheng Wang, Ziyun Li, Cao Gao, Yejoong Kim, Qing Dong, Yen-Po
  Chen, Laura Fick, Xun Sun, Ron Dreslinski, Trevor Mudge, Hun~Seok Kim, David
  Blaauw, and Dennis Sylvester.
\newblock 14.7 a 288$\mu$w programmable deep-learning processor with 270kb
  on-chip weight storage using non-uniform memory hierarchy for mobile
  intelligence.
\newblock In {\em Solid-State Circuits Conference (ISSCC), 2017 IEEE
  International}, pages 250--251. IEEE, 2017.

\bibitem{bengio2009curriculum}
Yoshua Bengio, J{\'e}r{\^o}me Louradour, Ronan Collobert, and Jason Weston.
\newblock Curriculum learning.
\newblock In {\em Proceedings of the 26th annual international conference on
  machine learning}, pages 41--48. ACM, 2009.

\bibitem{chen2014diannao}
Tianshi Chen, Zidong Du, Ninghui Sun, Jia Wang, Chengyong Wu, Yunji Chen, and
  Olivier Temam.
\newblock Diannao: A small-footprint high-throughput accelerator for ubiquitous
  machine-learning.
\newblock {\em ACM Sigplan Notices}, 49(4):269--284, 2014.

\bibitem{chen2017eyeriss}
Yu-Hsin Chen, Tushar Krishna, Joel~S Emer, and Vivienne Sze.
\newblock Eyeriss: An energy-efficient reconfigurable accelerator for deep
  convolutional neural networks.
\newblock {\em IEEE Journal of Solid-State Circuits}, 52(1):127--138, 2017.

\bibitem{chen2014dadiannao}
Yunji Chen, Tao Luo, Shaoli Liu, Shijin Zhang, Liqiang He, Jia Wang, Ling Li,
  Tianshi Chen, Zhiwei Xu, Ninghui Sun, and Olivier Temam.
\newblock Dadiannao: A machine-learning supercomputer.
\newblock In {\em Proceedings of the 47th Annual IEEE/ACM International
  Symposium on Microarchitecture}, pages 609--622. IEEE Computer Society, 2014.

\bibitem{chollet2016xception}
Fran{\c{c}}ois Chollet.
\newblock Xception: Deep learning with depthwise separable convolutions.
\newblock {\em arXiv preprint arXiv:1610.02357}, 2016.

\bibitem{clemons2016patch}
Jason Clemons, Chih-Chi Cheng, Iuri Frosio, Daniel Johnson, and Stephen~W
  Keckler.
\newblock A patch memory system for image processing and computer vision.
\newblock In {\em Microarchitecture (MICRO), 2016 49th Annual IEEE/ACM
  International Symposium on}, pages 1--13. IEEE, 2016.

\bibitem{courbariaux2016binarized}
Matthieu Courbariaux, Itay Hubara, Daniel Soudry, Ran El-Yaniv, and Yoshua
  Bengio.
\newblock Binarized neural networks: Training deep neural networks with weights
  and activations constrained to+ 1 or-1.
\newblock {\em arXiv preprint arXiv:1602.02830}, 2016.

\bibitem{dicecco2017fpga}
Roberto DiCecco, Lin Sun, and Paul Chow.
\newblock Fpga-based training of convolutional neural networks with a reduced
  precision floating-point library.
\newblock In {\em Field Programmable Technology (ICFPT), 2017 International
  Conference on}, pages 239--242. IEEE, 2017.

\bibitem{ding2017c}
Caiwen Ding, Siyu Liao, Yanzhi Wang, Zhe Li, Ning Liu, Youwei Zhuo, Chao Wang,
  Xuehai Qian, Yu~Bai, Geng Yuan, Xiaolong Ma, Yipeng Zhang, Jian Tang, Qinru
  Qiu, Xue Lin, and Bo~Yuan.
\newblock Circnn: accelerating and compressing deep neural networks using
  block-circulant weight matrices.
\newblock In {\em Proceedings of the 50th Annual IEEE/ACM International
  Symposium on Microarchitecture}, pages 395--408. ACM, 2017.

\bibitem{du2015shidiannao}
Zidong Du, Robert Fasthuber, Tianshi Chen, Paolo Ienne, Ling Li, Tao Luo,
  Xiaobing Feng, Yunji Chen, and Olivier Temam.
\newblock Shidiannao: Shifting vision processing closer to the sensor.
\newblock In {\em ACM SIGARCH Computer Architecture News}, volume~43, pages
  92--104. ACM, 2015.

\bibitem{lin2017binary}
Xing T. Zhao R. Zhang Z. Srivastava M. B. Tu Z. Gupta R.~K. ELin, J.~H.
\newblock Binarized convolutional neural networks with separable filters for
  efficient hardware acceleration.
\newblock In {\em CVPR Workshops}, pages 344--352, 2017.

\bibitem{gray2017blocksparse}
Scott Gray, Alec Radford, and Diederik Kingma.
\newblock Gpu kernels for block-sparse weights.
\newblock
  \url{https://s3-us-west-2.amazonaws.com/openai-assets/blocksparse/blocksparsepaper.pdf},
  2017.
\newblock [Online; accessed 12-January-2018].

\bibitem{gysel2016hardware}
Philipp Gysel, Mohammad Motamedi, and Soheil Ghiasi.
\newblock Hardware-oriented approximation of convolutional neural networks.
\newblock {\em arXiv preprint arXiv:1604.03168}, 2016.

\bibitem{han2017ese}
Song Han, Junlong Kang, Huizi Mao, Yiming Hu, Xin Li, Yubin Li, Dongliang Xie,
  Hong Luo, Song Yao, Yu~Wang, Huazhong Yang, and William~J. Dally.
\newblock Ese: Efficient speech recognition engine with sparse lstm on fpga.
\newblock In {\em Proceedings of the 2017 ACM/SIGDA International Symposium on
  Field-Programmable Gate Arrays}, pages 75--84. ACM, 2017.

\bibitem{han2015deep}
Song Han, Huizi Mao, and William~J Dally.
\newblock Deep compression: Compressing deep neural networks with pruning,
  trained quantization and huffman coding.
\newblock {\em arXiv preprint arXiv:1510.00149}, 2015.

\bibitem{he2016deep}
Kaiming He, Xiangyu Zhang, Shaoqing Ren, and Jian Sun.
\newblock Deep residual learning for image recognition.
\newblock In {\em Proceedings of the IEEE conference on computer vision and
  pattern recognition}, pages 770--778, 2016.

\bibitem{he2017channel}
Yihui He, Xiangyu Zhang, and Jian Sun.
\newblock Channel pruning for accelerating very deep neural networks.

\bibitem{howard2017mobilenets}
Andrew~G Howard, Menglong Zhu, Bo~Chen, Dmitry Kalenichenko, Weijun Wang,
  Tobias Weyand, Marco Andreetto, and Hartwig Adam.
\newblock Mobilenets: Efficient convolutional neural networks for mobile vision
  applications.
\newblock {\em arXiv preprint arXiv:1704.04861}, 2017.

\bibitem{huang2017condensenet}
Gao Huang, Shichen Liu, Laurens van~der Maaten, and Kilian~Q Weinberger.
\newblock Condensenet: An efficient densenet using learned group convolutions.
\newblock {\em arXiv preprint arXiv:1711.09224}, 2017.

\bibitem{jouppi2017datacenter}
Norman~P. Jouppi, Cliff Young, Nishant Patil, David Patterson, Gaurav Agrawal,
  Raminder Bajwa, Sarah Bates, Suresh Bhatia, Nan Boden, Al~Borchers, Rick
  Boyle, Pierre-luc Cantin, Clifford Chao, Chris Clark, Jeremy Coriell, Mike
  Daley, Matt Dau, Jeffrey Dean, Ben Gelb, Tara~Vazir Ghaemmaghami, Rajendra
  Gottipati, William Gulland, Robert Hagmann, C.~Richard Ho, Doug Hogberg, John
  Hu, Robert Hundt, Dan Hurt, Julian Ibarz, Aaron Jaffey, Alek Jaworski,
  Alexander Kaplan, Harshit Khaitan, Daniel Killebrew, Andy Koch, Naveen Kumar,
  Steve Lacy, James Laudon, James Law, Diemthu Le, Chris Leary, Zhuyuan Liu,
  Kyle Lucke, Alan Lundin, Gordon MacKean, Adriana Maggiore, Maire Mahony,
  Kieran Miller, Rahul Nagarajan, Ravi Narayanaswami, Ray Ni, Kathy Nix, Thomas
  Norrie, Mark Omernick, Narayana Penukonda, Andy Phelps, Jonathan Ross, Matt
  Ross, Amir Salek, Emad Samadiani, Chris Severn, Gregory Sizikov, Matthew
  Snelham, Jed Souter, Dan Steinberg, Andy Swing, Mercedes Tan, Gregory
  Thorson, Bo~Tian, Horia Toma, Erick Tuttle, Vijay Vasudevan, Richard Walter,
  Walter Wang, Eric Wilcox, and Doe~Hyun Yoon.
\newblock In-datacenter performance analysis of a tensor processing unit.
\newblock In {\em Proceedings of the 44th Annual International Symposium on
  Computer Architecture}, ISCA '17, pages 1--12, New York, NY, USA, 2017. ACM.

\bibitem{khan2008spinnaker}
Muhammad~Mukaram Khan, David~R Lester, Luis~A Plana, A~Rast, Xin Jin, Eustace
  Painkras, and Stephen~B Furber.
\newblock Spinnaker: mapping neural networks onto a massively-parallel chip
  multiprocessor.
\newblock In {\em Neural Networks, 2008. IJCNN 2008.(IEEE World Congress on
  Computational Intelligence). IEEE International Joint Conference on}, pages
  2849--2856. Ieee, 2008.

\bibitem{krizhevsky2014cifar}
Alex Krizhevsky, Vinod Nair, and Geoffrey Hinton.
\newblock The cifar-10 dataset, 2014.

\bibitem{krizhevsky2012imagenet}
Alex Krizhevsky, Ilya Sutskever, and Geoffrey~E Hinton.
\newblock Imagenet classification with deep convolutional neural networks.
\newblock In {\em Advances in neural information processing systems}, pages
  1097--1105, 2012.

\bibitem{kung1982systolic}
H.~T. Kung.
\newblock Why systolic architectures?
\newblock {\em IEEE Computer}, 15:37--46, 1982.

\bibitem{kungLeisersonSystolicArrays1978}
H.~T. Kung and C.~E. Leiserson.
\newblock Systolic arrays (for vlsi).
\newblock In {\em Sparse Matrix Proceedings 1978}, pages 256--282. Society for
  Industrial and Applied Mathematics, 1979.

\bibitem{lecun1998mnist}
Yann LeCun.
\newblock The mnist database of handwritten digits.
\newblock {\em http://yann. lecun. com/exdb/mnist/}, 1998.

\bibitem{lecun1998gradient}
Yann LeCun, L{\'e}on Bottou, Yoshua Bengio, and Patrick Haffner.
\newblock Gradient-based learning applied to document recognition.
\newblock {\em Proceedings of the IEEE}, 86(11):2278--2324, 1998.

\bibitem{lin2016fixed}
Darryl Lin, Sachin Talathi, and Sreekanth Annapureddy.
\newblock Fixed point quantization of deep convolutional networks.
\newblock In {\em International Conference on Machine Learning}, pages
  2849--2858, 2016.

\bibitem{lin2015neural}
Zhouhan Lin, Matthieu Courbariaux, Roland Memisevic, and Yoshua Bengio.
\newblock Neural networks with few multiplications.
\newblock {\em arXiv preprint arXiv:1510.03009}, 2015.

\bibitem{loshchilov2016sgdr}
Ilya Loshchilov and Frank Hutter.
\newblock Sgdr: stochastic gradient descent with restarts.
\newblock {\em Learning}, 10:3, 2016.

\bibitem{luo2017thinet}
Jian-Hao Luo, Jianxin Wu, and Weiyao Lin.
\newblock Thinet: A filter level pruning method for deep neural network
  compression.
\newblock {\em arXiv preprint arXiv:1707.06342}, 2017.

\bibitem{ma2017optimizing}
Yufei Ma, Yu~Cao, Sarma Vrudhula, and Jae-sun Seo.
\newblock Optimizing loop operation and dataflow in fpga acceleration of deep
  convolutional neural networks.
\newblock In {\em Proceedings of the 2017 ACM/SIGDA International Symposium on
  Field-Programmable Gate Arrays}, pages 45--54. ACM, 2017.

\bibitem{Merritt2018ARM}
Rick Merritt.
\newblock Arm at risk on ai chip marketc.
\newblock {\em EE Times India}, April 2018.

\bibitem{narang2017block}
Sharan Narang, Eric Undersander, and Gregory~F. Diamos.
\newblock Block-sparse recurrent neural networks.
\newblock {\em CoRR}, abs/1711.02782, 2017.

\bibitem{baiduhotchips}
Jian Ouyang, Ephrem Wu, Jing Wang, Yupeng Li, and Hanlin Xie.
\newblock Xpu: A programmable fpga accelerator for diverse workloads.
\newblock {\em Hot Chips}, 2017.

\bibitem{park2016fpga}
Jinhwan Park and Wonyong Sung.
\newblock Fpga based implementation of deep neural networks using on-chip
  memory only.
\newblock In {\em Acoustics, Speech and Signal Processing (ICASSP), 2016 IEEE
  International Conference on}, pages 1011--1015. IEEE, 2016.

\bibitem{park2017scale}
Jongse Park, Hardik Sharma, Divya Mahajan, Joon~Kyung Kim, Preston Olds, and
  Hadi Esmaeilzadeh.
\newblock Scale-out acceleration for machine learning.
\newblock In {\em Proceedings of the 50th Annual IEEE/ACM International
  Symposium on Microarchitecture}, pages 367--381. ACM, 2017.

\bibitem{qiu2016going}
Jiantao Qiu, Jie Wang, Song Yao, Kaiyuan Guo, Boxun Li, Erjin Zhou, Jincheng
  Yu, Tianqi Tang, Ningyi Xu, Sen Song, Yu~Wang, and Huazhong Yang.
\newblock Going deeper with embedded fpga platform for convolutional neural
  network.
\newblock In {\em Proceedings of the 2016 ACM/SIGDA International Symposium on
  Field-Programmable Gate Arrays}, pages 26--35. ACM, 2016.

\bibitem{reagen2016minerva}
Brandon Reagen, Paul Whatmough, Robert Adolf, Saketh Rama, Hyunkwang Lee,
  Sae~Kyu Lee, Jos{\'e}~Miguel Hern{\'a}ndez-Lobato, Gu-Yeon Wei, and David
  Brooks.
\newblock Minerva: Enabling low-power, highly-accurate deep neural network
  accelerators.
\newblock In {\em ACM SIGARCH Computer Architecture News}, volume~44, pages
  267--278. IEEE Press, 2016.

\bibitem{ren2017sc}
Ao~Ren, Zhe Li, Caiwen Ding, Qinru Qiu, Yanzhi Wang, Ji~Li, Xuehai Qian, and
  Bo~Yuan.
\newblock Sc-dcnn: Highly-scalable deep convolutional neural network using
  stochastic computing.
\newblock {\em ACM SIGOPS Operating Systems Review}, 51(2):405--418, 2017.

\bibitem{rhu2016vdnn}
Minsoo Rhu, Natalia Gimelshein, Jason Clemons, Arslan Zulfiqar, and Stephen~W
  Keckler.
\newblock vdnn: Virtualized deep neural networks for scalable, memory-efficient
  neural network design.
\newblock In {\em Microarchitecture (MICRO), 2016 49th Annual IEEE/ACM
  International Symposium on}, pages 1--13. IEEE, 2016.

\bibitem{Rojas1996}
R.~Rojas.
\newblock {\em Neural Networks - A Systematic Introduction, Chapter 18:
  Hardware for Neural Networks}.
\newblock Springer-Verlag, 1996.

\bibitem{romero2014fitnets}
Adriana Romero, Nicolas Ballas, Samira~Ebrahimi Kahou, Antoine Chassang, Carlo
  Gatta, and Yoshua Bengio.
\newblock Fitnets: Hints for thin deep nets.
\newblock {\em arXiv preprint arXiv:1412.6550}, 2014.

\bibitem{ruder2016overview}
Sebastian Ruder.
\newblock An overview of gradient descent optimization algorithms.
\newblock {\em arXiv preprint arXiv:1609.04747}, 2016.

\bibitem{sankaradas2009massively}
Murugan Sankaradas, Venkata Jakkula, Srihari Cadambi, Srimat Chakradhar, Igor
  Durdanovic, Eric Cosatto, and Hans~Peter Graf.
\newblock A massively parallel coprocessor for convolutional neural networks.
\newblock In {\em Application-specific Systems, Architectures and Processors,
  2009. ASAP 2009. 20th IEEE International Conference on}, pages 53--60. IEEE,
  2009.

\bibitem{shen2017escher}
Yongming Shen, Michael Ferdman, and Peter Milder.
\newblock Escher: A cnn accelerator with flexible buffering to minimize
  off-chip transfer.
\newblock In {\em Field-Programmable Custom Computing Machines (FCCM), 2017
  IEEE 25th Annual International Symposium on}, pages 93--100. IEEE, 2017.

\bibitem{shen2017maximizing}
Yongming Shen, Michael Ferdman, and Peter Milder.
\newblock Maximizing cnn accelerator efficiency through resource partitioning.
\newblock In {\em Proceedings of the 44th Annual International Symposium on
  Computer Architecture}, pages 535--547. ACM, 2017.

\bibitem{simonyan2014very}
Karen Simonyan and Andrew Zisserman.
\newblock Very deep convolutional networks for large-scale image recognition.
\newblock {\em arXiv preprint arXiv:1409.1556}, 2014.

\bibitem{song2017pipelayer}
Linghao Song, Xuehai Qian, Hai Li, and Yiran Chen.
\newblock Pipelayer: A pipelined reram-based accelerator for deep learning.
\newblock In {\em High Performance Computer Architecture (HPCA), 2017 IEEE
  International Symposium on}, pages 541--552. IEEE, 2017.

\bibitem{truenorthcifar}
John V. Arthur Andrew S. Cassidy Rathinakumar Appuswamy Alexander Andreopoulos
  David J. Berg Jeffrey L. McKinstry Timothy Melano Davis R. Barch Carmelo di
  Nolfo Pallab Datta Arnon Amir Brian Taba Myron D.~Flickner Steven K.~Esser,
  Paul A.~Merolla and Dharmendra~S. Modha.
\newblock Convolutional networks for fast, energy-efficient neuromorphic
  computing.
\newblock {\em National Academy of Sciences}, 2016.

\bibitem{tarvainen2017mean}
Antti Tarvainen and Harri Valpola.
\newblock Mean teachers are better role models: Weight-averaged consistency
  targets improve semi-supervised deep learning results.
\newblock In {\em Advances in neural information processing systems}, pages
  1195--1204, 2017.

\bibitem{umuroglu2017finn}
Yaman Umuroglu, Nicholas~J Fraser, Giulio Gambardella, Michaela Blott, Philip
  Leong, Magnus Jahre, and Kees Vissers.
\newblock Finn: A framework for fast, scalable binarized neural network
  inference.
\newblock In {\em Proceedings of the 2017 ACM/SIGDA International Symposium on
  Field-Programmable Gate Arrays}, pages 65--74. ACM, 2017.

\bibitem{wang2017dlau}
Chao Wang, Lei Gong, Qi~Yu, Xi~Li, Yuan Xie, and Xuehai Zhou.
\newblock Dlau: A scalable deep learning accelerator unit on fpga.
\newblock {\em IEEE Transactions on Computer-Aided Design of Integrated
  Circuits and Systems}, 36(3):513--517, 2017.

\bibitem{wang2017chain}
Shihao Wang, Dajiang Zhou, Xushen Han, and Takeshi Yoshimura.
\newblock Chain-nn: An energy-efficient 1d chain architecture for accelerating
  deep convolutional neural networks.
\newblock In {\em 2017 Design, Automation \& Test in Europe Conference \&
  Exhibition (DATE)}, pages 1032--1037. IEEE, 2017.

\bibitem{wang2016re}
Ying Wang, Huawei Li, and Xiaowei Li.
\newblock Re-architecting the on-chip memory sub-system of machine-learning
  accelerator for embedded devices.
\newblock In {\em Proceedings of the 35th International Conference on
  Computer-Aided Design}, page~13. ACM, 2016.

\bibitem{wei2017automated}
Xuechao Wei, Cody~Hao Yu, Peng Zhang, Youxiang Chen, Yuxin Wang, Han Hu, Yun
  Liang, and Jason Cong.
\newblock Automated systolic array architecture synthesis for high throughput
  cnn inference on fpgas.
\newblock In {\em Design Automation Conference (DAC), 2017 54th ACM/EDAC/IEEE},
  pages 1--6. IEEE, 2017.

\bibitem{wen2016learning}
Wei Wen, Chunpeng Wu, Yandan Wang, Yiran Chen, and Hai Li.
\newblock Learning structured sparsity in deep neural networks.
\newblock In {\em Advances in Neural Information Processing Systems}, pages
  2074--2082, 2016.

\bibitem{wu2017shift}
Bichen Wu, Alvin Wan, Xiangyu Yue, Peter Jin, Sicheng Zhao, Noah Golmant, Amir
  Gholaminejad, Joseph Gonzalez, and Kurt Keutzer.
\newblock Shift: A zero flop, zero parameter alternative to spatial
  convolutions.
\newblock {\em arXiv preprint arXiv:1711.08141}, 2017.

\bibitem{zhang2016caffeine}
Chen Zhang, Zhenman Fang, Peipei Zhou, Peichen Pan, and Jason Cong.
\newblock Caffeine: Towards uniformed representation and acceleration for deep
  convolutional neural networks.
\newblock In {\em Computer-Aided Design (ICCAD), 2016 IEEE/ACM International
  Conference on}, pages 1--8. IEEE, 2016.

\bibitem{zhang2015optimizing}
Chen Zhang, Peng Li, Guangyu Sun, Yijin Guan, Bingjun Xiao, and Jason Cong.
\newblock Optimizing fpga-based accelerator design for deep convolutional
  neural networks.
\newblock In {\em Proceedings of the 2015 ACM/SIGDA International Symposium on
  Field-Programmable Gate Arrays}, pages 161--170. ACM, 2015.

\bibitem{zhang2016cambricon}
Shijin Zhang, Zidong Du, Lei Zhang, Huiying Lan, Shaoli Liu, Ling Li, Qi~Guo,
  Tianshi Chen, and Yunji Chen.
\newblock Cambricon-x: An accelerator for sparse neural networks.
\newblock In {\em The 49th Annual IEEE/ACM International Symposium on
  Microarchitecture}, page~20. IEEE Press, 2016.

\bibitem{zhao2017aep}
Lei Zhao, Youtao Zhang, and Jun Yang.
\newblock Aep: An error-bearing neural network accelerator for energy
  efficiency and model protection.
\newblock In {\em Computer-Aided Design (ICCAD), 2017 IEEE/ACM International
  Conference on}, pages 765--771. IEEE, 2017.

\bibitem{zhao2017accelerating}
Ritchie Zhao, Weinan Song, Wentao Zhang, Tianwei Xing, Jeng-Hau Lin, Mani
  Srivastava, Rajesh Gupta, and Zhiru Zhang.
\newblock Accelerating binarized convolutional neural networks with
  software-programmable fpgas.
\newblock In {\em Proceedings of the 2017 ACM/SIGDA International Symposium on
  Field-Programmable Gate Arrays}, pages 15--24. ACM, 2017.

\end{thebibliography}


@article{han2015deep,
  title={Deep compression: Compressing deep neural networks with pruning, trained quantization and huffman coding},
  author={Han, Song and Mao, Huizi and Dally, William J},
  journal={arXiv preprint arXiv:1510.00149},
  year={2015}
}


@article{howard2017mobilenets,
  title={Mobilenets: Efficient convolutional neural networks for mobile vision applications},
  author={Howard, Andrew G and Zhu, Menglong and Chen, Bo and Kalenichenko, Dmitry and Wang, Weijun and Weyand, Tobias and Andreetto, Marco and Adam, Hartwig},
  journal={arXiv preprint arXiv:1704.04861},
  year={2017}
}

@article{huang2017condensenet,
  title={CondenseNet: An Efficient DenseNet using Learned Group Convolutions},
  author={Huang, Gao and Liu, Shichen and van der Maaten, Laurens and Weinberger, Kilian Q},
  journal={arXiv preprint arXiv:1711.09224},
  year={2017}
}

@article{paszke2017automatic,
  title={Automatic differentiation in PyTorch},
  author={Paszke, Adam and Gross, Sam and Chintala, Soumith and Chanan, Gregory and Yang, Edward and DeVito, Zachary and Lin, Zeming and Desmaison, Alban and Antiga, Luca and Lerer, Adam},
  year={2017}
}


@article{chen2017eyeriss,
  title={Eyeriss: An energy-efficient reconfigurable accelerator for deep convolutional neural networks},
  author={Chen, Yu-Hsin and Krishna, Tushar and Emer, Joel S and Sze, Vivienne},
  journal={IEEE Journal of Solid-State Circuits},
  volume={52},
  number={1},
  pages={127--138},
  year={2017},
  publisher={IEEE}
}

@inproceedings{tarvainen2017mean,
  title={Mean teachers are better role models: Weight-averaged consistency targets improve semi-supervised deep learning results},
  author={Tarvainen, Antti and Valpola, Harri},
  booktitle={Advances in neural information processing systems},
  pages={1195--1204},
  year={2017}
}

@article{romero2014fitnets,
  title={Fitnets: Hints for thin deep nets},
  author={Romero, Adriana and Ballas, Nicolas and Kahou, Samira Ebrahimi and Chassang, Antoine and Gatta, Carlo and Bengio, Yoshua},
  journal={arXiv preprint arXiv:1412.6550},
  year={2014}
}

@inproceedings{bengio2009curriculum,
  title={Curriculum learning},
  author={Bengio, Yoshua and Louradour, J{\'e}r{\^o}me and Collobert, Ronan and Weston, Jason},
  booktitle={Proceedings of the 26th annual international conference on machine learning},
  pages={41--48},
  year={2009},
  organization={ACM}
}

@article{ren2017sc,
  title={Sc-dcnn: Highly-scalable deep convolutional neural network using stochastic computing},
  author={Ren, Ao and Li, Zhe and Ding, Caiwen and Qiu, Qinru and Wang, Yanzhi and Li, Ji and Qian, Xuehai and Yuan, Bo},
  journal={ACM SIGOPS Operating Systems Review},
  volume={51},
  number={2},
  pages={405--418},
  year={2017},
  publisher={ACM}
}

@article{akopyan2015truenorth,
  title={Truenorth: Design and tool flow of a 65 mw 1 million neuron programmable neurosynaptic chip},
  author={Akopyan, Filipp and Sawada, Jun and Cassidy, Andrew and Alvarez-Icaza, Rodrigo and Arthur, John and Merolla, Paul and Imam, Nabil and Nakamura, Yutaka and Datta, Pallab and Nam, Gi-Joon and Taba, Brian and Beakes, Michael and Brezzo, Bernard and Kuang, Jente and Manohar, Rajit and Risk, William and Jackson, Bryan and Modha, Dharmendra},
  journal={IEEE Transactions on Computer-Aided Design of Integrated Circuits and Systems},
  volume={34},
  number={10},
  pages={1537--1557},
  year={2015},
  publisher={IEEE}
}

@inproceedings{khan2008spinnaker,
  title={SpiNNaker: mapping neural networks onto a massively-parallel chip multiprocessor},
  author={Khan, Muhammad Mukaram and Lester, David R and Plana, Luis A and Rast, A and Jin, Xin and Painkras, Eustace and Furber, Stephen B},
  booktitle={Neural Networks, 2008. IJCNN 2008.(IEEE World Congress on Computational Intelligence). IEEE International Joint Conference on},
  pages={2849--2856},
  year={2008},
  organization={Ieee}
}

@article{Merritt2018ARM,
  title={ARM at Risk on AI Chip Marketc},
  author={Merritt, Rick},
  year={2018},
  Month={April},
	Journal={EE Times India}
}

@inproceedings{albericio2016cnvlutin,
  title={Cnvlutin: Ineffectual-neuron-free deep neural network computing},
  author={Albericio, Jorge and Judd, Patrick and Hetherington, Tayler and Aamodt, Tor and Jerger, Natalie Enright and Moshovos, Andreas},
  booktitle={ACM SIGARCH Computer Architecture News},
  volume={44},
  number={3},
  pages={1--13},
  year={2016},
  organization={IEEE Press}
}

@inproceedings{zhang2016cambricon,
  title={Cambricon-x: An accelerator for sparse neural networks},
  author={Zhang, Shijin and Du, Zidong and Zhang, Lei and Lan, Huiying and Liu, Shaoli and Li, Ling and Guo, Qi and Chen, Tianshi and Chen, Yunji},
  booktitle={The 49th Annual IEEE/ACM International Symposium on Microarchitecture},
  pages={20},
  year={2016},
  organization={IEEE Press}
}

@inproceedings{nair2010rectified,
  title={Rectified linear units improve restricted boltzmann machines},
  author={Nair, Vinod and Hinton, Geoffrey E},
  booktitle={Proceedings of the 27th international conference on machine learning (ICML-10)},
  pages={807--814},
  year={2010}
}

@article{ioffe2015batch,
  title={Batch normalization: Accelerating deep network training by reducing internal covariate shift},
  author={Ioffe, Sergey and Szegedy, Christian},
  journal={arXiv preprint arXiv:1502.03167},
  year={2015}
}

@article{jacob2017quantization,
  title={Quantization and training of neural networks for efficient integer-arithmetic-only inference},
  author={Jacob, Benoit and Kligys, Skirmantas and Chen, Bo and Zhu, Menglong and Tang, Matthew and Howard, Andrew and Adam, Hartwig and Kalenichenko, Dmitry}
}

@inproceedings{parashar2017scnn,
  title={Scnn: An accelerator for compressed-sparse convolutional neural networks},
  author={Parashar, Angshuman and Rhu, Minsoo and Mukkara, Anurag and Puglielli, Antonio and Venkatesan, Rangharajan and Khailany, Brucek and Emer, Joel and Keckler, Stephen W and Dally, William J},
  booktitle={ACM SIGARCH Computer Architecture News},
  volume={45},
  number={2},
  pages={27--40},
  year={2017},
  organization={ACM}
}

@inproceedings{huang2017densely,
  title={Densely connected convolutional networks},
  author={Huang, Gao and Liu, Zhuang}
}

@article{wu2017shift,
  title={Shift: A Zero FLOP, Zero Parameter Alternative to Spatial Convolutions},
  author={Wu, Bichen and Wan, Alvin and Yue, Xiangyu and Jin, Peter and Zhao, Sicheng and Golmant, Noah and Gholaminejad, Amir and Gonzalez, Joseph and Keutzer, Kurt},
  journal={arXiv preprint arXiv:1711.08141},
  year={2017}
}

@article{kung1982systolic,
  title={Why systolic architectures?},
  author={Kung, H. T.},
  volume={15},
  issue={1},
  year={1982},
  journal = {IEEE Computer},
  pages={37--46}
}

@book{Rojas1996,
AUTHOR = {R. Rojas},
TITLE = {Neural Networks - A Systematic Introduction, Chapter 18: Hardware for Neural Networks},
PUBLISHER = {Springer-Verlag},
YEAR = {1996},
Chapter = {18}
%Chapter = {Hardware for Neural Networks}
}

@article{brentkung1982,
  title={A Regular Layout for Parallel Adders},
  author={Brent, R. P. and Kung, H. T.},
  journal={IEEE Transactions on Computers},
  volume={C-31},
  Issue={3},
  pages={260--264},
  year={1982}
}

@misc{krizhevsky2014cifar,
  title={The CIFAR-10 dataset},
  author={Krizhevsky, Alex and Nair, Vinod and Hinton, Geoffrey},
  journal={2013-11-14]. http://www. cs. toronto. edu/\~{} kriz/cifar. html},
  year={2014}
}

@inproceedings{kungLeisersonSystolicArrays1978,
  title={Systolic Arrays (for VLSI)},
  author={Kung, H. T. and Leiserson, C. E.},
  year={1979},
  booktitle={Sparse Matrix Proceedings 1978},
  pages={256--282},
  organization={Society for Industrial and Applied Mathematics}
}


@inproceedings{jouppi2017datacenter,
 author = {Jouppi, Norman P. and Young, Cliff and Patil, Nishant and Patterson, David and Agrawal, Gaurav and Bajwa, Raminder and Bates, Sarah and Bhatia, Suresh and Boden, Nan and Borchers, Al and Boyle, Rick and Cantin, Pierre-luc and Chao, Clifford and Clark, Chris and Coriell, Jeremy and Daley, Mike and Dau, Matt and Dean, Jeffrey and Gelb, Ben and Ghaemmaghami, Tara Vazir and Gottipati, Rajendra and Gulland, William and Hagmann, Robert and Ho, C. Richard and Hogberg, Doug and Hu, John and Hundt, Robert and Hurt, Dan and Ibarz, Julian and Jaffey, Aaron and Jaworski, Alek and Kaplan, Alexander and Khaitan, Harshit and Killebrew, Daniel and Koch, Andy and Kumar, Naveen and Lacy, Steve and Laudon, James and Law, James and Le, Diemthu and Leary, Chris and Liu, Zhuyuan and Lucke, Kyle and Lundin, Alan and MacKean, Gordon and Maggiore, Adriana and Mahony, Maire and Miller, Kieran and Nagarajan, Rahul and Narayanaswami, Ravi and Ni, Ray and Nix, Kathy and Norrie, Thomas and Omernick, Mark and Penukonda, Narayana and Phelps, Andy and Ross, Jonathan and Ross, Matt and Salek, Amir and Samadiani, Emad and Severn, Chris and Sizikov, Gregory and Snelham, Matthew and Souter, Jed and Steinberg, Dan and Swing, Andy and Tan, Mercedes and Thorson, Gregory and Tian, Bo and Toma, Horia and Tuttle, Erick and Vasudevan, Vijay and Walter, Richard and Wang, Walter and Wilcox, Eric and Yoon, Doe Hyun},
 title = {In-Datacenter Performance Analysis of a Tensor Processing Unit},
 booktitle = {Proceedings of the 44th Annual International Symposium on Computer Architecture},
 series = {ISCA '17},
 year = {2017},
 isbn = {978-1-4503-4892-8},
 location = {Toronto, ON, Canada},
 pages = {1--12},
 numpages = {12},
 url = {http://doi.acm.org/10.1145/3079856.3080246},
 doi = {10.1145/3079856.3080246},
 acmid = {3080246},
 publisher = {ACM},
 address = {New York, NY, USA},
 keywords = {CNN, DNN, GPU, LSTM, MLP, RNN, TPU, TensorFlow, accelerator, deep learning, domain-specific architecture, neural network},
} 


@article{chollet2016xception,
  title={Xception: Deep Learning with Depthwise Separable Convolutions},
  author={Chollet, Fran{\c{c}}ois},
  journal={arXiv preprint arXiv:1610.02357},
  year={2016}
}

@inproceedings{krizhevsky2012imagenet,
  title={Imagenet classification with deep convolutional neural networks},
  author={Krizhevsky, Alex and Sutskever, Ilya and Hinton, Geoffrey E},
  booktitle={Advances in neural information processing systems},
  pages={1097--1105},
  year={2012}
}

@article{simonyan2014very,
  title={Very deep convolutional networks for large-scale image recognition},
  author={Simonyan, Karen and Zisserman, Andrew},
  journal={arXiv preprint arXiv:1409.1556},
  year={2014}
}

@article{tibshirani1996regression,
  title={Regression shrinkage and selection via the lasso},
  author={Tibshirani, Robert},
  journal={Journal of the Royal Statistical Society. Series B (Methodological)},
  pages={267--288},
  year={1996},
  publisher={JSTOR}
}

@article{yuan2006model,
  title={Model selection and estimation in regression with grouped variables},
  author={Yuan, Ming and Lin, Yi},
  journal={Journal of the Royal Statistical Society: Series B (Statistical Methodology)},
  volume={68},
  number={1},
  pages={49--67},
  year={2006},
  publisher={Wiley Online Library}
}

@inproceedings{wen2016learning,
  title={Learning structured sparsity in deep neural networks},
  author={Wen, Wei and Wu, Chunpeng and Wang, Yandan and Chen, Yiran and Li, Hai},
  booktitle={Advances in Neural Information Processing Systems},
  pages={2074--2082},
  year={2016}
}

@article{narang2017block,
  author    = {Sharan Narang and
               Eric Undersander and
               Gregory F. Diamos},
  title     = {Block-Sparse Recurrent Neural Networks},
  journal   = {CoRR},
  volume    = {abs/1711.02782},
  year      = {2017},
  url       = {http://arxiv.org/abs/1711.02782},
  archivePrefix = {arXiv},
  eprint    = {1711.02782},
  timestamp = {Fri, 01 Dec 2017 14:22:24 +0100},
  biburl    = {http://dblp.org/rec/bib/journals/corr/abs-1711-02782},
  bibsource = {dblp computer science bibliography, http://dblp.org}
}

@misc{gray2017blocksparse,
  title={GPU Kernels for Block-Sparse Weights},
  author={Gray, Scott and Radford, Alec and Kingma, Diederik},
  howpublished="\url{https://s3-us-west-2.amazonaws.com/openai-assets/blocksparse/blocksparsepaper.pdf}",
  year={2017},
  note = "[Online; accessed 12-January-2018]"
}

@inproceedings{lin2016fixed,
  title={Fixed point quantization of deep convolutional networks},
  author={Lin, Darryl and Talathi, Sachin and Annapureddy, Sreekanth},
  booktitle={International Conference on Machine Learning},
  pages={2849--2858},
  year={2016}
}

@article{courbariaux2016binarized,
  title={Binarized neural networks: Training deep neural networks with weights and activations constrained to+ 1 or-1},
  author={Courbariaux, Matthieu and Hubara, Itay and Soudry, Daniel and El-Yaniv, Ran and Bengio, Yoshua},
  journal={arXiv preprint arXiv:1602.02830},
  year={2016}
}

@article{mcdanel2017embedded,
  title={Embedded binarized neural networks},
  author={McDanel, Bradley and Teerapittayanon, Surat and Kung, HT},
  journal={arXiv preprint arXiv:1709.02260},
  year={2017}
}

@inproceedings{he2016deep,
  title={Deep residual learning for image recognition},
  author={He, Kaiming and Zhang, Xiangyu and Ren, Shaoqing and Sun, Jian},
  booktitle={Proceedings of the IEEE conference on computer vision and pattern recognition},
  pages={770--778},
  year={2016}
}

@article{lecun1998mnist,
  title={The MNIST database of handwritten digits},
  author={LeCun, Yann},
  journal={http://yann. lecun. com/exdb/mnist/},
  year={1998}
}

@misc{fashion,
  author       = {Han Xiao and Kashif Rasul and Roland Vollgraf},
  title        = {Fashion-MNIST: a Novel Image Dataset for Benchmarking Machine Learning Algorithms},
  date         = {2017-08-28},
  year         = {2017},
  eprintclass  = {cs.LG},
  eprinttype   = {arXiv},
  eprint       = {cs.LG/1708.07747},
}

@inproceedings{deng2009imagenet,
  title={Imagenet: A large-scale hierarchical image database},
  author={Deng, Jia and Dong, Wei and Socher, Richard and Li, Li-Jia and Li, Kai and Fei-Fei, Li},
  booktitle={Computer Vision and Pattern Recognition, 2009. CVPR 2009. IEEE Conference on},
  pages={248--255},
  year={2009},
  organization={IEEE}
}

%FPGA Papers
@inproceedings{zhang2016caffeine,
  title={Caffeine: Towards uniformed representation and acceleration for deep convolutional neural networks},
  author={Zhang, Chen and Fang, Zhenman and Zhou, Peipei and Pan, Peichen and Cong, Jason},
  booktitle={Computer-Aided Design (ICCAD), 2016 IEEE/ACM International Conference on},
  pages={1--8},
  year={2016},
  organization={IEEE}
}

@inproceedings{ma2017optimizing,
  title={Optimizing loop operation and dataflow in FPGA acceleration of deep convolutional neural networks},
  author={Ma, Yufei and Cao, Yu and Vrudhula, Sarma and Seo, Jae-sun},
  booktitle={Proceedings of the 2017 ACM/SIGDA International Symposium on Field-Programmable Gate Arrays},
  pages={45--54},
  year={2017},
  organization={ACM}
}

@inproceedings{ding2017c,
  title={CirCNN: accelerating and compressing deep neural networks using block-circulant weight matrices},
  author={Ding, Caiwen and Liao, Siyu and Wang, Yanzhi and Li, Zhe and Liu, Ning and Zhuo, Youwei and Wang, Chao and Qian, Xuehai and Bai, Yu and Yuan, Geng and Ma, Xiaolong and Zhang, Yipeng and Tang, Jian and Qiu, Qinru and Lin, Xue and Yuan, Bo},
  booktitle={Proceedings of the 50th Annual IEEE/ACM International Symposium on Microarchitecture},
  pages={395--408},
  year={2017},
  organization={ACM}
}

@inproceedings{zhang2015optimizing,
  title={Optimizing fpga-based accelerator design for deep convolutional neural networks},
  author={Zhang, Chen and Li, Peng and Sun, Guangyu and Guan, Yijin and Xiao, Bingjun and Cong, Jason},
  booktitle={Proceedings of the 2015 ACM/SIGDA International Symposium on Field-Programmable Gate Arrays},
  pages={161--170},
  year={2015},
  organization={ACM}
}

@inproceedings{zhao2017accelerating,
  title={Accelerating binarized convolutional neural networks with software-programmable fpgas},
  author={Zhao, Ritchie and Song, Weinan and Zhang, Wentao and Xing, Tianwei and Lin, Jeng-Hau and Srivastava, Mani and Gupta, Rajesh and Zhang, Zhiru},
  booktitle={Proceedings of the 2017 ACM/SIGDA International Symposium on Field-Programmable Gate Arrays},
  pages={15--24},
  year={2017},
  organization={ACM}
}

@inproceedings{qiu2016going,
  title={Going deeper with embedded fpga platform for convolutional neural network},
  author={Qiu, Jiantao and Wang, Jie and Yao, Song and Guo, Kaiyuan and Li, Boxun and Zhou, Erjin and Yu, Jincheng and Tang, Tianqi and Xu, Ningyi and Song, Sen and Wang, Yu and Yang, Huazhong},
  booktitle={Proceedings of the 2016 ACM/SIGDA International Symposium on Field-Programmable Gate Arrays},
  pages={26--35},
  year={2016},
  organization={ACM}
}

@inproceedings{dicecco2017fpga,
  title={FPGA-based training of convolutional neural networks with a reduced precision floating-point library},
  author={DiCecco, Roberto and Sun, Lin and Chow, Paul},
  booktitle={Field Programmable Technology (ICFPT), 2017 International Conference on},
  pages={239--242},
  year={2017},
  organization={IEEE}
}

@inproceedings{sankaradas2009massively,
  title={A massively parallel coprocessor for convolutional neural networks},
  author={Sankaradas, Murugan and Jakkula, Venkata and Cadambi, Srihari and Chakradhar, Srimat and Durdanovic, Igor and Cosatto, Eric and Graf, Hans Peter},
  booktitle={Application-specific Systems, Architectures and Processors, 2009. ASAP 2009. 20th IEEE International Conference on},
  pages={53--60},
  year={2009},
  organization={IEEE}
}

@inproceedings{wei2017automated,
  title={Automated systolic array architecture synthesis for high throughput CNN inference on FPGAs},
  author={Wei, Xuechao and Yu, Cody Hao and Zhang, Peng and Chen, Youxiang and Wang, Yuxin and Hu, Han and Liang, Yun and Cong, Jason},
  booktitle={Design Automation Conference (DAC), 2017 54th ACM/EDAC/IEEE},
  pages={1--6},
  year={2017},
  organization={IEEE}
}

@inproceedings{jialiang2017opencl,
  title={An OpenCL™ deep learning accelerator on Arria 10},
  author={Aydonat, Utku and O'Connell, Shane and Capalija, Davor and Ling, Andrew C and Chiu, Gordon R},
  booktitle={Proceedings of the 2017 ACM/SIGDA International Symposium on Field-Programmable Gate Arrays},
  pages={25--34},
  year={2017},
  organization={ACM}
}

@inproceedings{aydonat2017opencl,
  title={Improving the performance of OpenCL-based FPGA acelerator for convolution neural network},
  author={Zhang Jialiang, and Jing Li},
  booktitle={Proceedings of the 2017 ACM/SIGDA International Symposium on Field-Programmable Gate Arrays},
  pages={55--64},
  year={2017},
  organization={ACM}
}

@article{wang2017dlau,
  title={DLAU: A scalable deep learning accelerator unit on FPGA},
  author={Wang, Chao and Gong, Lei and Yu, Qi and Li, Xi and Xie, Yuan and Zhou, Xuehai},
  journal={IEEE Transactions on Computer-Aided Design of Integrated Circuits and Systems},
  volume={36},
  number={3},
  pages={513--517},
  year={2017},
  publisher={IEEE}
}

@inproceedings{han2017ese,
  title={Ese: Efficient speech recognition engine with sparse lstm on fpga},
  author={Han, Song and Kang, Junlong and Mao, Huizi and Hu, Yiming and Li, Xin and Li, Yubin and Xie, Dongliang and Luo, Hong and Yao, Song and Wang, Yu and Yang, Huazhong and Dally, William J.},
  booktitle={Proceedings of the 2017 ACM/SIGDA International Symposium on Field-Programmable Gate Arrays},
  pages={75--84},
  year={2017},
  organization={ACM}
}

@inproceedings{park2016fpga,
  title={FPGA based implementation of deep neural networks using on-chip memory only},
  author={Park, Jinhwan and Sung, Wonyong},
  booktitle={Acoustics, Speech and Signal Processing (ICASSP), 2016 IEEE International Conference on},
  pages={1011--1015},
  year={2016},
  organization={IEEE}
}

@inproceedings{albericio2017bit,
  title={Bit-pragmatic deep neural network computing},
  author={Albericio, Jorge and Delm{\'a}s, Alberto and Judd, Patrick and Sharify, Sayeh and O'Leary, Gerard and Genov, Roman and Moshovos, Andreas},
  booktitle={Proceedings of the 50th Annual IEEE/ACM International Symposium on Microarchitecture},
  pages={382--394},
  year={2017},
  organization={ACM}
}

@inproceedings{park2017scale,
  title={Scale-out acceleration for machine learning},
  author={Park, Jongse and Sharma, Hardik and Mahajan, Divya and Kim, Joon Kyung and Olds, Preston and Esmaeilzadeh, Hadi},
  booktitle={Proceedings of the 50th Annual IEEE/ACM International Symposium on Microarchitecture},
  pages={367--381},
  year={2017},
  organization={ACM}
}

@inproceedings{shin201714,
  title={14.2 DNPU: An 8.1 TOPS/W reconfigurable CNN-RNN processor for general-purpose deep neural networks},
  author={Shin, Dongjoo and Lee, Jinmook and Lee, Jinsu and Yoo, Hoi-Jun},
  booktitle={Solid-State Circuits Conference (ISSCC), 2017 IEEE International},
  pages={240--241},
  year={2017},
  organization={IEEE}
}

@inproceedings{whatmough201714,
  title={14.3 a 28nm soc with a 1.2 ghz 568nj/prediction sparse deep-neural-network engine with> 0.1 timing error rate tolerance for iot applications},
  author={Whatmough, Paul N and Lee, Sae Kyu and Lee, Hyunkwang and Rama, Saketh and Brooks, David and Wei, Gu-Yeon},
  booktitle={Solid-State Circuits Conference (ISSCC), 2017 IEEE International},
  pages={242--243},
  year={2017},
  organization={IEEE}
}

@inproceedings{zhao2017aep,
  title={AEP: An error-bearing neural network accelerator for energy efficiency and model protection},
  author={Zhao, Lei and Zhang, Youtao and Yang, Jun},
  booktitle={Computer-Aided Design (ICCAD), 2017 IEEE/ACM International Conference on},
  pages={765--771},
  year={2017},
  organization={IEEE}
}

@inproceedings{reagen2016minerva,
  title={Minerva: Enabling low-power, highly-accurate deep neural network accelerators},
  author={Reagen, Brandon and Whatmough, Paul and Adolf, Robert and Rama, Saketh and Lee, Hyunkwang and Lee, Sae Kyu and Hern{\'a}ndez-Lobato, Jos{\'e} Miguel and Wei, Gu-Yeon and Brooks, David},
  booktitle={ACM SIGARCH Computer Architecture News},
  volume={44},
  number={3},
  pages={267--278},
  year={2016},
  organization={IEEE Press}
}

@inproceedings{razlighi2017looknn,
  title={Looknn: Neural network with no multiplication},
  author={Razlighi, Mohammad Samragh and Imani, Mohsen and Koushanfar, Farinaz and Rosing, Tajana},
  booktitle={2017 Design, Automation \& Test in Europe Conference \& Exhibition (DATE)},
  pages={1775--1780},
  year={2017},
  organization={IEEE}
}

@inproceedings{wang2017chain,
  title={Chain-NN: An energy-efficient 1D chain architecture for accelerating deep convolutional neural networks},
  author={Wang, Shihao and Zhou, Dajiang and Han, Xushen and Yoshimura, Takeshi},
  booktitle={2017 Design, Automation \& Test in Europe Conference \& Exhibition (DATE)},
  pages={1032--1037},
  year={2017},
  organization={IEEE}
}

@inproceedings{rouhani2017deep3,
  title={Deep3: Leveraging three levels of parallelism for efficient deep learning},
  author={Rouhani, Bita Darvish and Mirhoseini, Azalia and Koushanfar, Farinaz},
  booktitle={Proceedings of the 54th Annual Design Automation Conference 2017},
  pages={61},
  year={2017},
  organization={ACM}
}

@inproceedings{samragh2017customizing,
  title={Customizing neural networks for efficient fpga implementation},
  author={Samragh, Mohammad and Ghasemzadeh, Mohammad and Koushanfar, Farinaz},
  booktitle={Field-Programmable Custom Computing Machines (FCCM), 2017 IEEE 25th Annual International Symposium on},
  pages={85--92},
  year={2017},
  organization={IEEE}
}

@inproceedings{shen2017escher,
  title={Escher: A cnn accelerator with flexible buffering to minimize off-chip transfer},
  author={Shen, Yongming and Ferdman, Michael and Milder, Peter},
  booktitle={Field-Programmable Custom Computing Machines (FCCM), 2017 IEEE 25th Annual International Symposium on},
  pages={93--100},
  year={2017},
  organization={IEEE}
}

@inproceedings{yu2017scalpel,
  title={Scalpel: Customizing dnn pruning to the underlying hardware parallelism},
  author={Yu, Jiecao and Lukefahr, Andrew and Palframan, David and Dasika, Ganesh and Das, Reetuparna and Mahlke, Scott},
  booktitle={Proceedings of the 44th Annual International Symposium on Computer Architecture},
  pages={548--560},
  year={2017},
  organization={ACM}
}

@inproceedings{parashar2017scnn,
  title={Scnn: An accelerator for compressed-sparse convolutional neural networks},
  author={Parashar, Angshuman and Rhu, Minsoo and Mukkara, Anurag and Puglielli, Antonio and Venkatesan, Rangharajan and Khailany, Brucek and Emer, Joel and Keckler, Stephen W and Dally, William J},
  booktitle={Proceedings of the 44th Annual International Symposium on Computer Architecture},
  pages={27--40},
  year={2017},
  organization={ACM}
}

@inproceedings{shen2017maximizing,
  title={Maximizing CNN accelerator efficiency through resource partitioning},
  author={Shen, Yongming and Ferdman, Michael and Milder, Peter},
  booktitle={Proceedings of the 44th Annual International Symposium on Computer Architecture},
  pages={535--547},
  year={2017},
  organization={ACM}
}

@inproceedings{song2017pipelayer,
  title={PipeLayer: A pipelined ReRAM-based accelerator for deep learning},
  author={Song, Linghao and Qian, Xuehai and Li, Hai and Chen, Yiran},
  booktitle={High Performance Computer Architecture (HPCA), 2017 IEEE International Symposium on},
  pages={541--552},
  year={2017},
  organization={IEEE}
}

@inproceedings{bang201714,
  title={14.7 a 288$\mu$w programmable deep-learning processor with 270kb on-chip weight storage using non-uniform memory hierarchy for mobile intelligence},
  author={Bang, Suyoung and Wang, Jingcheng and Li, Ziyun and Gao, Cao and Kim, Yejoong and Dong, Qing and Chen, Yen-Po and Fick, Laura and Sun, Xun and Dreslinski, Ron and Mudge, Trevor and Kim, Hun Seok and Blaauw, David and Sylvester, Dennis},
  booktitle={Solid-State Circuits Conference (ISSCC), 2017 IEEE International},
  pages={250--251},
  year={2017},
  organization={IEEE}
}

@inproceedings{umuroglu2017finn,
  title={Finn: A framework for fast, scalable binarized neural network inference},
  author={Umuroglu, Yaman and Fraser, Nicholas J and Gambardella, Giulio and Blott, Michaela and Leong, Philip and Jahre, Magnus and Vissers, Kees},
  booktitle={Proceedings of the 2017 ACM/SIGDA International Symposium on Field-Programmable Gate Arrays},
  pages={65--74},
  year={2017},
  organization={ACM}
}

@inproceedings{alwani2016fused,
  title={Fused-layer CNN accelerators},
  author={Alwani, Manoj and Chen, Han and Ferdman, Michael and Milder, Peter},
  booktitle={Microarchitecture (MICRO), 2016 49th Annual IEEE/ACM International Symposium on},
  pages={1--12},
  year={2016},
  organization={IEEE}
}

%--
@inproceedings{zhang2016cambricon,
  title={Cambricon-X: An accelerator for sparse neural networks},
  author={Zhang, Shijin and Du, Zidong and Zhang, Lei and Lan, Huiying and Liu, Shaoli and Li, Ling and Guo, Qi and Chen, Tianshi and Chen, Yunji},
  booktitle={Microarchitecture (MICRO), 2016 49th Annual IEEE/ACM International Symposium on},
  pages={1--12},
  year={2016},
  organization={IEEE}
}

@inproceedings{judd2016stripes,
  title={Stripes: Bit-serial deep neural network computing},
  author={Judd, Patrick and Albericio, Jorge and Hetherington, Tayler and Aamodt, Tor M and Moshovos, Andreas},
  booktitle={Microarchitecture (MICRO), 2016 49th Annual IEEE/ACM International Symposium on},
  pages={1--12},
  year={2016},
  organization={IEEE}
}

@inproceedings{rhu2016vdnn,
  title={vDNN: Virtualized deep neural networks for scalable, memory-efficient neural network design},
  author={Rhu, Minsoo and Gimelshein, Natalia and Clemons, Jason and Zulfiqar, Arslan and Keckler, Stephen W},
  booktitle={Microarchitecture (MICRO), 2016 49th Annual IEEE/ACM International Symposium on},
  pages={1--13},
  year={2016},
  organization={IEEE}
}

@inproceedings{sharma2016high,
  title={From high-level deep neural models to FPGAs},
  author={Sharma, Hardik and Park, Jongse and Mahajan, Divya and Amaro, Emmanuel and Kim, Joon Kyung and Shao, Chenkai and Mishra, Asit and Esmaeilzadeh, Hadi},
  booktitle={Microarchitecture (MICRO), 2016 49th Annual IEEE/ACM International Symposium on},
  pages={1--12},
  year={2016},
  organization={IEEE}
}

@inproceedings{yazdani2016ultra,
  title={An ultra low-power hardware accelerator for automatic speech recognition},
  author={Yazdani, Reza and Segura, Albert and Arnau, Jose-Maria and Gonzalez, Antonio},
  booktitle={Microarchitecture (MICRO), 2016 49th Annual IEEE/ACM International Symposium on},
  pages={1--12},
  year={2016},
  organization={IEEE}
}

@inproceedings{clemons2016patch,
  title={A patch memory system for image processing and computer vision},
  author={Clemons, Jason and Cheng, Chih-Chi and Frosio, Iuri and Johnson, Daniel and Keckler, Stephen W},
  booktitle={Microarchitecture (MICRO), 2016 49th Annual IEEE/ACM International Symposium on},
  pages={1--13},
  year={2016},
  organization={IEEE}
}

@inproceedings{rajamanikkam2016boostnoc,
  title={BoostNoC: power efficient network-on-chip architecture for near threshold computing},
  author={Rajamanikkam, Chidhambaranathan and JS, Rajesh and Chakraborty, Koushik and Roy, Sanghamitra},
  booktitle={Proceedings of the 35th International Conference on Computer-Aided Design},
  pages={124},
  year={2016},
  organization={ACM}
}

@inproceedings{venkataramani2016efficient,
  title={Efficient embedded learning for IoT devices},
  author={Venkataramani, Swagath and Roy, Kaushik and Raghunathan, Anand},
  booktitle={Design Automation Conference (ASP-DAC), 2016 21st Asia and South Pacific},
  pages={308--311},
  year={2016},
  organization={IEEE}
}

@inproceedings{wang2016re,
  title={Re-architecting the on-chip memory sub-system of machine-learning accelerator for embedded devices},
  author={Wang, Ying and Li, Huawei and Li, Xiaowei},
  booktitle={Proceedings of the 35th International Conference on Computer-Aided Design},
  pages={13},
  year={2016},
  organization={ACM}
}

@inproceedings{han2016eie,
  title={EIE: efficient inference engine on compressed deep neural network},
  author={Han, Song and Liu, Xingyu and Mao, Huizi and Pu, Jing and Pedram, Ardavan and Horowitz, Mark A and Dally, William J},
  booktitle={Computer Architecture (ISCA), 2016 ACM/IEEE 43rd Annual International Symposium on},
  pages={243--254},
  year={2016},
  organization={IEEE}
}

@inproceedings{chen2016eyeriss,
  title={Eyeriss: A spatial architecture for energy-efficient dataflow for convolutional neural networks},
  author={Chen, Yu-Hsin and Emer, Joel and Sze, Vivienne},
  booktitle={ACM SIGARCH Computer Architecture News},
  volume={44},
  number={3},
  pages={367--379},
  year={2016},
  organization={IEEE Press}
}

@inproceedings{albericio2016cnvlutin,
  title={Cnvlutin: Ineffectual-neuron-free deep neural network computing},
  author={Albericio, Jorge and Judd, Patrick and Hetherington, Tayler and Aamodt, Tor and Jerger, Natalie Enright and Moshovos, Andreas},
  booktitle={ACM SIGARCH Computer Architecture News},
  volume={44},
  number={3},
  pages={1--13},
  year={2016},
  organization={IEEE Press}
}

@article{chen2014diannao,
  title={Diannao: A small-footprint high-throughput accelerator for ubiquitous machine-learning},
  author={Chen, Tianshi and Du, Zidong and Sun, Ninghui and Wang, Jia and Wu, Chengyong and Chen, Yunji and Temam, Olivier},
  journal={ACM Sigplan Notices},
  volume={49},
  number={4},
  pages={269--284},
  year={2014},
  publisher={ACM}
}

@inproceedings{chen2014dadiannao,
  title={Dadiannao: A machine-learning supercomputer},
  author={Chen, Yunji and Luo, Tao and Liu, Shaoli and Zhang, Shijin and He, Liqiang and Wang, Jia and Li, Ling and Chen, Tianshi and Xu, Zhiwei and Sun, Ninghui and Temam, Olivier},
  booktitle={Proceedings of the 47th Annual IEEE/ACM International Symposium on Microarchitecture},
  pages={609--622},
  year={2014},
  organization={IEEE Computer Society}
}

@inproceedings{du2015shidiannao,
  title={ShiDianNao: Shifting vision processing closer to the sensor},
  author={Du, Zidong and Fasthuber, Robert and Chen, Tianshi and Ienne, Paolo and Li, Ling and Luo, Tao and Feng, Xiaobing and Chen, Yunji and Temam, Olivier},
  booktitle={ACM SIGARCH Computer Architecture News},
  volume={43},
  number={3},
  pages={92--104},
  year={2015},
  organization={ACM}
}

@article{lecun1998gradient,
  title={Gradient-based learning applied to document recognition},
  author={LeCun, Yann and Bottou, L{\'e}on and Bengio, Yoshua and Haffner, Patrick},
  journal={Proceedings of the IEEE},
  volume={86},
  number={11},
  pages={2278--2324},
  year={1998},
  publisher={IEEE}
}

@inproceedings{esser2015backpropagation,
  title={Backpropagation for energy-efficient neuromorphic computing},
  author={Esser, Steve K and Appuswamy, Rathinakumar and Merolla, Paul and Arthur, John V and Modha, Dharmendra S},
  booktitle={Advances in Neural Information Processing Systems},
  pages={1117--1125},
  year={2015}
}

@MISC{vivaldo-dataset,
      TITLE = "7 Series DSP 48E1 Slice ",
      NOTE = "\url{https://www.xilinx.com/support/documentation/user_guides/ug479_7Series_DSP48E1.pdf}"
}

@article{luo2017thinet,
  title={Thinet: A filter level pruning method for deep neural network compression},
  author={Luo, Jian-Hao and Wu, Jianxin and Lin, Weiyao},
  journal={arXiv preprint arXiv:1707.06342},
  year={2017}
}

@inproceedings{he2017channel,
  title={Channel pruning for accelerating very deep neural networks},
  author={He, Yihui and Zhang, Xiangyu and Sun, Jian}
}

@article{ruder2016overview,
  title={An overview of gradient descent optimization algorithms},
  author={Ruder, Sebastian},
  journal={arXiv preprint arXiv:1609.04747},
  year={2016}
}

@inproceedings{lin2017binary,
  title={Binarized Convolutional Neural Networks with Separable Filters for Efficient Hardware Acceleration},
  author={ELin, J. H., Xing, T., Zhao, R., Zhang, Z., Srivastava, M. B., Tu, Z., Gupta, R. K.},
  booktitle={CVPR Workshops},
  pages={344--352},
  year={2017}
}

@article{loshchilov2016sgdr,
  title={SGDR: stochastic gradient descent with restarts},
  author={Loshchilov, Ilya and Hutter, Frank},
  journal={Learning},
  volume={10},
  pages={3},
  year={2016}
}

@article{lin2015neural,
  title={Neural networks with few multiplications},
  author={Lin, Zhouhan and Courbariaux, Matthieu and Memisevic, Roland and Bengio, Yoshua},
  journal={arXiv preprint arXiv:1510.03009},
  year={2015}
}

@article{gysel2016hardware,
  title={Hardware-oriented approximation of convolutional neural networks},
  author={Gysel, Philipp and Motamedi, Mohammad and Ghiasi, Soheil},
  journal={arXiv preprint arXiv:1604.03168},
  year={2016}
}

@article{baiduhotchips,
  title={XPU: A Programmable FPGA Accelerator for Diverse Workloads},
  author={Ouyang, Jian and Wu, Ephrem and Wang, Jing and Li, Yupeng and Xie, Hanlin},
  journal={Hot Chips},
  year={2017}
}

@article{truenorthcifar,
  title={Convolutional networks for fast, energy-efficient neuromorphic computing},
  author={Steven K. Esser, Paul A. Merolla, John V. Arthur, Andrew S. Cassidy, Rathinakumar Appuswamy, Alexander Andreopoulos, David J. Berg, Jeffrey L. McKinstry, Timothy Melano, Davis R. Barch, Carmelo di Nolfo, Pallab Datta, Arnon Amir, Brian Taba, Myron D. Flickner, and Dharmendra S. Modha},
  journal={National Academy of Sciences},
  year={2016}
}

@MISC{synopsysdesigncompiler,
      TITLE = "Design Compiler: RTL Synthesis ",
      NOTE = "\url{https://www.synopsys.com/support/training/rtl-synthesis/design-compiler-rtl-synthesis.html}"
}

@MISC{nangatelib,
      TITLE = "NanGate FreePDK45 Open Cell Library ",
      NOTE = "\url{http://www.nangate.com/?page_id=2325}"
}

@MISC{cacti,
      TITLE = "CACTI: An integrated cache and memory access time, cycle time, area, leakage, and dynamic power model",
      NOTE = "\url{https://github.com/HewlettPackard/cacti}"
}
@MISC{vivado,
      TITLE = "Vivado Design Suite - HLx Editions Productivity. Multiplied",
      NOTE = "\url{https://www.xilinx.com/products/design-tools/vivado.html}"
}
@MISC{xcku035,
      TITLE = "Xilinx Inc. XCKU035-1FBVA676C",
      NOTE = "\url{https://www.digikey.ca/product-detail/en/xilinx-inc/XCKU035-1FBVA676C/122-1989-ND/6132038}"
}

\end{document}